\documentclass[10pt,twocolumn,letterpaper]{article}

\usepackage{cvpr}              

\definecolor{cvprblue}{rgb}{0.21,0.49,0.74}
\usepackage[pagebackref,breaklinks,colorlinks,allcolors=cvprblue]{hyperref}

\def\paperID{17582} 
\def\confName{CVPR}
\def\confYear{2025}

\title{Lessons and Insights from a Unifying Study of \\ Parameter-Efficient Fine-Tuning (PEFT) in Visual Recognition}

\author{%
  \textbf{Zheda Mai$^{1}$}, \textbf{Ping Zhang$^{1}$}, \textbf{Cheng-Hao Tu$^1$}, \textbf{Hong-You Chen$^1$}, \textbf{Li Zhang$^2$}, \textbf{Wei-Lun Chao$^1$} \\
  $^1$The Ohio State University, $^2$Google Research. 
}

\usepackage[utf8]{inputenc} 
\usepackage[T1]{fontenc}    
\usepackage{hyperref}       
\usepackage{url}            
\usepackage{booktabs}       
\usepackage{amsfonts}       
\usepackage{nicefrac}       
\usepackage{microtype}      
\usepackage{xcolor}         
\usepackage{colortbl}

\usepackage[linesnumbered,ruled,vlined,algo2e]{algorithm2e}

\usepackage{wrapfig}
\usepackage{amssymb}
\usepackage{pifont}
\usepackage{bbding}
\usepackage{graphicx}
\usepackage{subcaption}
\usepackage{float}
\usepackage{wrapfig}

\usepackage{amssymb}
\usepackage{amsmath,amsfonts}
\usepackage{amsopn}
\usepackage{bm} 
\usepackage{multirow}
\usepackage{flushend}
\usepackage{tabularx}

\newcommand{\vct}[1]{\boldsymbol{#1}} 
\newcommand{\mat}[1]{\boldsymbol{#1}} 
\newcommand{\cst}[1]{\mathsf{#1}}  

\newcommand{\field}[1]{\mathbb{#1}}
\newcommand{\R}{\field{R}} 

\newcommand{\ProbOpr}[1]{\mathbb{#1}}

\newcommand{\expect}[2]{%
\ifthenelse{\equal{#2}{}}{\ProbOpr{E}_{#1}}
{\ifthenelse{\equal{#1}{}}{\ProbOpr{E}\left[#2\right]}{\ProbOpr{E}_{#1}\left[#2\right]}}} 

\newcommand{\vb}{\vct{b}}

\newcommand{\vx}{{\vct{x}}}

\newcommand{\vw}{\vct{w}}

\newcommand{\mU}{\mat{U}}
\newcommand{\mA}{\mat{A}}
\newcommand{\mB}{\mat{B}}

\newcommand{\mW}{\mat{W}}

\newcommand{\mZ}{\mat{Z}}
\newcommand{\mQ}{\mat{Q}}
\newcommand{\mP}{\mat{P}}

\newcommand{\mI}{\mat{I}}
\newcommand{\mK}{\mat{K}}
\newcommand{\mV}{\mat{V}}

\newcommand{\eat}[1]{}
\newcommand{\method}[1]{\textsc{#1}}

\usepackage{graphicx}
\usepackage{arydshln}
\usepackage{subcaption} 
\usepackage{tablefootnote}
\usepackage[stable]{footmisc}
\usepackage{makecell}
\newcommand\mypara[1]{\vspace{0.6mm}\noindent\textbf{#1}}
\usepackage{cellspace}
\usepackage{multirow}

\usepackage[utf8]{inputenc} 
\usepackage[T1]{fontenc}    
\usepackage{hyperref}       
\usepackage{url}            
\usepackage{booktabs}       
\usepackage{amsfonts}       
\usepackage{nicefrac}       
\usepackage{microtype}      
\usepackage{xcolor}         
\usepackage{colortbl}

\usepackage{wrapfig}
\usepackage{amssymb}
\usepackage{pifont}
\usepackage{bbding}
\usepackage{graphicx}
\usepackage{subcaption}
\usepackage{float}
\usepackage{wrapfig}
\usepackage{graphicx}
\usepackage{arydshln}
\usepackage{subcaption} 
\usepackage{tablefootnote}
\usepackage[stable]{footmisc}
\usepackage{makecell}
\usepackage{cellspace}
\usepackage{multirow}

\newcommand{\zheda}[1]{{\color{magenta}[Zheda: #1]}}

\usepackage{tikz}
\usepackage{stackengine}

\usepackage{stackengine}
\usepackage{stfloats}
\usepackage{placeins}

\stackMath
\newcommand\crossline{%
  \stackon[-0.9ex]{\textcolor{red}{\boldsymbol{\times}}}{\textcolor{red}{\rule{2ex}{0.3mm}}}%
}
\newcommand{\greentriangle}{\tikz\fill[green] (0,0) -- (1ex,0) -- (0.5ex,1ex) -- cycle;}
\newcommand{\redcircle}{\tikz\draw[red, fill=red] (0,0) circle (0.5ex);}

\author{%
  \textbf{Zheda Mai$^{1}$}, \textbf{Ping Zhang$^{1}$}, \textbf{Cheng-Hao Tu$^1$},\\ \textbf{Hong-You Chen$^1$}, \textbf{Quang-Huy Nguyen$^1$}, \textbf{Li Zhang$^2$}, \textbf{Wei-Lun Chao$^1$} \\
  $^1$The Ohio State University, $^2$Google Research. 
}

\definecolor{mydarkgreen}{rgb}{0,0.7,0}

\begin{document}
\maketitle

\begin{abstract}
Parameter-efficient fine-tuning (PEFT) has attracted significant attention due to the growth of pre-trained model sizes and the need to fine-tune (FT) them for superior downstream performance. Despite a surge in new PEFT methods, a systematic study to understand their performance and suitable application scenarios is lacking, leaving questions like~\emph{``when to apply PEFT''}~and~\emph{``which method to use''}~largely unanswered, especially in visual recognition. In this paper, we conduct a unifying empirical study of representative PEFT methods with Vision Transformers. We systematically tune their hyperparameters to fairly compare their accuracy on downstream tasks. Our study offers a practical user guide and unveils several new insights. First, if tuned carefully, different PEFT methods achieve similar accuracy in the low-shot benchmark VTAB-1K. This includes simple approaches like FT the bias terms that were reported inferior. Second, despite similar accuracy, we find that PEFT methods make different mistakes and high-confidence predictions, likely due to their different inductive biases. Such an inconsistency (or complementarity) opens up the opportunity for ensemble methods, and we make preliminary attempts at this. Third, going beyond the commonly used low-shot tasks, we find that PEFT is also useful in many-shot regimes, achieving comparable or better accuracy than full FT while using significantly fewer parameters. Lastly, we investigate PEFT's ability to preserve a pre-trained model's robustness to distribution shifts (\eg, CLIP). Perhaps not surprisingly, PEFT approaches outperform full FT alone. However, with weight-space ensembles, full FT  can better balance target distribution and distribution shift performance, suggesting a future research direction for robust PEFT\footnote{\url{https://github.com/OSU-MLB/ViT_PEFT_Vision}.   }. 

\end{abstract}

\section{Introduction}
{Pre-training and then fine-tuning (FT)} has become the standard practice to tackle visual recognition problems \citep{bommasani2021opportunities}. 
The community-wide enthusiasm for open-sourcing has made it possible to access large, powerful pre-trained models learned from a gigantic amount of data, e.g., ImageNet-21K~\citep{ridnik2021imagenet} or LAION-5B~\citep{schuhmann2022laion}. 
More research focus has thus been on how to FT such large models~\citep{yu2023visual}. Among existing efforts, parameter-efficient transfer learning (PEFT), has attracted increasing attention lately~\citep{han2024parameter, ding2023parameter}. Instead of FT the whole model (\ie, full FT) or the last fully connected layer (\ie, linear probing), PEFT approaches seek to update or insert a relatively small number of parameters to the pre-trained model \citep{xin2024parameter}. Doing so has several noticeable advantages. First, as named, PEFT is parameter-efficient. For one downstream task (\eg, recognizing bird species or car brands), it only needs to learn and store a tiny fraction of parameters on top of the pre-trained model. Second, accuracy-wise, PEFT has been shown to consistently outperform linear probing and often beat full FT, as reported on the commonly used low-shot image classification benchmark VTAB-1K~\citep{zhai2019large}.


\begin{figure*}[t]
\centering
\begin{subfigure}{0.39\linewidth}
  \centering
  \includegraphics[width=\linewidth]{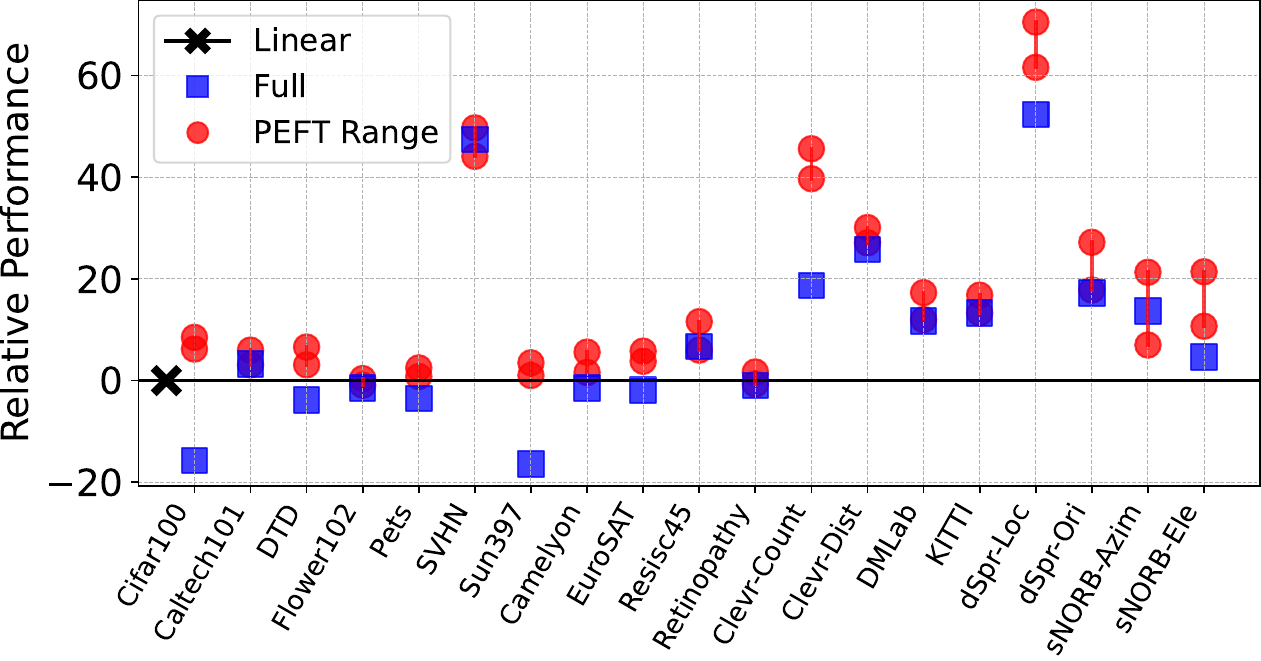}
  \caption{Accuracy gain vs.~linear probing on VTAB-1K (19 tasks)}
  \label{fig:vtab_vis}
\end{subfigure}
\hfill
\begin{subfigure}{0.275\linewidth}
  \centering
  \includegraphics[width=0.8\linewidth]{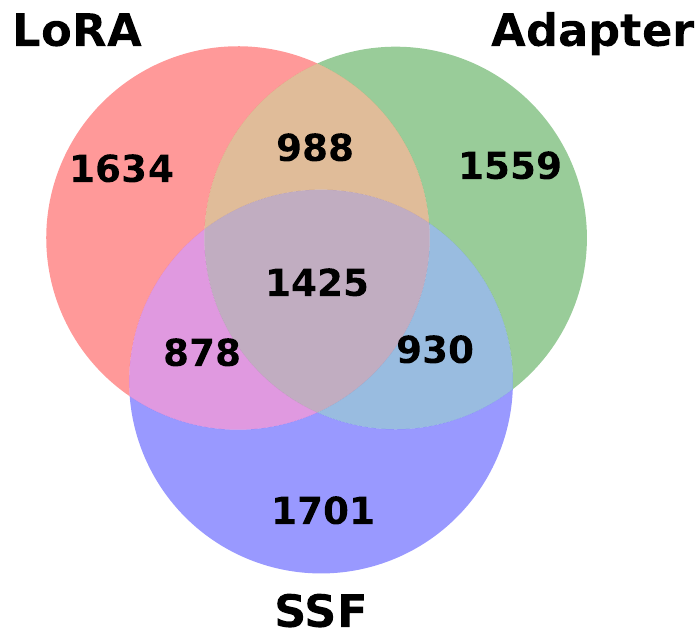}
  \caption{Prediction overlaps (5K most confident)}
  \label{fig:highconf}
\end{subfigure}
\hfill
\begin{subfigure}{0.325\linewidth}
  \centering
  \includegraphics[width=\linewidth]{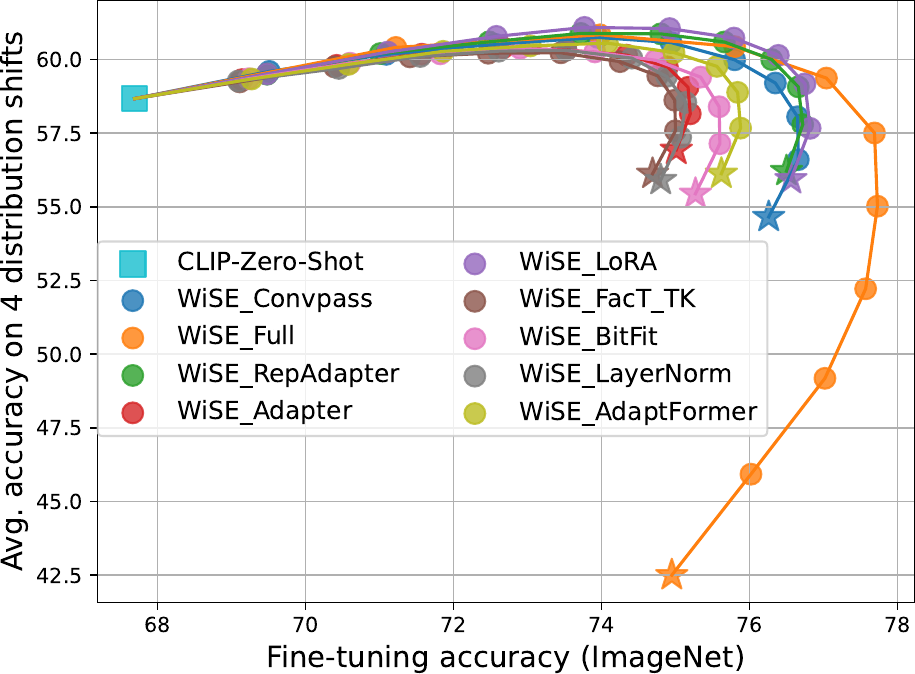}
  \caption{Target distribution vs.~distribution shifts}
  \label{fig:merge}
\end{subfigure}

\caption{\small Highlights of our insights. \textbf{(a) Downstream accuracy:} with proper implementation and fair tuning, different PEFT methods achieve similar accuracy ({\color{red}$\bullet$-$\bullet$}: the range from the most to the least accurate methods) and consistently outperform linear probing ({$\times$}) and full FT ({\color{blue}$\blacksquare$}) on VTAB-1K. \textbf{(b) Diverse predictions:} despite reaching similar downstream performance, different PEFT methods produce diverse predictions. This opens new opportunities for ensemble approaches and other learning paradigms (\eg semi-supervised learning) that can exploit the prediction discrepancies. \textbf{(c) Distribution shift accuracy:} FT a CLIP ViT-B/16, known for its generalizability across domains, with PEFT on ImageNet-1K (100 samples/class) better preserves the distribution shift accuracy (Y-axis, averaged across ImageNet-(V2, S, R, A) than {\color{orange}full FT}, evidenced by the $\star$ points. Interestingly,  \emph{weight-space ensembles (WiSE)~\citep{wortsman2022robust} is applicable between PEFT's FT model and the pre-trained model} ({\color{cyan}$\blacksquare$}), but not as effective as applying it to the fully FT model.  Details are in~\autoref{sec:few}, \autoref{sec:complememtary} and~\autoref{sec: robust}. }
\vskip -7pt
\end{figure*}

To date, a plethora of PEFT approaches have been proposed, bringing in inspiring ideas and promising results. Along with this come several excellent surveys that summarize existing PEFT approaches \citep{yu2023visual, xin2024parameter, ding2023parameter}. \emph{Yet, a systematic understanding of the PEFT paradigm seems still missing.} For example, with so many PEFT approaches, there is a lack of unifying references for when and how to apply them. 
Though superior accuracy was reported on the low-shot benchmark VTAB-1K, there is not much discussion on how PEFT approaches achieve it. 
Does it result from PEFT's ability to promote transferability or prevent over-fitting?  
The current evaluation also raises the question of whether PEFT is useful beyond a low-shot scenario. 
Last but not least, besides superior accuracy, do existing PEFT approaches offer different, ideally, complementary information? 



Attempting to answer these questions, we conduct a unifying empirical study of representative PEFT methods for Vision Transformers~\citep{dosovitskiy2020image}, including Low-Rank Adaptation (LoRA)~\citep{hu2021lora}, Visual Prompt Tuning (VPT)~\citep{jia2022visual}, Adapter~\citep{houlsby2019parameter}, and \textit{ten} other approaches.  
We systematically tune their hyperparameters to fairly compare their accuracy on the low-shot benchmark VTAB-1K.
This includes learning rate, weight decay, and method-specific parameters like the PEFT module sizes.
Besides VTAB-1K, we examine PEFT methods on full-size downstream datasets such as CIFAR-100~\citep{krizhevsky2009learning}, RESISC for remote sensing~\citep{cheng2017remote}, and Clevr-Distance for depth classification~\citep{zhai2019large, johnson2017clevr}. We also conduct a study on ImageNet~\citep{deng2009imagenet} and its variants with domain shifts~\citep{hendrycks2021faces, Gao_2023, hendrycks2021natural, recht2019imagenet} for robustness evaluation. 

We summarize our key findings and analyses as follows:

\noindent\textbf{Representative PEFT methods perform similarly on VTAB-1K, when properly implemented and tuned (\autoref{fig:vtab_vis}).} This includes methods previously considered less effective, such as BitFit~\citep{zaken2022bitfit}, which FT only the bias terms of the frozen backbone. Methods originally proposed for NLP, like Adapter \citep{houlsby2019parameter} and LoRA~\citep{hu2021lora} also exhibit impressive performance when their bottleneck dimensions are carefully tuned. Among all the hyperparameters, we find the {drop path rate}~\citep{huang2016deep} particularly important. Ignoring it (\ie, setting it to $0$) significantly degrades the performance, potentially due to over-fitting. Overall, PEFT methods consistently outperform linear probing and full FT on all 19 image classification tasks (with $1,000$ training samples) in VTAB-1K.



\noindent\textbf{While similarly accurate on average, PEFT approaches make different predictions (\autoref{fig:highconf}).} The above finding seems daunting: \emph{if existing PEFT approaches all perform similarly in terms of accuracy, do we learn anything useful beyond a single approach?} This is particularly worrisome given that they FT the same backbone using the same downstream data. Fortunately, our analysis shows that different PEFT methods learn differently from the same data, resulting in \textit{diverse prediction errors and confidence}. We attribute this to their inductive bias differences \citep{neyshabur2014search} --- they explicitly specify different parameters to be updated or inserted. This opens up the door to leverage their discrepancy for improvement, \eg, through \textit{ensemble methods~\citep{dietterich2000ensemble,zhou2012ensemble} or co-training~\citep{blum1998combining,balcan2004co}} and we provide preliminary studies.
\noindent\textbf{PEFT is also effective in many-shot regimes.} We extend PEFT beyond low-shot regime and find it effective even with ample downstream training data --- {PEFT can be on par or surpass full FT}. This suggests that adjusting only a fraction of parameters in a suitably pre-trained backbone (\eg, pre-trained on ImageNet-21K~\citep{dosovitskiy2020image}) could already offer a sufficient capacity~\citep{zhang2021understanding} to reach a performant hypothesis for downstream tasks. 


\noindent\textbf{PEFT appears more robust than full FT to distribution shifts, but WiSE overturns this advantage (\autoref{fig:merge}).}
We also assess PEFT's robustness to distribution shifts, following~\citep{wortsman2022robust}. We consider a CLIP backbone~\citep{radford2021learning}, known for its superior generalizability to distribution shifts, and FT it with PEFT on ImageNet-1K. We found that PEFT retains CLIP's generalizability (\eg, to samples from ImageNet-(V2, S, R, A)) better than full FT. This may not be surprising. What is interesting is that the weight-space ensembles (WiSE) between the FT and pre-trained models~\citep{wortsman2022robust} is compatible with PEFT as well to further improve the robustness without sacrificing the target accuracy. To the best of our knowledge, \textit{we are the first to explore WiSE for PEFT.} Nevertheless, full FT with WiSE can achieve even higher accuracy in both downstream and distribution shift data than PEFT, suggesting a further research direction in robust PEFT.


\noindent\textbf{What lead to PEFT's success?} We attempt to answer this fundamental question by analyzing the findings in our study. On VTAB-1K with 19 tasks, we identify two scenarios: (1) in certain tasks, full FT outperforms linear probing, suggesting the need to update the backbone; (2) in other tasks, linear probing outperforms full FT, suggesting either the backbone is good enough or updating it risks over-fitting. The superior accuracy of PEFT in both scenarios suggests that PEFT acts as an \emph{effective regularizer} during low-shot training. Still using VTAB but with ample training data, we find that for scenario (1) tasks, PEFT performs similarly with full FT, suggesting that its regularization role does not impede model learning from abundant data. For scenario (2) tasks, PEFT can surprisingly still outperform full FT, indicating that it effectively transfers (or preserves) crucial pre-trained knowledge that full FT may discard. Overall, PEFT succeeds as a \textbf{high-capacity learner} equipped with an \textbf{effective regularizer}.

\noindent\textbf{Contributions.} Instead of chasing the leaderboard, we systematically scrutinize existing methods via a unifying study. Our contribution is thus not a technical novelty, but: (1) a \textit{systematic framework} for reproducible evaluations of PEFT methods; (2) a set of  \textit{empirical recommendations} on when and how to use PEFT methods for practitioners (\autoref{sec:few}, \autoref{sec:many}); (3) \textit{new insights for future research} including leveraging PEFT's {prediction differences} (\autoref{sec:complememtary}) and exploring {robust fine-tuning} (\autoref{sec: robust}).

\section{Background}

\subsection{Large pre-trained models} 


Building upon networks with millions (or billions) of parameters and massive training data, large pre-trained models have led to groundbreaking results in various downstream tasks~\citep{liang2024foundation, moor2023foundation} and shown emerging capabilities not observed previously~\citep{khan2022transformers, li2024multimodal, bommasani2021opportunities}. For example, a Vision Transformer (ViT)~\citep{dosovitskiy2020image} trained with ImageNet-21K ($14$M images) leads to consistent gains v.s.~a ViT trained with ImageNet-1K ($1.3$M images)~\citep{dosovitskiy2020image}. ViTs pre-trained with millions of image-text pairs via a contrastive objective function (\eg, CLIP-ViT)~\citep{radford2021learning, cherti2023reproducible} show an unprecedented zero-shot capability and robustness to distribution shifts~\citep{radford2021learning}. We focus on the ImageNet-21K-ViT and CLIP-ViT in this paper.

\noindent\textbf{Vision Transformer (ViT).} A ViT contains $M$ Transformer layers consisting of a multi-head self-attention (MSA) block, a multi-level perceptron (MLP) block, two Layer Normalization (LN) blocks~\citep{ba2016layer}, and two residual links. The $m$-th Transformer layer can be formulated as
\begin{align}
\mZ_{m}^{\prime} & =\operatorname{MSA}\left(\operatorname{LN}\left(\mZ_{m-1}\right)\right)+\mZ_{m-1}, \\
\mZ_{m} & =\operatorname{MLP}\left(\operatorname{LN}\left(\mZ_{m}^{\prime}\right)\right)+\mZ_{m}^{\prime}, \label{eqeq_MLP}
\end{align}
where $\mZ_{m-1}$ is the output of the preceding $(m-1)$-th Transformer layer. Without loss of generality, let us consider a single-head MSA where the input $\mZ$ is first projected into three matrices, $\mQ$, $\mK$, and $\mV$. The output of this block is then formulated as:
\begin{align}
\mQ = \mW_Q \mZ,& \quad \mK = \mW_K \mZ, \quad \mV = \mW_V \mZ, \label{eq_project} \\
&\mV\times\cst{Softmax}(\frac{\mK^\top\mQ}{\sqrt{D}}). 
\label{eq_MSA}
\end{align}

\subsection{Parameter-Efficient Fine-Tuning (PEFT)}

\label{sec:PEFT-all}
Fine-tuning is arguably the most common way to tailor a pre-trained model for downstream tasks~\cite{zhuang2020comprehensive, wortsman2022robust, tu2024holistic}. As the size of pre-trained models gets larger, updating and storing all the parameters for one downstream task becomes inefficient. PEFT has thus emerged as a promising paradigm.  PEFT was originally developed in NLP~\citep{he2021towards,lester2021power,he2022hyperprompt,mao2022unipelt,sung2021training,zaken2022bitfit,asai2022attentional,vu2022spot,liu2022ptuning,su2022transferability,zhong2022panda} and has attracted increasing attention in vision~\citep{jia2022visual,chen2022adaptformer,convpass,zhang2022neural,liu2022polyhistor,ssf}.
Existing approaches can generally be categorized into four groups: prompt-based, adapter-based, direct selective tuning, and efficient selective tuning. 
\emph{We focus on visual recognition and compare representative PEFT approaches applicable to ViTs.}



\noindent\textbf{Prompt-based.} Prompt-based method emerged in NLP~\citep{promptnlpsurvey, petlnkp} whose core concept is augmenting the input data with task-specific prompts. \textbf{Visual Prompt Tuning (VPT)}~\citep{jia2022visual} adapts such an idea to ViTs. Its deep version (VPT-Deep) prepends a set of soft prompts to the input tokens of each Transformer layer (\ie, $\{\mZ_m\}_{m=0}^{M-1}$) and only optimizes the prompts during FT. Other representative works in this category include~\citep{tu2023visual, yu2023visualtuning,promptVLsurvey, tsai2023convolutional, bahng2022exploring, wang2024lion}.  

\noindent\textbf{Adapter-based.}
They typically introduce additional trainable parameters to a frozen pre-trained model~\citep{petlnkp}. Initially developed for domain adaptation~\citep{rebuffi2017learning, rebuffi2018efficient} and continual learning~\citep{rosenfeld2018incremental, mai2022online}, they have been extended to adapt Transformer-based models in NLP and vision~\citep{houlsby2019parameter, liu2022polyhistor,xin2024parameter,zhao2024sct, yu2023visualtuning}. We consider five popular methods. As the first adapter-based method, \textbf{Houl.~Adapter} ~\citep{houlsby2019parameter} inserts two Adapters (a two-layer bottleneck-structured MLP with a residual link) into each Transformer layer, one after the MSA block and one after the MLP block. 
\textbf{Pfeif.~Adapter}~\citep{pfeiffer2021adapterfusion} inserts the Adapter solely after the MLP block, shown effective in recent studies~\citep{hu2021lora}.
\textbf{AdaptFormer}~\citep{chen2022adaptformer} adds an Adapter in parallel with the original MLP block, different from the sequential design of previous methods. \textbf{ConvPass}~\citep{convpass} inserts a convolutional-based Adapter that explicitly encodes visual inductive biases by performing 2D convolution over nearby patch tokens. This module is inserted parallel to the MSA and MLP blocks.
\textbf{RepAdapter}~\citep{luo2023rep} introduces linear Adapters with group-wise transformations~\citep{luo2022towards}, placed sequentially after MSA and MLP.


\noindent\textbf{Direct selective tuning.} They selectively update a subset of parameters of the backbone, striking a balance between full FT and linear probing. We consider three approaches. 
\textbf{BitFit}~\citep{zaken2022bitfit} updates the bias terms, including those in the patch embeddings projection, the Q/K/V weights, the MLP and LN blocks.
\textbf{LayerNorm}~\citep{ln-tune} updates the parameters of the LN blocks. \textbf{DiffFit}~\citep{xie2023difffit} updates both the bias terms and the LN blocks and inserts learnable factors to scale the features after the MSA and the MLP blocks. Instead of updating parameters, \textbf{SSF}~\citep{ssf} linearly adapts intermediate features, motivated by feature modulation~\citep{huang2017arbitrary}. 


\noindent\textbf{Efficient selective tuning.} Instead of directly updating parameters, these methods learns \emph{additive residuals} (\eg, $\Delta \mW$) to the original parameters (\eg, $\mW$). By injecting a low-rank constraint to the residuals, this category effectively reduces the learnable parameters.
\textbf{LoRA}~\citep{hu2021lora}, arguably the most well-known approach, parameterizes the residuals by low-rank decomposition to update the weights. Concretely, to update a $\mW\in\R^{D\times D}$ matrix, LoRA learns $\mW_{\text{down}}\in\R^{r \times D}$ and $\mW_{\text{up}}\in\R^{D \times r}$ with $r \ll D$, and forms the additive residual by $\Delta \mW=\mW_{\text{up}}\mW_{\text{down}}\in\R^{D \times D}$.
\textbf{FacT}~\citep{jie2023fact} extends the idea of matrix decomposition into tensor decomposition. It stacks the $D\times D$ learnable matrices in all the Transformer layers into a 3D tensor and learns an additive residual parameterized by Tensor-Train (TT)~\citep{tt} and Tucker (TK)~\citep{tk} formulations. More detailed summary of ViT and a survey of PEFT methods can be found in \autoref{-sec: survey}.

\begin{table*}[t]
\normalsize
\centering
\resizebox{\textwidth}{!}{
\begin{tabular}{l|ccccccc:c|cccc:c|cccccccc:c|c|c}
\toprule
& \multicolumn{8}{c|}{Natural}&\multicolumn{5}{c|}{Specialized}&\multicolumn{9}{c}{Structured}&\\
Method& {\rotatebox[origin=l]{90}{\colorbox{orange!30}{CIFAR-100}}}&{\rotatebox[origin=l]{90}{\colorbox{blue!30}{Caltech101}}}&{\rotatebox[origin=l]{90}{\colorbox{orange!30}{DTD}}}&{\rotatebox[origin=l]{90}{\colorbox{orange!30}{Flowers102}}}&{\rotatebox[origin=l]{90}{\colorbox{orange!30}{Pets}}}&{\rotatebox[origin=l]{90}{\colorbox{blue!30}{SVHN}}}&{\rotatebox[origin=l]{90}{\colorbox{orange!30}{Sun397}}}&{\rotatebox[origin=l]{90}{Mean}}&{\rotatebox[origin=l]{90}{\colorbox{orange!30}{Camelyon}}}&{\rotatebox[origin=l]{90}{\colorbox{orange!30}{EuroSAT}}}&{\rotatebox[origin=l]{90}{\colorbox{blue!30}{Resisc45}}}&{\rotatebox[origin=l]{90}{\colorbox{orange!30}{Retinopathy}}}&{\rotatebox[origin=l]{90}{Mean}}&{\rotatebox[origin=l]{90}{\colorbox{blue!30}{Clevr-Count}}}&{\rotatebox[origin=l]{90}{\colorbox{blue!30}{Clevr-Dist}}}&{\rotatebox[origin=l]{90}{\colorbox{blue!30}{DMLab}}}&{\rotatebox[origin=l]{90}{\colorbox{blue!30}{KITTI-Dist}}}&{\rotatebox[origin=l]{90}{\colorbox{blue!30}{dSpr-Loc}}}&{\rotatebox[origin=l]{90}{\colorbox{blue!30}{dSpr-Ori}}}&{\rotatebox[origin=l]{90}{\colorbox{blue!30}{sNORB-Azim}}}&{\rotatebox[origin=l]{90}{\colorbox{blue!30}{sNORB-Elev}}}&{\rotatebox[origin=l]{90}{Mean}}&{\rotatebox[origin=l]{90}{Overall Mean}} & {\rotatebox[origin=l]{90}{Tunable Params}} \\
\midrule
Linear         & 78.1     & 86.6       & 65.7 & 98.9      & 89.3 & 41.5 & 53.2   & 72.5 & 83.1     & 90.0      & 74.9     & 74.6        & 80.6 & 37.5        & 35.1       & 36.5  & 64.6  & 16.2     & 29.4     & 17.3       & 23.7      & 32.5 & 61.9    & 0         \\
Full           & 62.4     & 89.9       & 61.9 & 97.4      & 85.8 & 88.9 & 36.8   & 76.7 & 81.6     & 88.1    & 81.6     & 73.6        & 81.2 & 56.2        & 60.9       & 48.2  & 77.9  & 68.5     & 46.6     & 31.0         & 28.3      & 52.2 & 70.0    & 85.8      \\
\midrule
\textcolor{gray}{VPT-Shallow}    & \textcolor{gray}{80.2}     & \textcolor{gray}{88.7}       & \textcolor{gray}{67.9} & \textcolor{gray}{99.1}      & \textcolor{gray}{89.6} & \textcolor{gray}{77.0}   & \textcolor{gray}{54.2}   & \textcolor{gray}{79.4} & \textcolor{gray}{81.8}     & \textcolor{gray}{90.3}    & \textcolor{gray}{77.2}     & \textcolor{gray}{74.4}        & \textcolor{gray}{80.9} & \textcolor{gray}{42.2}        & \textcolor{gray}{52.4}       & 38    & \textcolor{gray}{66.5}  & \textcolor{gray}{52.4}     & \textcolor{gray}{43.1}     & \textcolor{gray}{15.2}       & \textcolor{gray}{23.2}      & \textcolor{gray}{41.6} & \textcolor{gray}{67.3}    & \textcolor{gray}{0.07}      \\
VPT-Deep       & 84.8     & 91.5       & 69.4 & 99.1      & 91.0   & 85.6 & 54.7   & 81.8 & 86.4     & 94.9    & 84.2     & 73.9        & 84.9 & 79.3        & 62.4       & 48.5  & 77.9  & 80.3     & 56.4     & 33.2       & 43.8      & 60.2 & 75.6    & 0.43      \\
\midrule
BitFit         & 86.5     & 90.5       & 70.3 & 98.9      & 91.0   & 91.2 & 54.2   & 82.6 & 86.7     & 95.0      & 85.3     & 75.5        & 85.6 & 77.2        & 63.2       & 51.2  & 79.2  & 78.6     & 53.9     & 30.1       & 34.7      & 58.5 & 75.6    & 0.1       \\
DiffFit        & 86.3     & 90.2       & 71.2 & 99.2      & 91.7 & 91.2 & 56.1   & 83.2 & 85.8     & 94.1    & 80.9     & 75.2        & 84.0 & 80.1        & 63.4       & 50.9  & 81.0    & 77.8     & 52.8     & 30.7       & 35.5      & 59.0 & 75.4    & 0.14      \\
LayerNorm      & 86.0     & 89.7       & 72.2 & 99.1      & 91.4 & 90.0   & 56.1   & 83.0 & 84.7     & 93.8    & 83.0       & 75.2        & 84.2 & 77.5        & 62.2       & 49.9  & 78.1  & 78.0       & 52.1     & 24.3       & 34.4      & 57.1 & 74.7    & 0.04      \\
SSF            & 86.6     & 89.8       & 68.8 & 99.1      & 91.4 & 91.2 & 56.5   & 82.8 & 86.1     & 94.5    & 83.2     & 74.8        & 84.7 & 80.1        & 63.6       & 53.0    & 81.4  & 85.6     & 52.1     & 31.9       & 37.2      & 60.6 & 76.0    & 0.21      \\
\midrule
Pfeif. Adapter & 86.3       & 91.5       & 72.1 & 99.2      & 91.4 & 88.5 & 55.7   & 83.0 & 86.2     & 95.5    & 85.3     & 76.2        & 85.8 & 83.1        & 65.2       & 51.4  & 80.2  & 83.3     & 56.6     & 33.8       & 41.1      & 61.8 & 76.9    & 0.67       \\
Houl. Adapter  & 84.3     & 92.1       & 72.3 & 98        & 91.7 & 90.0   & 55.4   & 83.2 & 88.7     & 95.3    & 86.5     & 75.2        & 86.4 & 82.9        & 63.6       & 53.8  & 79.6  & 84.4     & 54.3     & 34.2       & 44.3      & 62.1 & 77.2    & 0.77      \\
AdaptFormer    & 85.8     & 91.8       & 70.5 & 99.2      & 91.8 & 89.4 & 56.7   & 83.2 & 86.8     & 95.0      & 86.5     & 76.3        & 86.2 & 82.9        & 64.1       & 52.8  & 80.0    & 84.7     & 53.0       & 33.0         & 41.4      & 61.5 & 76.9    & 0.46      \\
RepAdapter     & 86.0     & 92.5       & 69.1 & 99.1      & 90.9 & 90.9 & 55.4   & 82.9 & 86.9     & 95.3    & 86.0       & 75.4        & 85.9 & 82.5        & 63.5       & 51.4  & 80.2  & 85.4     & 52.1     & 35.7       & 41.7      & 61.6 & 76.8    & 0.53      \\
Convpass       & 85.0     & 92.1       & 72.0   & 99.3      & 91.3 & 90.8 & 55.9   & 83.5 & 87.7     & 95.8    & 85.9     & 75.9        & 86.3 & 82.3        & 65.2       & 53.8  & 78.1  & 86.5     & 55.3     & 38.6       & 45.1      & 63.1 & 77.6    & 0.49      \\
\midrule
LoRA           & 85.7     & 92.6       & 69.8 & 99.1      & 90.5 & 88.5 & 55.5   & 82.6 & 87.5     & 94.9    & 85.9     & 75.7        & 86.0 & 82.9        & 63.9       & 51.8  & 79.9  & 86.6     & 47.2     & 33.4       & 42.5      & 61.0 & 76.5    & 0.55      \\
FacT\_TT       & 85.8     & 91.8       & 71.5 & 99.3      & 91.1 & 90.8 & 55.9   & 83.4 & 87.7     & 94.9    & 85.0       & 75.6        & 85.8 & 83.0          & 64.0         & 49.0    & 79.3  & 85.8     & 53.1     & 32.8       & 43.7      & 61.3 & 76.8    & 0.13      \\
FacT\_TK       & 86.2     & 92.5       & 71.8 & 99.1      & 90.1 & 91.2 & 56.2   & 83.4 & 85.8     & 95.5    & 86.0       & 75.7        & 85.8 & 82.7        & 65.1       & 51.5  & 78.9  & 86.7     & 53.1     & 27.8       & 40.8      & 60.8 & 76.6    & 0.23   \\  
\midrule
Relative Std Dev& 0.81 & 1.13 & 1.78 & 0.34 & 0.54 & 1.82 & 1.24 & 0.54 & 1.20 & 0.59 & 1.95 & 0.83 & 0.94 & 2.67 & 1.50 & 3.22 & 1.37 & 4.11 & 4.46 & 11.02 & 9.30 & 2.70 & 1.09 & - \\
\bottomrule
\end{tabular}}
\vskip-7pt
\caption{\small PEFT methods performances on VTAB-1K (19 tasks from 3 groups) by \method{Top-1 Accuracy}.  Based on the results across PEFT, linear probing, and full FT, we identify two task groups ({\color{blue!30}{purple}} and {\color{orange!30}{orange}}), as discussed in \autoref{sec: why}.}
\label{tab: vtab}
\vskip-7pt
\end{table*}

\subsection{Related work and comparison}
\label{sec: related}
The community-wide enthusiasm for PEFT has led to multiple survey articles~\citep{yu2023visual, xin2024parameter, han2024parameter}. Meanwhile, several empirical and theoretical studies were presented, mostly based on NLP tasks, attempting to provide a holistic understanding. \citep{he2021towards,mao2021unipelt} provided unified
views to methodologically connect PEFT methods. \citep{chen2022revisiting,ding2023parameter,he2021effectiveness} and \citep{he2022parameter, xin2024v} empirically compared PEFT methods on NLP and vision tasks, respectively, while \citep{fu2023effectiveness} offered a theoretical stability and generalization analysis. Accuracy-wise, \citep{chen2022revisiting,ding2023parameter,he2021effectiveness} found that PEFT is robust to over-fitting and quite effective in NLP tasks under low-data regimes. \emph{This is, however, not the case for vision tasks: \citep{he2022parameter} showed that representative PEFT methods like LoRA and Adapter can't consistently outperform either full FT or linear probing.}
In terms of why PEFT works, \citep{fu2023effectiveness} framed PEFT as sparse fine-tuning and showed that it imposes a regularization by controlling stability; \citep{ding2023parameter,he2022parameter} framed PEFT as (subspace) optimization; \citep{ding2023parameter} further discussed the theoretical principle inspired by optimal control.

Our study strengthens and complements the above studies and offers new insights. First, we compared over \textit{ten} PEFT methods and carefully tuned the hyperparameters, aiming to accurately assess each method’s performance. This is particularly crucial for the vision community, where comprehensive references are still limited, and simpler methods like BitFit have often been deemed inferior, while other approaches have shown discrepancies compared to NLP studies.  
Second, we go beyond a \emph{competition} perspective to investigate a \emph{complementary} perspective of PEFT approaches. We show that different PEFT approaches offer \textit{effective base learners for model ensembles}. Third, we go beyond downstream accuracy to investigate PEFT's effectiveness in maintaining \textit{out-of-distribution robustness}. Finally, we analyze both \textit{low-shot} and \textit{many-shot} settings, revealing distinct patterns among PEFT, full FT, and linear probing, extending the understanding of PEFT. 

\section{PEFT Methods in Low-Shots Regime}
\label{sec:few}




\begin{figure*}[htbp]
    \centering
    \small
    \begin{tabular}{@{}c@{\hspace{0.01\linewidth}}c@{\hspace{0.01\linewidth}}c@{}}
        \includegraphics[width=0.3\linewidth]{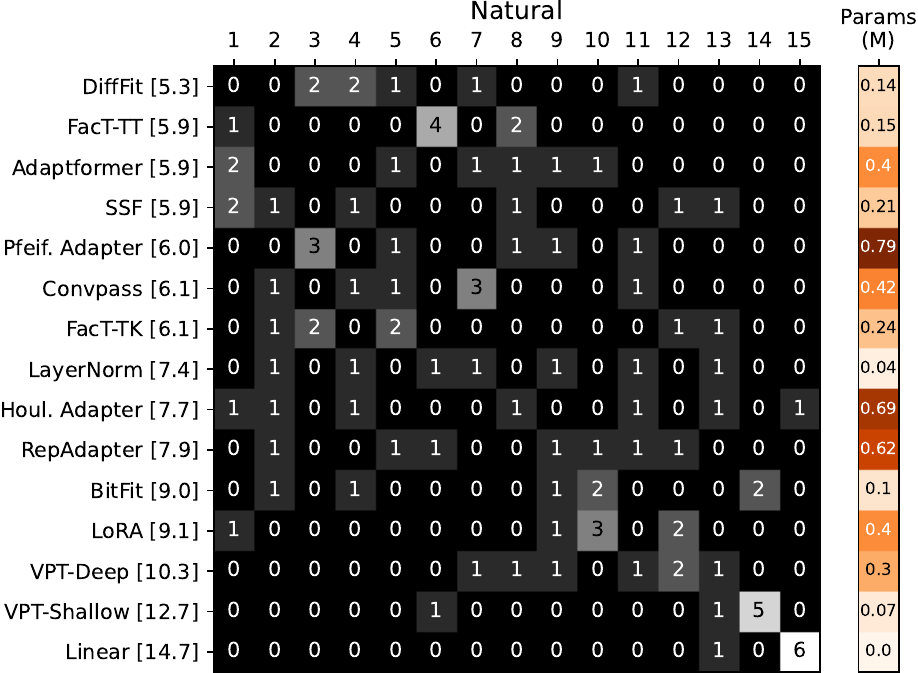} &
        \includegraphics[width=0.3\linewidth]{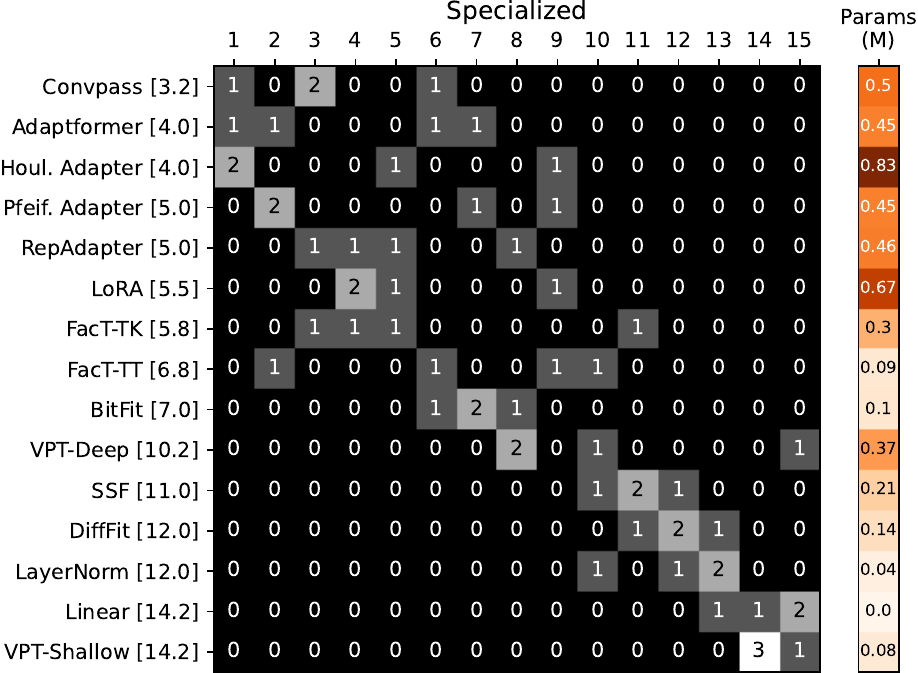} &
        \includegraphics[width=0.3\linewidth]{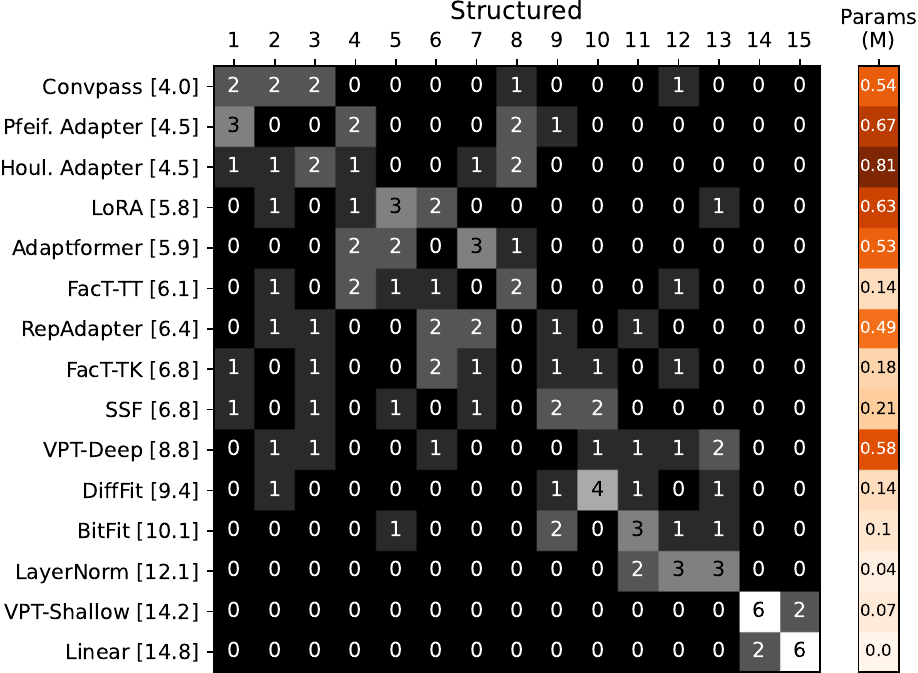} \\
    \end{tabular}
    \vskip-11pt
    \caption{\small Ranking frequency of 15 methods (14 PEFT + linear probing) for three groups in VTAB-1K. Element $(i, j)$ is the number of times method $i$ ranks $j^{th}$ in each group. Methods are ordered by mean ranks (in brackets). The parameters column shows the \# of trainable parameters in millions. More details are in \autoref{-sec:results}.}
    \label{fig:ranking}
    \vskip-10pt
\end{figure*}


Pre-trained models are meant to ease downstream applications. One representative scenario is low-shot learning: supervised FT of the pre-trained model with a small number of samples per class. Indeed, low-shot learning has been widely used to evaluate PEFT performance. 

 \noindent\textbf{Dataset.} \textbf{VTAB-1K}~\citep{zhai2019large} consists of 19 classification tasks from three groups.  \textbf{Natural} comprises natural images captured with standard cameras.  \textbf{Specialized} contains images captured by specialist equipment for remote sensing and medical purposes. \textbf{Structured} evaluates scene structure comprehension, including object counting and depth estimation. Following~\cite{zhai2019large}, we split the \textbf{1000} training image 80/20 for hyperparameter tuning. The reported \method{Top-1 Accuracy} is obtained after training over the 1000 images and evaluating on the original test set. 

 \noindent\textbf{Methods.} We consider linear probing, full FT, and \textbf{14} PEFT methods: {2} {prompt-based}~\citep{jia2022visual}, {5} {adapter-based}~\citep{houlsby2019parameter,pfeiffer2021adapterfusion,chen2022adaptformer,convpass,luo2023rep}, {4} {direct selective}~\citep{zaken2022bitfit,ln-tune,xie2023difffit,ssf}, and {3} {efficient selective}~\citep{hu2021lora,jie2023fact}. Please refer to~\autoref{sec:PEFT-all} and~\autoref{-sec: survey} for details.

 
 \noindent\textbf{Setup.} We employ the ViT-B/16~\citep{dosovitskiy2020image} pre-trained on ImageNet-21K~\citep{deng2009imagenet} as the backbone. The prediction head is randomly initialized for each dataset.  We systematically tune 1) learning rate, 2) weight decay, and 3) method-specifics like the PEFT parameter sizes, which are often left intact in previous studies. We set a cap for PEFT size $\leq1.5\%$ of ViT-B/16. We also turn the drop path rate~\citep{huang2016deep} on (0.1) or off (0). A detailed experiment setup is provided in \autoref{-sec:setup}, and more experiment results, including DINOv2~\cite{oquab2024dinov2} and larger backbones (ViT-L and ViT-H), are provided in \autoref{-sec:results}.



 \noindent\textbf{Results.}  As shown in \autoref{fig:vtab_vis} and \autoref{tab: vtab}, PEFT methods generally outperform both linear probing and full FT across datasets. Additionally, with proper implementation and fair hyperparameter tuning, we surprisingly found that most PEFT methods perform similarly as the relative standard deviations (divided by the means) in all three groups are quite low. Simple methods (\eg, Bitfit) and PEFT methods originally proposed for NLP (\eg, LoRA and Adapter), which were previously reported as inferior due to unoptimized implementations and hyperparameter tuning, now demonstrate competitive performance with SOTA visual PEFT methods.  To understand the relative advantages of different approaches, we provide the ranking frequency of PEFT methods across different groups in \autoref{fig:ranking}, where the element $(i, j)$ in each ranking matrix represents the frequency that method $i$ ranks $j^{th}$ in each group. Methods are ordered by their mean ranks (in brackets), and the parameters column indicates the number of trainable parameters in millions. In natural group, simpler methods with fewer trainable parameters---such as DiffFit and Fact-TT---offer a cost-effective solution without compromising performance. Conversely, in specialized and structured groups, methods with more parameters generally yield better performance. We hypothesize that this performance discrepancy arises from the \textbf{domain similarity} between the pre-trained domain (ImageNet) and the downstream domains. The natural group, sharing a stronger affinity with ImageNet, allows simpler methods like BitFit to adjust the features effectively. In contrast, the specialized and structured groups necessitate more complex methods with more trainable parameters to bridge the domain gap.

\noindent\textbf{Recipes.} 
In low-shot regimes, when the downstream data are similar to the pre-trained data, simple methods (\eg, DiffFit) offer solid accuracy with fewer parameters. Conversely, if there is a substantial domain gap, more complex methods with more parameters often achieve higher accuracy. Since low-shot training is especially prone to over-fitting, we recommend activating a nonzero drop path rate—commonly set to zero by default—that stochastically drops a Transformer block per sample \citep{huang2016deep}. \textit{All} methods can benefit from such a randomization-based regularization, as shown in \autoref{fig:dpr} in the Appendix.

\section{Different PEFT Approaches Offer Complementary Information}
\label{sec:complememtary}

The previous section demonstrates that all PEFT methods perform similarly across various domains. Given that different PEFT methods are trained on the same downstream data using the same backbone and achieve comparable accuracy, one might expect them to learn similar knowledge from the data, resulting in similar predictions. Contrary to this expectation, our findings below reveal that different PEFT methods acquire \textbf{distinct} and \textbf{complementary} knowledge from the \textit{same} downstream data, even when built upon the \textit{same} backbone, leading to \textbf{diverse} predictions.
\begin{figure*}[htbp]
\centering
\begin{tabular}{@{}c@{\hspace{0.1mm}}c@{}}
\begin{subfigure}[t]{0.75\textwidth}
    \centering
    \includegraphics[width=0.32\textwidth]{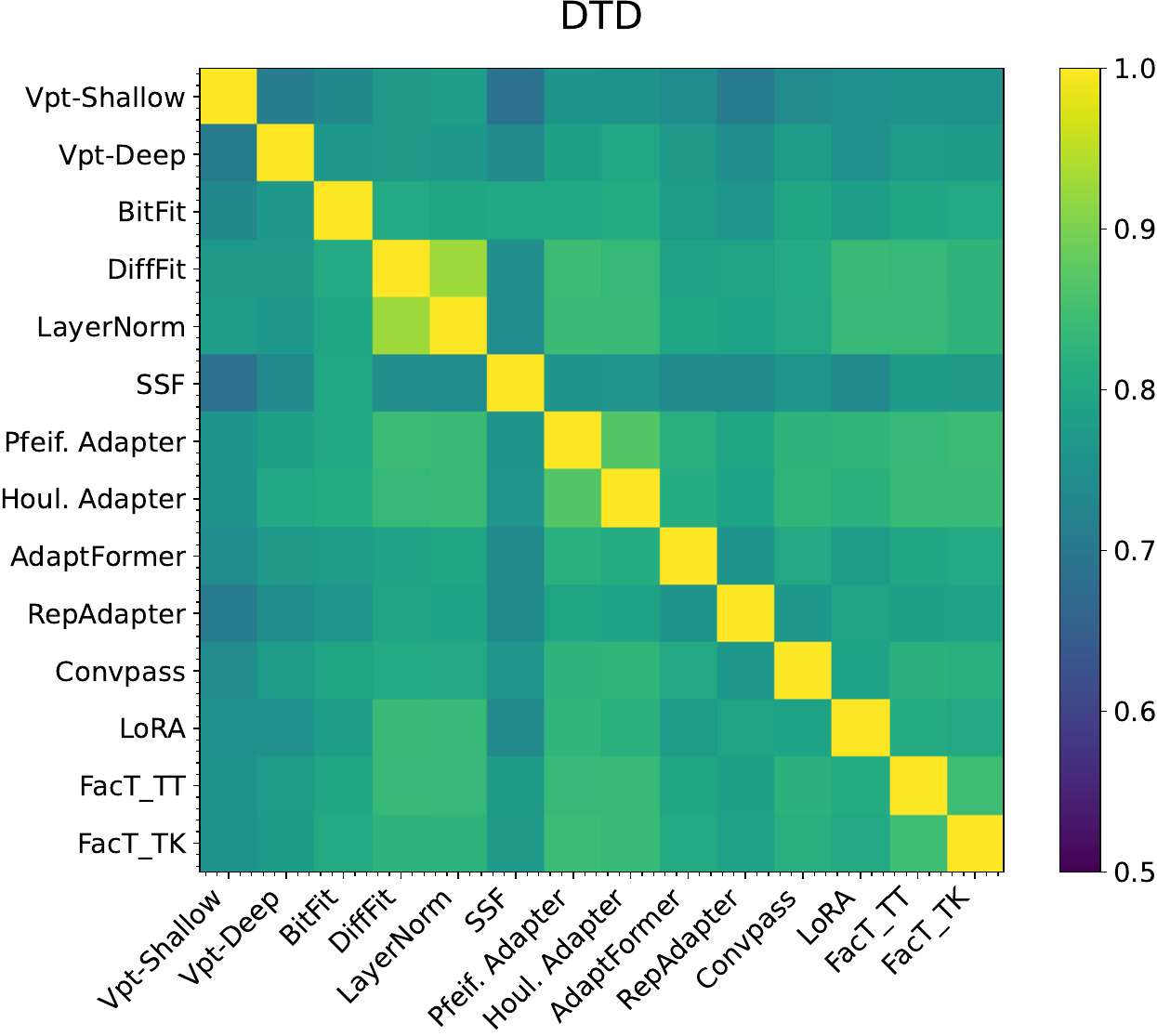}
    \includegraphics[width=0.32\textwidth]{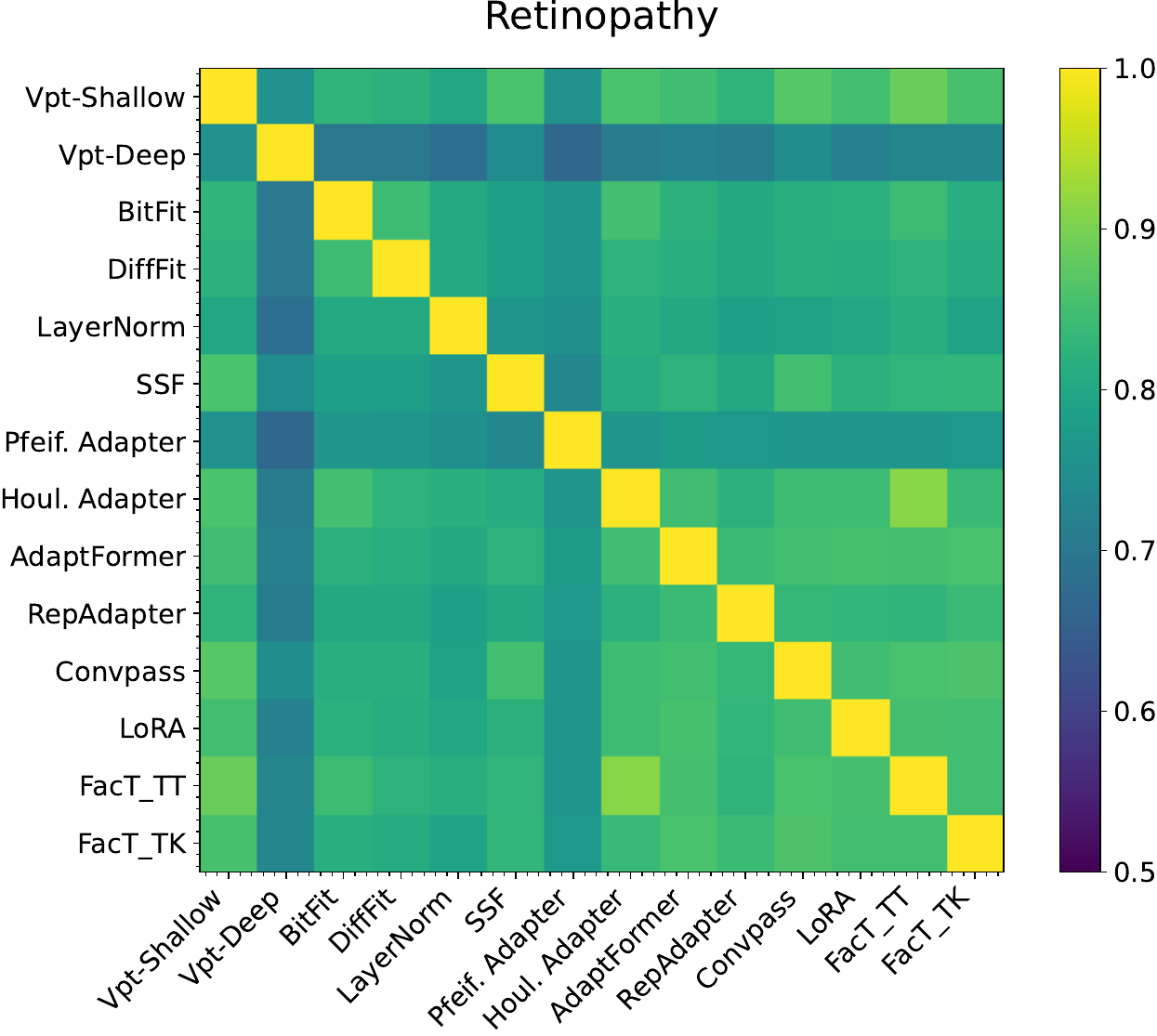}
    \includegraphics[width=0.32\textwidth]{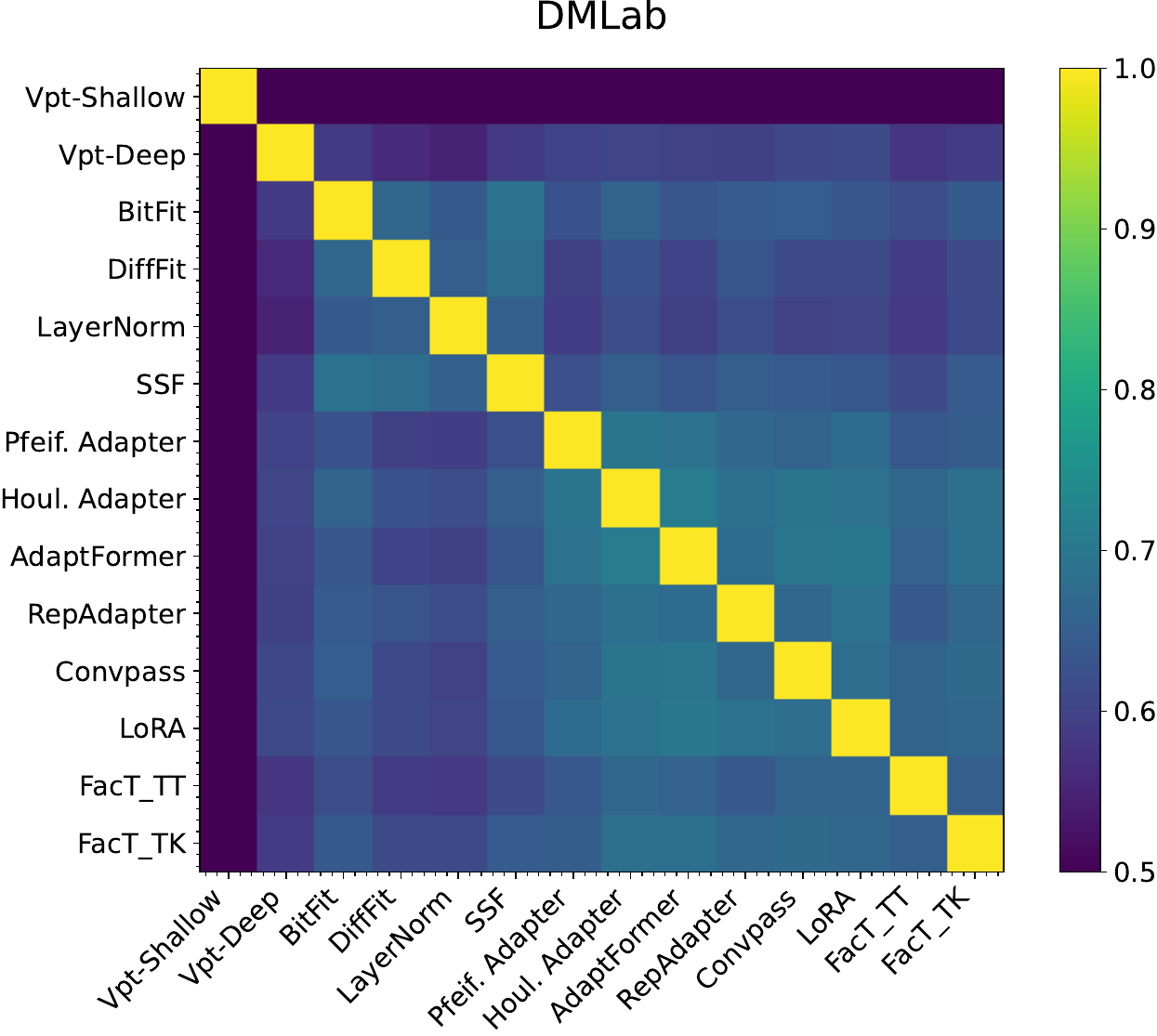}
    \caption{}
    \label{fig:diversit_pred_part1}
\end{subfigure}
&
\begin{subfigure}[t]{0.22\textwidth}
    \centering
    \includegraphics[width=\textwidth]{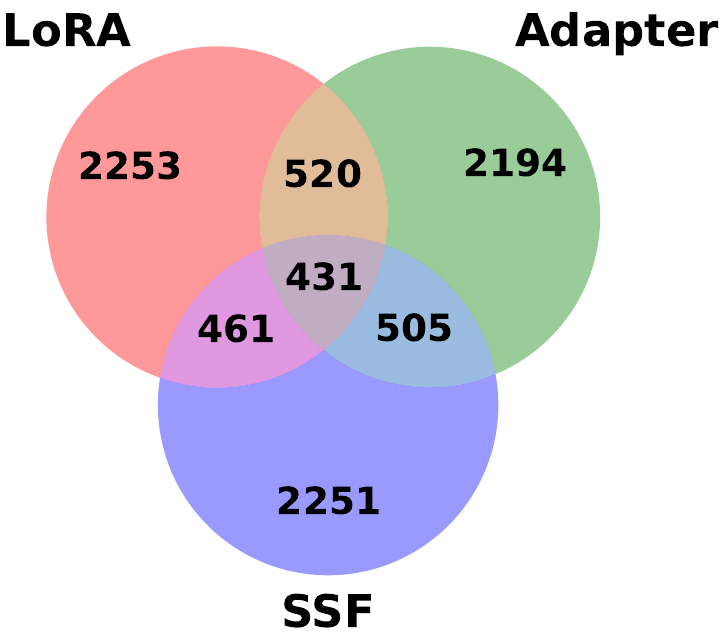}
    \caption{}
    \label{fig:diversit_pred_part2}
\end{subfigure}
\end{tabular}
\vskip-10pt
\caption{%
  \small \small (a) Prediction similarity analysis: element $(i,j)$ shows
      the percentage of samples that method $i$ and $j$ predict the same. 
      Although different methods achieve similar accuracy, they have diverse 
      predictions. (b)The wrong prediction overlaps of LoRA, Adapter, and SSF for the 5K least confident data. Correct prediction overlaps for the 5K most confident data are shown in~\autoref{fig:highconf}. They are FT on CIFAR100 (VTAB-1K). More results for (a) and (b) are in \autoref{-sec:results}.
}
\label{fig:my_figure}
\vskip-10pt
\end{figure*}

\begin{figure}[htbp]
\centering
\includegraphics[width=0.9\linewidth]{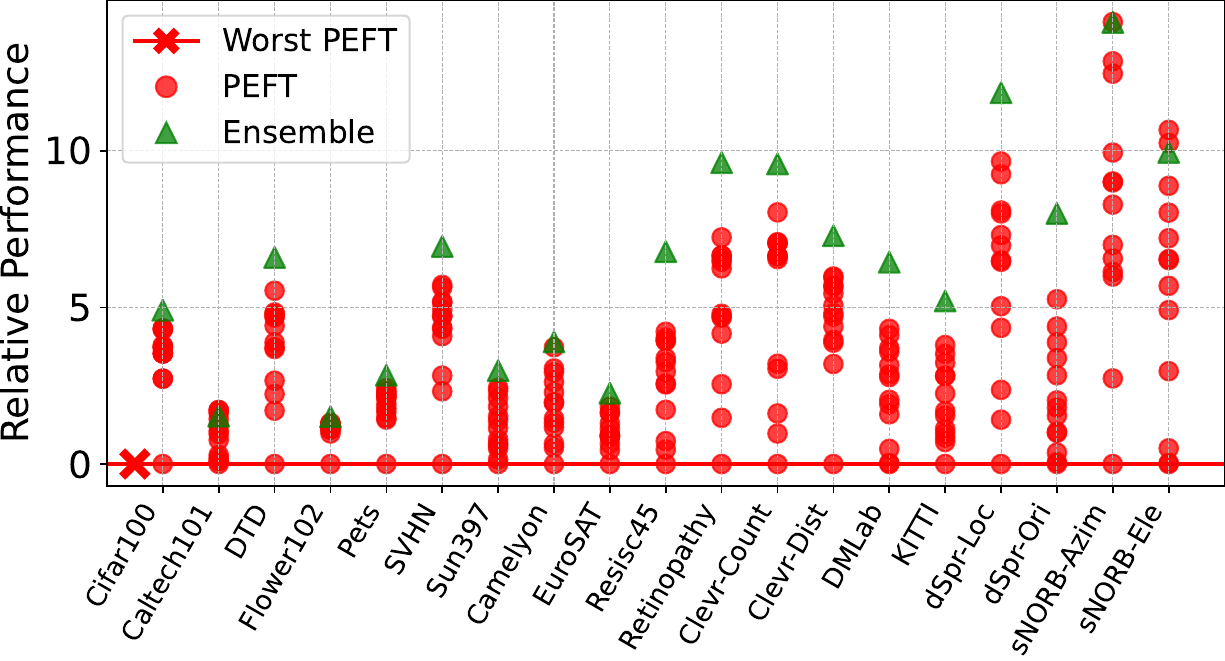}
\vskip-10pt
\caption{\small Ensemble (majority vote) shows consistent gain on most datasets thanks to the diverse predictions. }
  \label{fig:ensemble}
\vskip-10pt
\end{figure}

We start by analyzing their prediction similarity on the same dataset in VTAB-1K. It is expected that their predictions are similar for datasets with very high accuracy, such as Flowers102 (avg 99.1\%) and Caltech101 (avg 91.4\%). Beyond them, we find that most PEFT methods show diverse predictions in other datasets in VTAB-1K. \autoref{fig:diversit_pred_part1} shows the prediction similarities between 14 PEFT methods in DTD, Retinopathy, and DMLab, which belong to natural, specialized, and structured groups, respectively. In DTD and Retinopathy, most methods differ in about 20\% of their predictions, while in DMLab, this difference increases to approximately 35\%, even though they achieve similar accuracies. This prediction diversity may be attributed to the different \textit{inductive biases}~\citep{neyshabur2014search} of PEFT methods --- they explicitly select specific parameters to update or insert different modules at various locations within the model. More analyses and details are offered in \autoref{-sec:results}.

Such diverse predictions across methods open up the possibility of leveraging their heterogeneity for further improvement. The most straightforward approach is ensemble~\citep{gontijo2021no}, \eg, majority vote over methods. \autoref{fig:ensemble} demonstrates the ensemble performance gain over all the PEFT methods in each dataset, where we use the worst PEFT method as the baseline. Thanks to the diverse predictions across methods, the ensemble can provide consistent gain. 



Also, we analyze if PEFT methods make similar correct predictions for high-confidence samples and similar mistakes for low-confidence samples. 
\autoref{fig:highconf} and  \autoref{fig:diversit_pred_part2} show the correct prediction overlap for the 5K most confident samples (per method) and the wrong prediction overlap for the 5K least confident samples (per method). For demonstration purposes, we select one method from each PEFT category (LoRA, Adapter, SSF) and they are FT on CIFAR-100 in VTAB-1K. Methods within the same category also show diverse predictions (\autoref{-sec:results}). Since they make different predictions in both high and low-confidence regimes, this paves the way for new possibilities of using different PEFT methods to generate \textbf{diverse pseudo-labels} for semi-supervised learning~\citep{yang2022survey, gan2024erasing, gansemi},  domain adaptation~\citep{farahani2021brief, englert2024exploring, tang2024source}, and continual learning~\cite{mai2022online, mai2021supervised, shim2021online, lomonaco2022cvpr, mai2020batch}. For example, in semi-supervised learning, we can {FT different PEFT methods on the labeled data to generate \textit{diverse and accurate pseudo-labels} by selecting highly confident predictions from each PEFT method}. 

\section{PEFT Methods in Many-Shot Regime}

\label{sec:many}
Recent works in NLP~\cite{chen2022revisiting} have indicated that PEFT methods may not perform as competitively as full FT when data is abundant. We thus aim to investigate PEFT's performance in many-shot regimes by addressing the following questions: (1) \textit{Should we use PEFT or full FT when data is sufficient?} (2) \textit{How should we adjust the number of trainable parameters for PEFT methods in many-shot regimes? }

\noindent\textbf{Dataset.} We select one representative dataset from each group in VTAB: (1) CIFAR-100~\cite{krizhevsky2009learning}, a natural image dataset comprising 50K training images across 100 classes; (2) RESISC~\cite{cheng2017remote}, a remote sensing dataset for scene classification with 25.2K training samples across 45 classes; and (3) Clevr-Distance~\cite{zhai2019large, johnson2017clevr}, a synthetic image dataset for predicting the depth of the closest object with 6 depth classes and 70K samples. The reported results are obtained by training on the \textbf{full}  training set and evaluating on the original test set. 


 \noindent\textbf{Setup.} The model setup mostly follows the VTAB-1K experiment.   More details about setup and hyperparameter search are provided in \autoref{-sec:setup}.

\begin{figure*}[t]
    \centering
    \small
    \begin{tabular}{@{}c@{\hspace{0.01\linewidth}}c@{\hspace{0.01\linewidth}}c@{}}
        \includegraphics[width=0.33\linewidth]{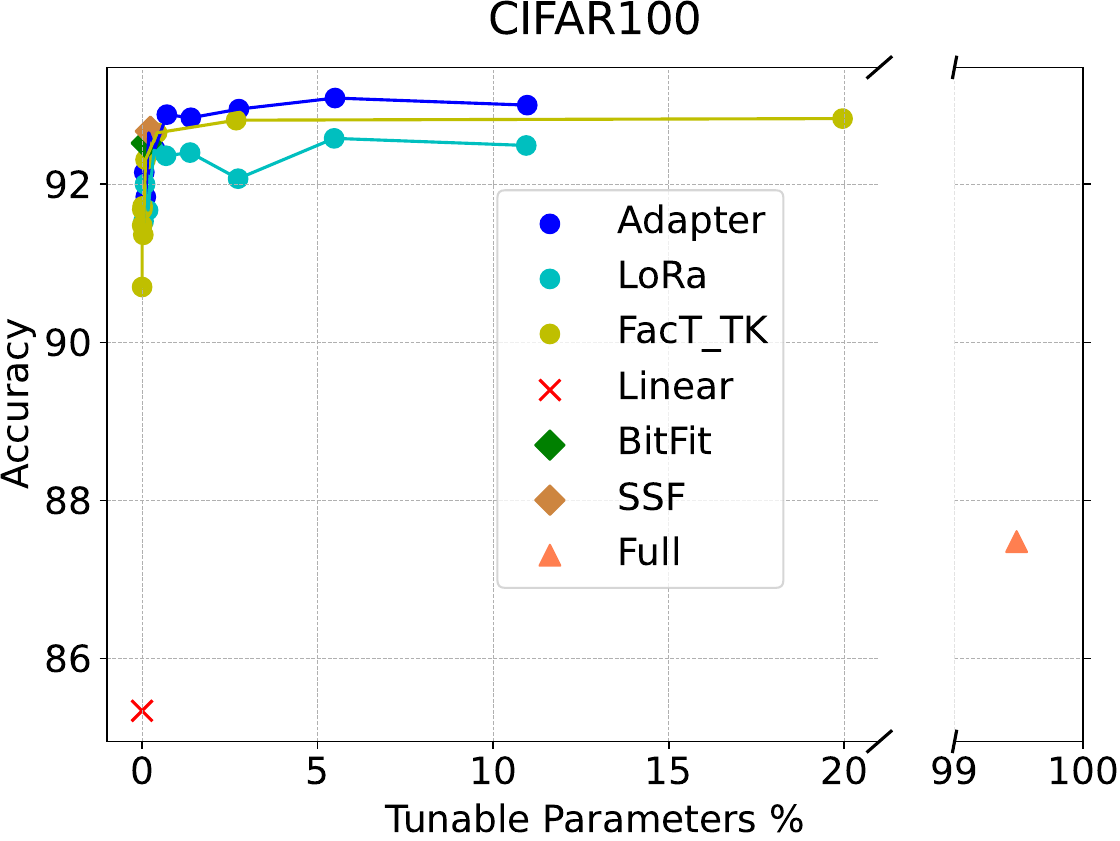} &
        \includegraphics[width=0.33\linewidth]{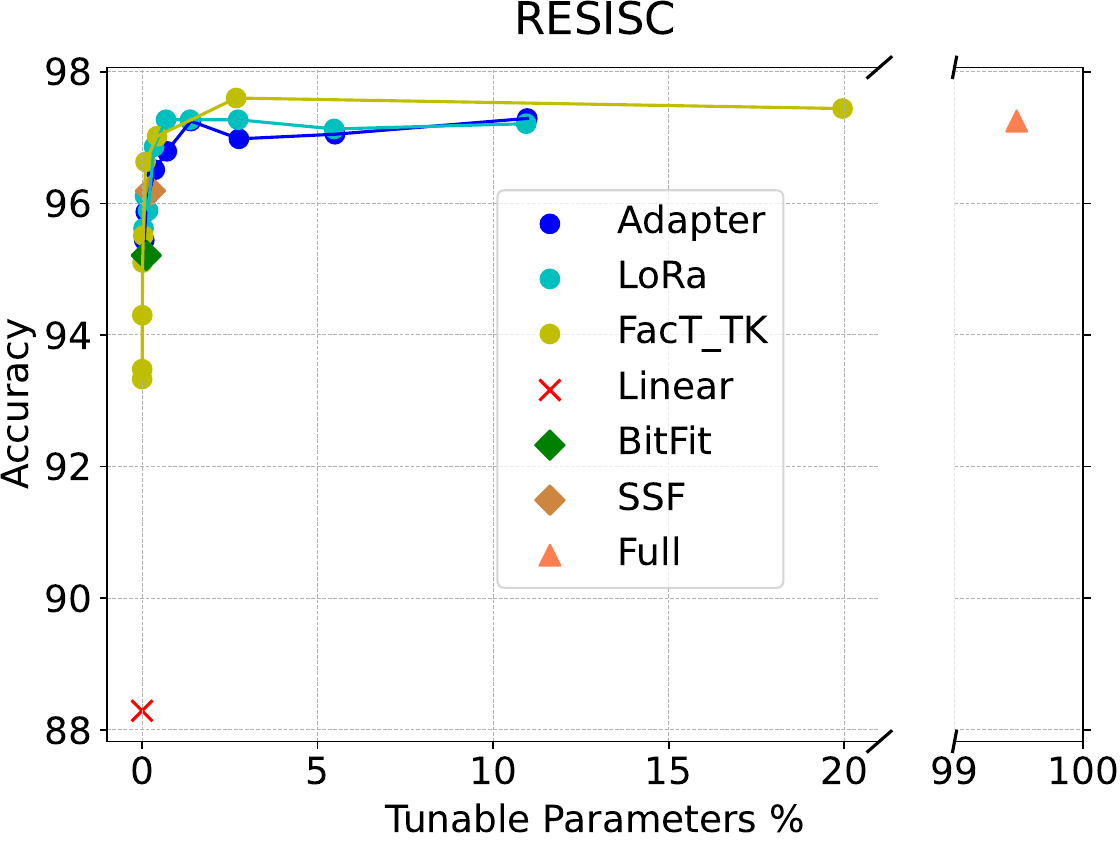} &
        \includegraphics[width=0.33\linewidth]{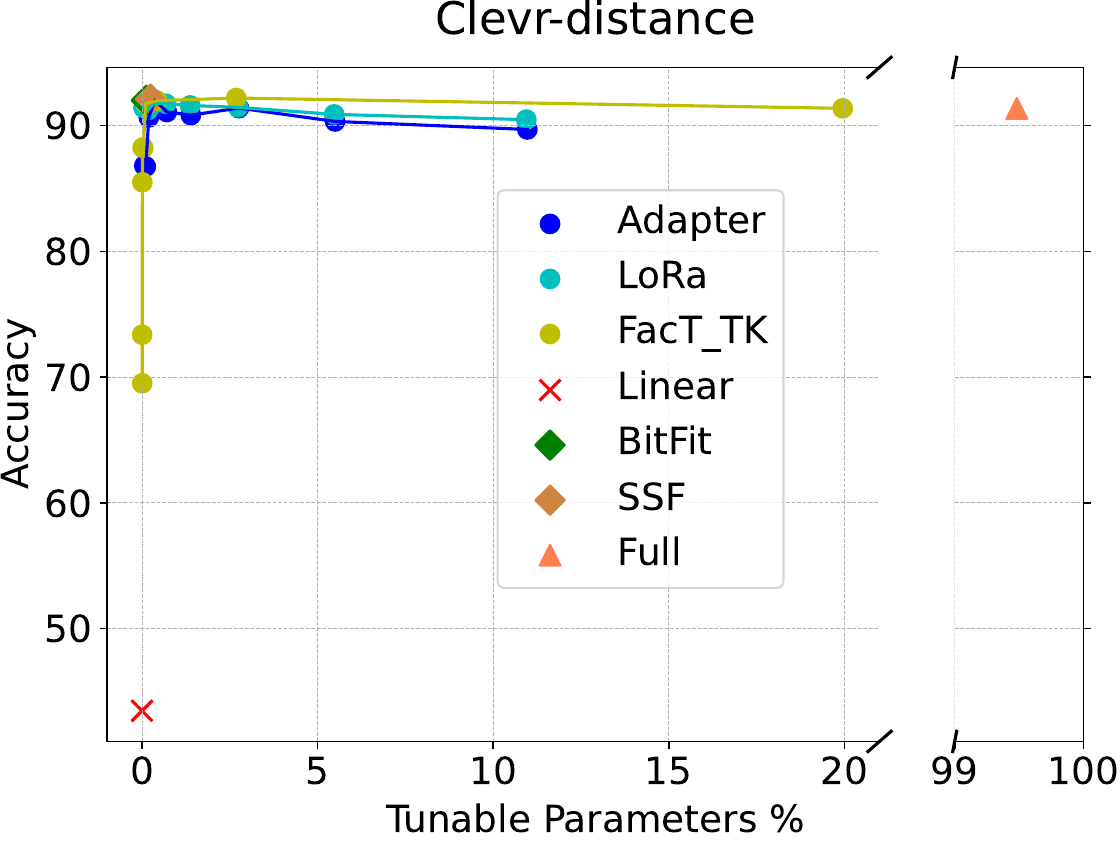} \\
    \end{tabular}
    \vskip -7pt
    \caption{\small PEFT accuracy in many-shot regimes, with different parameter sizes (X-axis) on three datasets from different domains. Even 2\%-5\% trainable parameters allow the models to have sufficient capacity to learn from full data. Details are in \autoref{-sec:results}). }
    \label{fig:many_shots}
    \vskip -10pt    
\end{figure*}

\noindent\textbf{Results.} In many-shot regimes, with sufficient downstream data, full FT may catch up and eventually outperform PEFT methods. However, from \autoref{fig:many_shots}, we found that even in many-shot regimes, PEFT can achieve \textbf{comparable results with full FT}, even just using $2\%$ of fine-tuning parameters. The performance gain, however, quickly \textbf{diminishes and plateaus after 5\%} of tunable parameters. By comparing the results on the domain-close CIFAR-100 and domain-different RESISC and Clevr, we have some further observations. Downstream tasks with larger domain gaps often require updating more parameters to achieve high accuracy. With sufficient downstream data, full FT is less prone to over-fitting and, indeed, attains a high accuracy. But interestingly, PEFT methods, with only \textbf{2\%$\sim$5\%} of tunable parameters, achieve similar accuracy, suggesting that its design principle does offer \textbf{sufficient effective capacity} for the model to learn~\cite{zhang2021understanding}. Downstream tasks with smaller domain gaps suggest that the pre-trained model had learned sufficient knowledge about them; full FT  thus risks washing such knowledge away. In fact, we found that PEFT notably outperforms full FT on CIFAR-100, suggesting it as a more \emph{robust transfer learning} algorithm for downstream tasks.

\noindent\textbf{Recipes.} 
In many-shot regimes, PEFT methods with sufficient parameters (2$\sim$5\%) appear more favorable than full FT and linear probing. On the one hand, they achieve comparable and even better accuracy than full FT. On the other hand, the tunable parameters remain manageable. {The parameter efficiency of PEFT also often implies \textit{less training GPU memory usage and training time}, making PEFT methods a favorable alternative in many-shot regimes.}  For a downstream domain that is close to the pre-training domain, PEFT shows much pronounced \emph{transferability}. For a downstream domain that is quite different, the limited tunable parameters (controversially, 2$\sim$5\% already amount to a few million) already allow the model to learn sufficiently.

\section{Why Do PEFT Methods Work?\footnotemark}

\label{sec: why}
\footnotetext[\value{footnote}]{\noindent Our intention is not to offer a definitive explanation about why PEFT works. As discussed in \autoref{sec: related}, there is no broadly accepted theory for PEFT's success. We hope our empirical findings can support the ongoing efforts to uncover the fundamental principles behind PEFT.}

\begin{figure}[htbp]
    \centering
    \begin{tabular}{@{}cc@{}}
    \includegraphics[width=0.5\columnwidth]{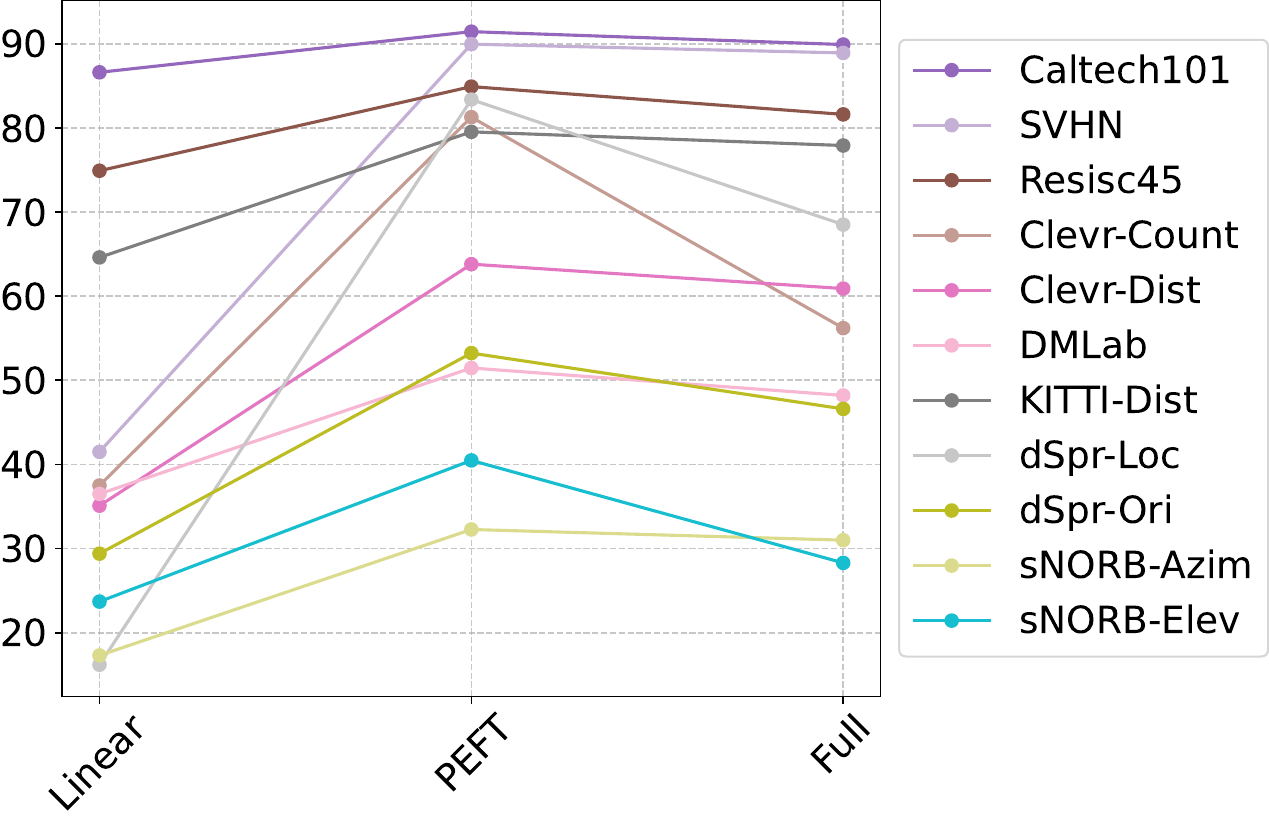} &
    \includegraphics[width=0.5\columnwidth]{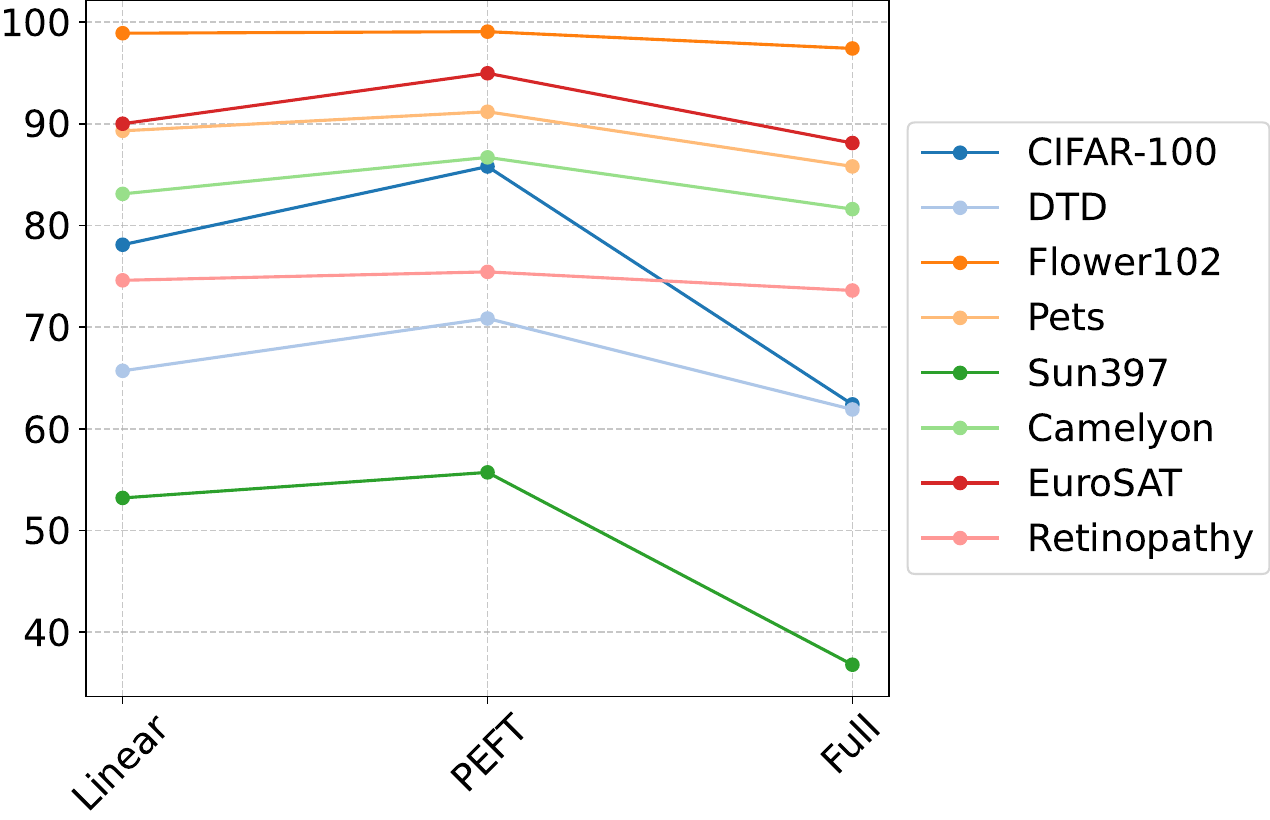} \\
    \parbox{0.48\columnwidth}{\centering (a)} &
    \parbox{0.48\columnwidth}{\centering (b)} \\
    \end{tabular}
    
    \vspace{0.1cm} 
    
    \begin{tabular}{@{}cc@{}}
        \includegraphics[width=0.5\columnwidth]{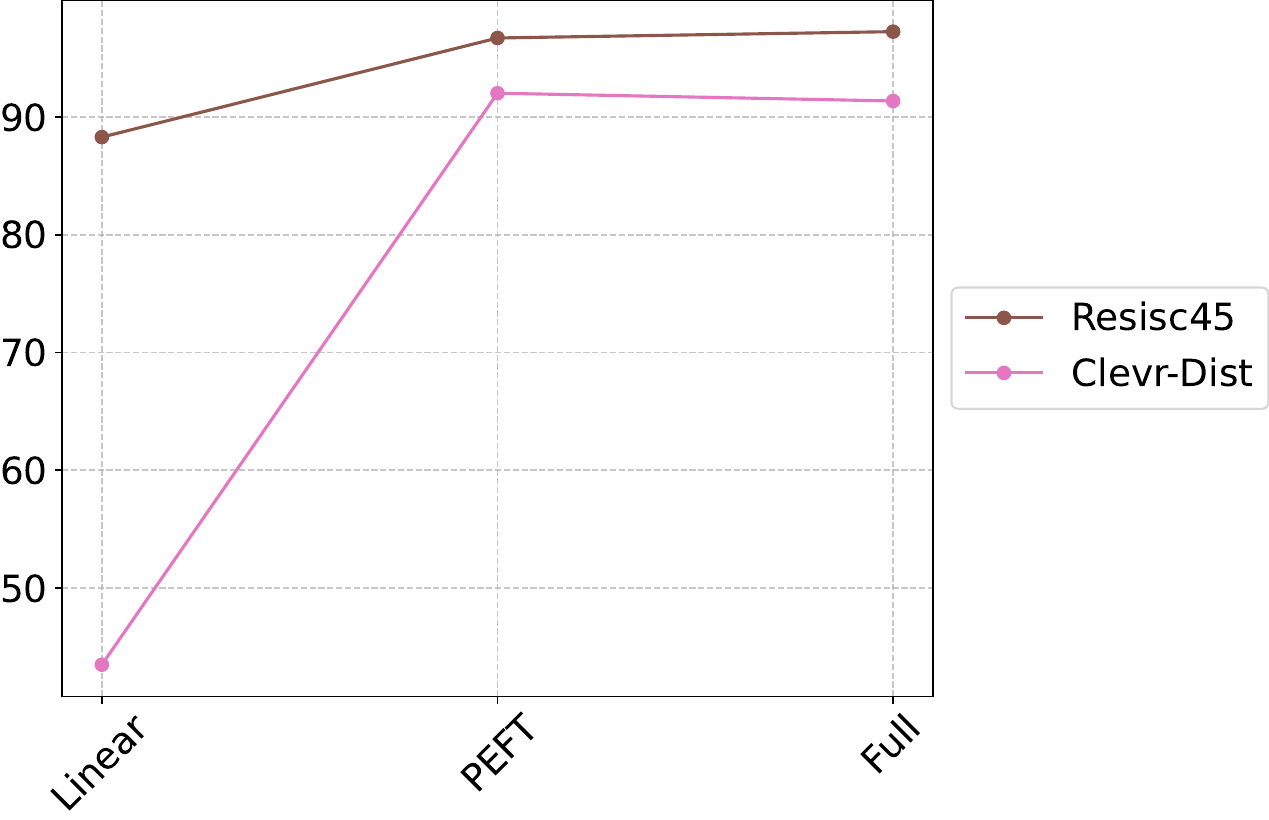} &
        \includegraphics[width=0.5\columnwidth]{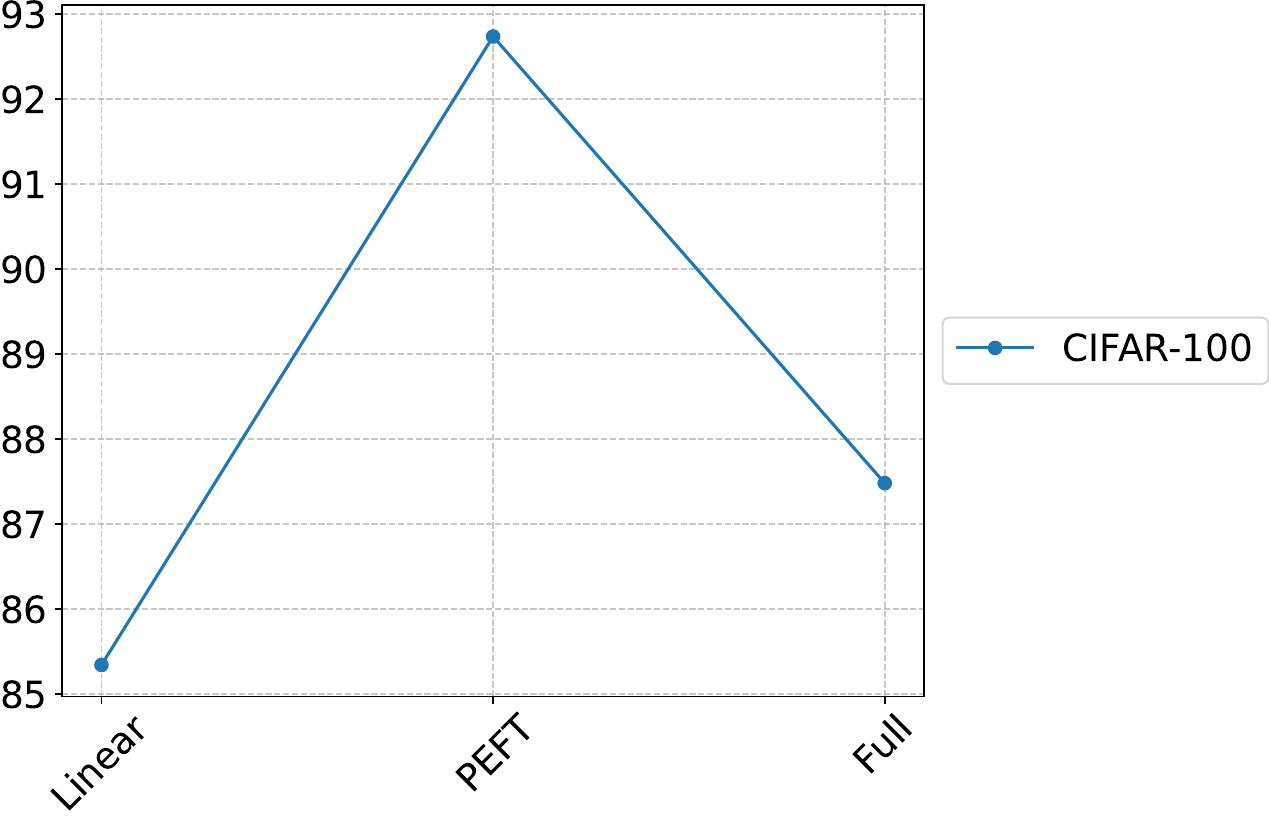} \\
        \parbox{0.48\columnwidth}{\centering (c)} &
        \parbox{0.48\columnwidth}{\centering (d)} \\
    \end{tabular}

    \vskip -7pt
    \caption{\small (a): VTAB-1K tasks in case \colorbox{blue!30}{\makebox(5,5){(1)}}, PEFT $>$ full $>$ linear. (b) VTAB-1K tasks in case \colorbox{orange!30}{\makebox(5,5){(2)}}, PEFT $>$ linear $>$ full. (c) RESISC \& Clevr in case \colorbox{blue!30}{\makebox(5,5){(1)}} with enough data, PEFT $\approx$ full $>$ linear. (d) CIFAR in case \colorbox{orange!30}{\makebox(5,5){(2)}} with enough data, PEFT $>$ full $>$ linear. Within each figure, left for linear, middle for PEFT, and right for full. More details are in \autoref{-sec:results}.}
    \label{fig:why}
    \vskip -7pt    
\end{figure}

Putting together~\autoref{sec:few} and~\autoref{sec:many}, we identify two distinct patterns regarding the performance among linear probing, full FT, and PEFT. 
Within 19 VTAB-1K tasks, we see: \colorbox{blue!30}{(1)} \emph{Full FT outperforms linear probing.} As linear probing reflects the pre-trained feature quality for downstream tasks, case (1) suggests the necessity to update the backbone to close the gap between pre-trained and downstream domains. \colorbox{orange!30}{(2)} \emph{Linear probing surpasses full FT,} suggesting the pre-trained features are good enough (at least in a low-shot scenario). Recklessly updating them may risk over-fitting. 
\autoref{fig:why} (a-b) summarizes the low-shot accuracy comparison 
based on the categorization above; each line corresponds to one task. Linear probing, PEFT, and full FT are located in order, from left to right, to reflect their tunable parameter sizes. PEFT's superiority in both cases showcases its \textbf{capacity} to learn and its \textbf{regularization role} to prevent over-fitting. 

We also draw the many-shot accuracy in \autoref{fig:why} (c-d) based on the same categorization: RESISC and Clevr in case \colorbox{blue!30}{(1)}, and CIFAR-100 in case \colorbox{orange!30}{(2)}. 
In the many-shot setting, full FT consistently outperforms linear probing, which seems to suggest \emph{no more risk of over-fitting}. However, on CIFAR-100 (\autoref{fig:why} (d)), we again see a noticeable gap between PEFT and full FT, just like in \autoref{fig:why} (b). Such a concave shape reminds us of the long-standing under-fitting-over-fitting curve, suggesting that even with sufficient downstream data, full FT still risks over-fitting. 

Considering PEFT's comparable performance to full FT on RESISC and Clevr with large domain gaps, we conclude that PEFT succeeds as a \textbf{high-capacity} learner equipped with an \textbf{effective regularizer}. The two roles trade-off well such that PEFT can excel in both low- and high-similarity domains under both low-shot and many-shot settings.

\begin{table*}
\centering
\resizebox{0.98\textwidth}{!}{%
\begin{tabular}{|c|c|c|c|c|c|c|c|c|c|}
\hline
         & Full  & BitFit & \begin{tabular}[c]{@{}c@{}}Layer-\\ Norm\end{tabular} & \begin{tabular}[c]{@{}c@{}}Houl. \\ Adapter\end{tabular} & \begin{tabular}[c]{@{}c@{}}Adapt-\\ Former\end{tabular} & \begin{tabular}[c]{@{}c@{}}Rep-\\ Adapter\end{tabular} & Convpass & LoRA  & FacT\_TK \\ \hline
100-shot ImageNet & 75.0 & 75.27  & 74.8                                                 & 75.0                                                    & 75.6                                                   & 76.5                                                  & 76.3    & 76.6 & 74.7    \\ \hline
Avg. distribution shift Acc     & 42.5 & 55.4 $\color{mydarkgreen}(12.9){\uparrow}$ & 55.9    $\color{mydarkgreen}(13.4){\uparrow}$                                             & 56.9      $\color{mydarkgreen}(14.4){\uparrow}$                                              & 56.1       $\color{mydarkgreen}(13.6){\uparrow}$                                            & 56.2        $\color{mydarkgreen}(13.7){\uparrow}$                                           & 54.7 $\color{mydarkgreen}(12.2){\uparrow}$    & 55.9 $\color{mydarkgreen}(13.4){\uparrow}$ & 56.1  $\color{mydarkgreen}(13.6){\uparrow}$   \\ \hline
\end{tabular}
}
\caption{\small The ``Avg. distribution shift Acc'' denotes the average performance of ImageNet-(V2, S, R, A) evaluated on the CLIP model FT on ImageNet. {\color{mydarkgreen}(${\uparrow}$) } indicates the gain over full FT. } 
\label{tab: robust}
\vskip -10pt 
\end{table*}

\section{How Robust are PEFT Methods to Distribution Shifts? }
\label{sec: robust}
Large pre-trained models such as CLIP~\cite{radford2021learning} have demonstrated unprecedented zero-shot accuracy across diverse data distributions. However, recent studies~\cite{wortsman2022robust, radford2021learning} have shown that FT on downstream data, while significantly boosting performance on the target distribution, often compromises the model's robustness to distribution shifts. Given that PEFT only updates a limited number of parameters in the model, we investigate whether PEFT can offer a more robust alternative to full FT.

 \noindent\textbf{Dataset.} We use 100-shot ImageNet-1K as our target distribution, with each class containing 100 images. Following~\cite{wortsman2022robust}, we consider 4 natural distribution shifts from ImageNet: \textbf{ImageNet-V2}~\cite{recht2019imagenet}, a new ImageNet test set collected with the original labeling protocol; \textbf{ImageNet-R}~\cite{hendrycks2021faces}, renditions for 200 ImageNet classes;  \textbf{ImageNet-S}~\cite{Gao_2023}, sketch images for 1K ImageNet classes; \textbf{ImageNet-A}~\cite{hendrycks2021natural}, a test set of natural images misclassified by a ImageNet pre-trained ResNet-50~\cite{he2015deep} for 200 ImageNet classes. 

\noindent\textbf{Setup.} We focus on the CLIP ViT-B/16 model, which comprises a visual encoder and a text encoder, pre-trained via contrastive learning on image-text pairs. Following~\cite{wortsman2022robust}, we add an FC layer as the head initialized using the class label text embedded by the text encoder. Subsequently, we discard the text encoder and apply PEFT methods to the visual encoder, FT only the PEFT modules and the head. More details about the CLIP model and experiment setup can be found in \autoref{-sec:setup}. 

\noindent\textbf{Results.} As shown in \autoref{tab: robust}, while some PEFT methods may not surpass full fine-tuning on the target distribution, they consistently demonstrate more robust performance on distribution-shifted data. This robustness can be attributed to PEFT updating only a small fraction of the parameters, thereby preserving the robust features of the foundational models.  Given the similar target distribution performance, \textit{should we blindly use PEFT methods for more robustness? }

\noindent\textbf{Weight-space ensembles (WiSE) for PEFT.} WiSE~\citep{wortsman2022robust}, which linearly interpolates the weights of a fully FT model with those of the original backbone, is a popular approach for boosting robustness. We explore whether WiSE can also enhance the robustness of PEFT. To apply WiSE to PEFT, we first linearly interpolate the prediction head with a mixing coefficient $\alpha$. For direct selective tuning methods (\eg BitFit), this involves merging the PEFT-tuned parameters and the original model. Since most Adapter-based methods include residual connections (Appendix~\ref{-sec: adapter}), we can adjust their impact by scaling the adapter modules with~$\alpha$. A similar approach applies to efficient selective methods (\eg LoRA) as they learn additive residuals to the original parameters. To the best of our knowledge, {we are the \textbf{first} to study WiSE for PEFT.}  As shown in \autoref{fig:merge} (more results in \autoref{-sec:results}), WiSE consistently improves both target and distribution shift performances of PEFT methods. For Adapter-based methods, WiSE can be considered feature ensembles, where $\alpha$ controls how strong the domain-specific features from the adapter module blend with the domain-agnostic ones from original backbones. For selective tuning, WiSE functions similarly to its application in full FT—exploiting the fact that the fine-tuned parameters remain near the original loss basin, allowing an ensemble that captures the best of both~\citep{ainsworthgit, ilharcoediting}. 

Interestingly, even though full FT is generally less robust than PEFT methods, WiSE elevates full FT’s performance above that of PEFT on both target distribution and distribution shift data, which suggests promising research directions to investigate the underlying mechanism and how to further improve the robustness of PEFT methods.


\section{Conclusion}
Instead of chasing the leaderboard, we conduct a unifying empirical study of PEFT, an emerging topic in the large model era. We provide an extendable framework for reproducible evaluations of PEFT methods in computer vision. We also have several new insights and implications, including PEFT methods' complementary expertise, suitable application regimes, and robustness to domain shifts. We expect our study to open new research directions and serve as a valuable user guide in practice.

\section*{Acknowledgment}
This research is supported by grants from the National Science Foundation (ICICLE: OAC-2112606). We are grateful for the generous support from the Google gift fund and the Ohio Supercomputer Center.



{
    \small
    \bibliographystyle{ieeenat_fullname}
    \bibliography{main}
}
\newpage
\newpage
\appendix
\section*{\LARGE Appendix}


We provide details that are omitted from the main paper. 


\begin{itemize}
    \item \autoref{-sec:setup}: Experiment and dataset details
    \item \autoref{-sec: survey}: Detailed descriptions of ViT and compared methods. 
    \item \autoref{-sec:results}: Additional results not presented in main paper
    \item \autoref{-sec:impacts}: Discussion about further impacts of this work

\end{itemize}

\section{Experiment and Dataset Details}
\label{-sec:setup}

\subsection{Experiment Details}


\paragraph{VTAB-1K}We employ AdamW optimizer~\cite{loshchilov2018decoupled} with a batch size of 64 and utilize the cosine decay learning rate scheduler. We train all methods with 100 epochs. The learning rate is tuned from [1e-3, 1e-2] and weight decay from [1e-4, 1e-3]. The method-specific hyperparameter searching grip is shown in \autoref{tb:hyperparams}, along with the tunable parameter ranges (in millions). Since most method-specific hyperparameters affect the number of tunable parameters in the PEFT methods, we set a cap on the tunable parameters for each PEFT method to be less than or equal to \textbf{1.5\%} of the total parameters in ViT-B/16, which is approximately equal to the number of parameters in the Query, Key, and Value matrices of a single MSA block. Consistent with the original VTAB-1k paper~\citep{zhai2019large}, most PEFT studies~\cite{jie2023fact, convpass,luo2023rep, jia2022visual, zhang2022neural, luo2023rep, ssf} don't apply data augmentation as it's challenging to identify a set of augmentations that uniformly benefits all 19 datasets\footnote{To demonstrate it, we apply simple data augmentations (RandomResizedCrop, RandomVerticalFlip and RandomHorizontalFlip) on three datasets in each group, as shown in \autoref{tab:simple_da}. }. To ensure that our results are directly comparable and that any performance differences are attributable to the methods themselves rather than data augmentation, we don't apply data augmentation.


\paragraph{Many-shot}We also employ AdamW optimizer with a batch size of 64 and a cosine decay learning rate scheduler. The learning rate is tuned from [5e-4, 1e-3] and weight decay keeps the same range of [1e-4, 1e-3]. We apply horizontal flipping for CIFAR100, horizontal and vertical flipping for Resisc, and no augmentation for Clevr. We train all methods with 40 epochs.

\paragraph{Robustness Model} CLIP models are trained on image-caption pairs collected from the web. Given a dataset of such pairs \(\{(x_1, s_1), \ldots, (x_B, s_B)\}\), these models learn an image encoder \(g\) and a text encoder \(h\) that aim to maximize the similarity \(\langle g(x_i), h(s_i) \rangle\) between matching image and caption embeddings while minimizing it for non-matching pairs. For zero-shot classification, the models predict the class of an input image \(x\) from a set of \(k\) class names \(C = \{c_1, \ldots, c_k\}\) by matching \(x\) with captions derived from these class names. Specifically, for each class \(c_j\), a caption is formulated as \(s_j = \text{``a photo of a } c_j\text{''}\). The predicted class is then determined by selecting the one whose caption embedding has the highest similarity with the image embedding: \(\hat{y} = \arg \max_j \langle g(x), h(s_j) \rangle\). Alternatively, a weight matrix \(W_{\text{zero-shot}} \in \mathbb{R}^{d \times k}\) can be constructed, where each column is the embedding \(h(s_j)\) corresponding to class \(c_j\). The model's output scores for each class are then computed as \(f(x) = g(x)^\top W_{\text{zero-shot}}\). To generate \(W_{\text{zero-shot}}\), we ensemble the 80 prompts provided by CLIP at \url{https:
//github.com/openai/CLIP}. 

\begin{figure*}[h]
    \centering
    \begin{subfigure}[b]{0.25\textwidth}
        \includegraphics[width=\textwidth]{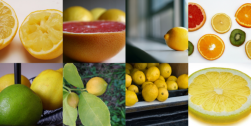}
        \caption{ImageNet~\cite{deng2009imagenet}}
    \end{subfigure}
    \hspace{0.05\textwidth}
    \begin{subfigure}[b]{0.25\textwidth}
        \includegraphics[width=\textwidth]{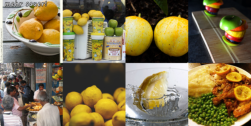}
        \caption{ImageNetV2~\cite{recht2019imagenet}}
    \end{subfigure}
    \hspace{0.05\textwidth}
    \begin{subfigure}[b]{0.25\textwidth}
        \includegraphics[width=\textwidth]{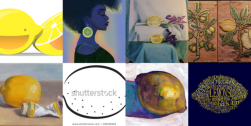}
        \caption{ImageNet-R~\cite{hendrycks2021faces}}
    \end{subfigure}

    \vspace{0.5cm}

    \begin{subfigure}[b]{0.25\textwidth}
        \includegraphics[width=\textwidth]{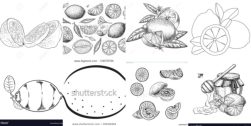}
        \caption{ImageNet Sketch~\cite{wang2019learning}}
    \end{subfigure}
    \hspace{0.05\textwidth}
    \begin{subfigure}[b]{0.25\textwidth}
        \includegraphics[width=\textwidth]{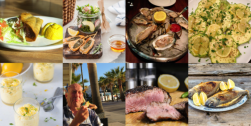}
        \caption{ImageNet-A~\cite{hendrycks2021natural}}
    \end{subfigure}
    
    \caption{Samples of the class lemon, from the fine-tuned dataset ImageNet and distribution shifts datasets (ImageNet-V2, ImageNet-R, ImageNet-S, and ImageNet-A). The CLIP model is fine-tuned with PEFT on ImageNet and evaluated on distribution shifts datasets to measure the robustness of fine-tuned models. The figures are modified based on~\cite{wortsman2022robust}. }
    \label{fig:robustness_setup}
\end{figure*}

\paragraph{Robustness Setup} In \autoref{sec:few} and \autoref{sec:many}, the fine-tuning train set and test set are from the same data distribution. In \autoref{sec: robust}, we fine-tune the CLIP model using ImageNet-1K training data (100 shots) and subsequently evaluate the fine-tuned model not only on the test set of ImageNet-1K but also on four additional datasets with distribution shifts: ImageNet-V2, ImageNet-R, ImageNet-S, and ImageNet-A, as shown in \autoref{fig:robustness_setup}. Following~\cite{wortsman2022robust}, we set a small learning rate as 3e-5 and weight decay as 5e-3. We use a strong data augmentation following~\cite{zhang2022neural}.

\paragraph{Computation}We used a workstation with eight NVIDIA RTX 6000 Ada GPUs, two AMD EPYC 9554 64-Core Processors, and 800GB of RAM. 

{\paragraph{What do we not investigate?} There are many aspects that one can ask about PEFT. Our study focuses more on their learning and prediction behaviors, not the
computation-specific properties like memory usage and FLOPS.}


\subsection{Dataset Details}
\paragraph{VTAB-1K} The processed VTAB-1K can be downloaded from our official code base to ensure reproducibility. 

\paragraph{Many-shot Datasets} We perform 90/10 train-val split for CIFAR-100, RESISC and Clevr-Distance. The split details are provided in our code base for reproducibility. We apply horizontal flipping for CIFAR100,  horizontal and vertical flipping for Resisc, and no augmentation for Clevr. All data are normalized by ImageNet mean and standard deviation.

\begin{table*}
\centering
\begin{tabular}{|c|c|c|}
\hline
\textbf{Method}         & \textbf{Hyperparamters}                                                                                                                                           & \textbf{\#Params (M)} \\ \hline
\textbf{VPT-Shallow}    & Prompt Number: {[}5, 10, 50, 100, 200{]}                                                                                                                                & 0.0003 $\sim$ 0.153    \\ \hline
\textbf{VPT-Deep}       & Prompt Number: {[}5, 10, 50, 100{]}                                                                                                                                     & 0.046 $\sim$ 0.921     \\ \hline
\textbf{BitFit}         & N/A                                                                                                                                                               & 0.102                 \\ \hline
\textbf{DiffFit}        & N/A                                                                                                                                                               & 0.140                 \\ \hline
\textbf{LayerNorm}      & N/A                                                                                                                                                               & 0.038                 \\ \hline
\textbf{SSF}            & N/A                                                                                                                                                               & 0.205                 \\ \hline
\textbf{Pfeif. Adapter} & \begin{tabular}[c]{@{}c@{}}Adapter Scale Factor: {[}0.01, 0.1, 1, 10{]}\\ Adapter Bottleneck: {[}4, 8, 16, 32{]}\end{tabular}                                           & 0.082 $\sim$0.599     \\ \hline
\textbf{Houl. Adapter}  & \begin{tabular}[c]{@{}c@{}}Adapter Scale Factor: {[}0.01, 0.1, 1, 10{]}\\ Adapter Bottleneck: {[}4, 8, 16, 32{]}\end{tabular}                                           & 0.165 $\sim$1.198     \\ \hline
\textbf{AdaptFormer}    & \begin{tabular}[c]{@{}c@{}}Adapter Scale Factor: {[}0.05, 0.1, 0.2{]}\\ Adapter Bottleneck: {[}4, 16, 32{]}\end{tabular}                                                & 0.082 $\sim$0.599     \\ \hline
\textbf{RepAdapter}     & \begin{tabular}[c]{@{}c@{}}RepAdapter Scale Factor: {[}0.1, 0.5, 1, 5, 10{]}\\ RepAdapter Bottleneck: {[}8, 16, 32{]}\end{tabular}                                      & 0.239 $\sim$0.903     \\ \hline
\textbf{Convpass}       & \begin{tabular}[c]{@{}c@{}}Convpass Scale Factor: {[}0.01, 0.1, 1, 10, 100{]}\\ Convpass Bottleneck: {[}8, 16{]}\\ Convpass Xavier Init: {[}True, False{]}\end{tabular} & 0.327 $\sim$0.664     \\ \hline
\textbf{LoRA}           & LoRA Bottleneck: {[}1, 8, 16, 32{]}                                                                                                                               & 0.036 $\sim$1.179     \\ \hline
\textbf{FacT\_TT}       & \begin{tabular}[c]{@{}c@{}}FacT Scale Factor: {[}0.01, 0.1, 1, 10, 100{]}\\ FacT Bottleneck: {[}8, 16, 32{]}\end{tabular}                                                      & 0.021 $\sim$0.196     \\ \hline
\textbf{FacT\_TK}       & \begin{tabular}[c]{@{}c@{}}FacT Bottleneck: {[}16, 32, 64{]}\\ FacT Scale Factor: {[}0.01, 0.1, 1, 10, 100{]}\end{tabular}                                                     & 0.030 $\sim$0.369     \\ \hline
\end{tabular}
\caption{Methods-specific hyperparameter searching grip for VTAB-1K experiment. }
\label{tb:hyperparams}
\end{table*}

\begin{table*}
\resizebox{\textwidth}{!}{
\begin{tabular}{|c|c|c|cccccccccccccccc|}
\hline
\multicolumn{1}{|l|}{}                 & \multicolumn{1}{l|}{}                  & \multicolumn{1}{l|}{} & \multicolumn{1}{c|}{\textbf{Linear}}                                     & \multicolumn{1}{c|}{\textbf{Full}}                                       & \multicolumn{1}{c|}{\textbf{VPT-Shallow}}                                & \multicolumn{1}{c|}{\textbf{VPT-Deep}}                                    & \multicolumn{1}{c|}{\textbf{BitFit}}                                     & \multicolumn{1}{c|}{\textbf{DiffFit}}                                    & \multicolumn{1}{c|}{\textbf{LayerNorm}}                                  & \multicolumn{1}{c|}{\textbf{SSF}}                                        & \multicolumn{1}{c|}{\textbf{\begin{tabular}[c]{@{}c@{}}Pfeif. \\ Adapter\end{tabular}}} & \multicolumn{1}{c|}{\textbf{\begin{tabular}[c]{@{}c@{}}Houl. \\ Adapter\end{tabular}}} & \multicolumn{1}{c|}{\textbf{\begin{tabular}[c]{@{}c@{}}Adapt-\\ Former\end{tabular}}} & \multicolumn{1}{c|}{\textbf{\begin{tabular}[c]{@{}c@{}}Rep-\\ Adapter\end{tabular}}} & \multicolumn{1}{c|}{\textbf{Convpass}}                                    & \multicolumn{1}{c|}{\textbf{LoRA}}                                       & \textbf{FacT\_TT}                                    & \textbf{Fact\_TK}                                    \\ \hline
                                       &                                        & \textbf{Simple DA}    & \cellcolor[HTML]{FFFFFF}{\color[HTML]{1F2328} 84.4}                      & \cellcolor[HTML]{FFFFFF}{\color[HTML]{1F2328} 76.8}                      & \cellcolor[HTML]{FFFFFF}{\color[HTML]{1F2328} 84.9}                      & \cellcolor[HTML]{FFFFFF}{\color[HTML]{1F2328} 84.8}                       & \cellcolor[HTML]{FFFFFF}{\color[HTML]{1F2328} 83.8}                      & \cellcolor[HTML]{FFFFFF}{\color[HTML]{1F2328} 85.7}                      & \cellcolor[HTML]{FFFFFF}{\color[HTML]{1F2328} 85.8}                      & \cellcolor[HTML]{FFFFFF}{\color[HTML]{1F2328} 86.1}                      & \cellcolor[HTML]{FFFFFF}{\color[HTML]{1F2328} 87.4}                      & \cellcolor[HTML]{FFFFFF}{\color[HTML]{1F2328} 86.0}                      & \cellcolor[HTML]{FFFFFF}{\color[HTML]{1F2328} 84.9}                      & \cellcolor[HTML]{FFFFFF}{\color[HTML]{1F2328} 86.4}                      & \cellcolor[HTML]{FFFFFF}{\color[HTML]{1F2328} 85.2}                       & \cellcolor[HTML]{FFFFFF}{\color[HTML]{1F2328} 86.8}                      & \cellcolor[HTML]{FFFFFF}{\color[HTML]{1F2328} 85.5}  & \cellcolor[HTML]{FFFFFF}{\color[HTML]{1F2328} 86.0}  \\ \cline{3-3} 
                                       &                                        & \textbf{Default}        & \cellcolor[HTML]{FFFFFF}{\color[HTML]{1F2328} 86.6}                      & \cellcolor[HTML]{FFFFFF}{\color[HTML]{1F2328} 89.9}                      & \cellcolor[HTML]{FFFFFF}{\color[HTML]{1F2328} 88.7}                      & \cellcolor[HTML]{FFFFFF}{\color[HTML]{1F2328} 91.5}                       & \cellcolor[HTML]{FFFFFF}{\color[HTML]{1F2328} 90.5}                      & \cellcolor[HTML]{FFFFFF}{\color[HTML]{1F2328} 90.2}                      & \cellcolor[HTML]{FFFFFF}{\color[HTML]{1F2328} 89.7}                      & \cellcolor[HTML]{FFFFFF}{\color[HTML]{1F2328} 89.8}                      & \cellcolor[HTML]{FFFFFF}{\color[HTML]{1F2328} 91.5}                      & \cellcolor[HTML]{FFFFFF}{\color[HTML]{1F2328} 92.1}                      & \cellcolor[HTML]{FFFFFF}{\color[HTML]{1F2328} 91.8}                      & \cellcolor[HTML]{FFFFFF}{\color[HTML]{1F2328} 92.5}                      & \cellcolor[HTML]{FFFFFF}{\color[HTML]{1F2328} 92.1}                       & \cellcolor[HTML]{FFFFFF}{\color[HTML]{1F2328} 92.6}                      & \cellcolor[HTML]{FFFFFF}{\color[HTML]{1F2328} 91.8}  & \cellcolor[HTML]{FFFFFF}{\color[HTML]{1F2328} 92.5}  \\ \cline{3-3} 
                                       & \multirow{-3}{*}{\textbf{Caltech101}}  & $\Delta$            & \cellcolor[HTML]{FFFFFF}{\color[HTML]{FF0000} -2.2}                      & \cellcolor[HTML]{FFFFFF}{\color[HTML]{FF0000} -13.1}                     & \cellcolor[HTML]{FFFFFF}{\color[HTML]{FF0000} -3.8}                      & \cellcolor[HTML]{FFFFFF}{\color[HTML]{FF0000} -6.7}                       & \cellcolor[HTML]{FFFFFF}{\color[HTML]{FF0000} -6.7}                      & \cellcolor[HTML]{FFFFFF}{\color[HTML]{FF0000} -4.5}                      & \cellcolor[HTML]{FFFFFF}{\color[HTML]{FF0000} -3.9}                      & \cellcolor[HTML]{FFFFFF}{\color[HTML]{FF0000} -3.7}                      & \cellcolor[HTML]{FFFFFF}{\color[HTML]{FF0000} -4.1}                      & \cellcolor[HTML]{FFFFFF}{\color[HTML]{FF0000} -6.1}                      & \cellcolor[HTML]{FFFFFF}{\color[HTML]{FF0000} -6.9}                      & \cellcolor[HTML]{FFFFFF}{\color[HTML]{FF0000} -6.1}                      & \cellcolor[HTML]{FFFFFF}{\color[HTML]{FF0000} -6.9}                       & \cellcolor[HTML]{FFFFFF}{\color[HTML]{FF0000} -5.8}                      & \cellcolor[HTML]{FFFFFF}{\color[HTML]{FF0000} -6.3}  & \cellcolor[HTML]{FFFFFF}{\color[HTML]{FF0000} -6.5}  \\ \hline 
                                       &                                        & \textbf{Simple DA}    & \cellcolor[HTML]{FFFFFF}{\color[HTML]{1F2328} 67.5}                      & \cellcolor[HTML]{FFFFFF}{\color[HTML]{1F2328} 57.8}                      & \cellcolor[HTML]{FFFFFF}{\color[HTML]{1F2328} 69.0}                      & \cellcolor[HTML]{FFFFFF}{\color[HTML]{1F2328} 71.1}                       & \cellcolor[HTML]{FFFFFF}{\color[HTML]{1F2328} 70.7}                      & \cellcolor[HTML]{FFFFFF}{\color[HTML]{1F2328} 73.7}                      & \cellcolor[HTML]{FFFFFF}{\color[HTML]{1F2328} 73.5}                      & \cellcolor[HTML]{FFFFFF}{\color[HTML]{1F2328} 68.4}                      & \cellcolor[HTML]{FFFFFF}{\color[HTML]{1F2328} 72.7}                      & \cellcolor[HTML]{FFFFFF}{\color[HTML]{1F2328} 72.6}                      & \cellcolor[HTML]{FFFFFF}{\color[HTML]{1F2328} 70.6}                      & \cellcolor[HTML]{FFFFFF}{\color[HTML]{1F2328} 71.5}                      & \cellcolor[HTML]{FFFFFF}{\color[HTML]{1F2328} 71.8}                       & \cellcolor[HTML]{FFFFFF}{\color[HTML]{1F2328} 73.0}                      & \cellcolor[HTML]{FFFFFF}{\color[HTML]{1F2328} 72.2}  & \cellcolor[HTML]{FFFFFF}{\color[HTML]{1F2328} 71.9}  \\ \cline{3-3} 
                                       &                                        & \textbf{Default}        & \cellcolor[HTML]{FFFFFF}{\color[HTML]{1F2328} 65.7}                      & \cellcolor[HTML]{FFFFFF}{\color[HTML]{1F2328} 61.9}                      & \cellcolor[HTML]{FFFFFF}{\color[HTML]{1F2328} 67.9}                      & \cellcolor[HTML]{FFFFFF}{\color[HTML]{1F2328} 69.4}                       & \cellcolor[HTML]{FFFFFF}{\color[HTML]{1F2328} 70.3}                      & \cellcolor[HTML]{FFFFFF}{\color[HTML]{1F2328} 71.2}                      & \cellcolor[HTML]{FFFFFF}{\color[HTML]{1F2328} 72.2}                      & \cellcolor[HTML]{FFFFFF}{\color[HTML]{1F2328} 68.8}                      & \cellcolor[HTML]{FFFFFF}{\color[HTML]{1F2328} 72.1}                      & \cellcolor[HTML]{FFFFFF}{\color[HTML]{1F2328} 72.3}                      & \cellcolor[HTML]{FFFFFF}{\color[HTML]{1F2328} 70.5}                      & \cellcolor[HTML]{FFFFFF}{\color[HTML]{1F2328} 69.1}                      & \cellcolor[HTML]{FFFFFF}{\color[HTML]{1F2328} 72.0}                       & \cellcolor[HTML]{FFFFFF}{\color[HTML]{1F2328} 69.8}                      & \cellcolor[HTML]{FFFFFF}{\color[HTML]{1F2328} 71.5}  & \cellcolor[HTML]{FFFFFF}{\color[HTML]{1F2328} 71.8}  \\ \cline{3-3} 
                                       & \multirow{-3}{*}{\textbf{DTD}}         & $\Delta$            & \cellcolor[HTML]{FFFFFF}{\color[HTML]{38761D} 1.8}                       & \cellcolor[HTML]{FFFFFF}{\color[HTML]{FF0000} -4.1}                      & \cellcolor[HTML]{FFFFFF}{\color[HTML]{38761D} 1.1}                       & \cellcolor[HTML]{FFFFFF}{\color[HTML]{38761D} 1.7}                        & \cellcolor[HTML]{FFFFFF}{\color[HTML]{38761D} 0.4}                       & \cellcolor[HTML]{FFFFFF}{\color[HTML]{38761D} 2.5}                       & \cellcolor[HTML]{FFFFFF}{\color[HTML]{38761D} 1.3}                       & \cellcolor[HTML]{FFFFFF}{\color[HTML]{FF0000} -0.4}                      & \cellcolor[HTML]{FFFFFF}{\color[HTML]{38761D} 0.6}                       & \cellcolor[HTML]{FFFFFF}{\color[HTML]{38761D} 0.3}                       & \cellcolor[HTML]{FFFFFF}{\color[HTML]{38761D} 0.1}                       & \cellcolor[HTML]{FFFFFF}{\color[HTML]{38761D} 2.4}                       & \cellcolor[HTML]{FFFFFF}{\color[HTML]{FF0000} -0.2}                       & \cellcolor[HTML]{FFFFFF}{\color[HTML]{38761D} 3.2}                       & \cellcolor[HTML]{FFFFFF}{\color[HTML]{38761D} 0.7}   & \cellcolor[HTML]{FFFFFF}{\color[HTML]{38761D} 0.1}   \\ \hline 
                                       &                                        & \textbf{Simple DA}    & \cellcolor[HTML]{FFFFFF}{\color[HTML]{1F2328} 98.1}                      & \cellcolor[HTML]{FFFFFF}{\color[HTML]{1F2328} 92.2}                      & \cellcolor[HTML]{FFFFFF}{\color[HTML]{1F2328} 98.2}                      & \cellcolor[HTML]{FFFFFF}{\color[HTML]{1F2328} 98.6}                       & \cellcolor[HTML]{FFFFFF}{\color[HTML]{1F2328} 98.0}                      & \cellcolor[HTML]{FFFFFF}{\color[HTML]{1F2328} 98.8}                      & \cellcolor[HTML]{FFFFFF}{\color[HTML]{1F2328} 98.8}                      & \cellcolor[HTML]{FFFFFF}{\color[HTML]{1F2328} 98.8}                      & \cellcolor[HTML]{FFFFFF}{\color[HTML]{1F2328} 98.4}                      & \cellcolor[HTML]{FFFFFF}{\color[HTML]{1F2328} 97.5}                      & \cellcolor[HTML]{FFFFFF}{\color[HTML]{1F2328} 98.7}                      & \cellcolor[HTML]{FFFFFF}{\color[HTML]{1F2328} 98.0}                      & \cellcolor[HTML]{FFFFFF}{\color[HTML]{1F2328} 98.9}                       & \cellcolor[HTML]{FFFFFF}{\color[HTML]{1F2328} 98.7}                      & \cellcolor[HTML]{FFFFFF}{\color[HTML]{1F2328} 98.7}  & \cellcolor[HTML]{FFFFFF}{\color[HTML]{1F2328} 98.7}  \\ \cline{3-3} 
                                       &                                        & \textbf{Default}        & \cellcolor[HTML]{FFFFFF}{\color[HTML]{1F2328} 98.9}                      & \cellcolor[HTML]{FFFFFF}{\color[HTML]{1F2328} 97.4}                      & \cellcolor[HTML]{FFFFFF}{\color[HTML]{1F2328} 99.1}                      & \cellcolor[HTML]{FFFFFF}{\color[HTML]{1F2328} 99.1}                       & \cellcolor[HTML]{FFFFFF}{\color[HTML]{1F2328} 98.9}                      & \cellcolor[HTML]{FFFFFF}{\color[HTML]{1F2328} 99.2}                      & \cellcolor[HTML]{FFFFFF}{\color[HTML]{1F2328} 99.1}                      & \cellcolor[HTML]{FFFFFF}{\color[HTML]{1F2328} 99.1}                      & \cellcolor[HTML]{FFFFFF}{\color[HTML]{1F2328} 99.2}                      & \cellcolor[HTML]{FFFFFF}{\color[HTML]{1F2328} 98.0}                      & \cellcolor[HTML]{FFFFFF}{\color[HTML]{1F2328} 99.2}                      & \cellcolor[HTML]{FFFFFF}{\color[HTML]{1F2328} 99.1}                      & \cellcolor[HTML]{FFFFFF}{\color[HTML]{1F2328} 99.3}                       & \cellcolor[HTML]{FFFFFF}{\color[HTML]{1F2328} 99.1}                      & \cellcolor[HTML]{FFFFFF}{\color[HTML]{1F2328} 99.3}  & \cellcolor[HTML]{FFFFFF}{\color[HTML]{1F2328} 99.1}  \\ \cline{3-3} 
\multirow{-9}{*}{\textbf{Natural}}     & \multirow{-3}{*}{\textbf{Flower102}}   & $\Delta$            & \cellcolor[HTML]{FFFFFF}{\color[HTML]{FF0000} -0.8}                      & \cellcolor[HTML]{FFFFFF}{\color[HTML]{FF0000} -5.2}                      & \cellcolor[HTML]{FFFFFF}{\color[HTML]{FF0000} -0.9}                      & \cellcolor[HTML]{FFFFFF}{\color[HTML]{FF0000} -0.5}                       & \cellcolor[HTML]{FFFFFF}{\color[HTML]{FF0000} -0.9}                      & \cellcolor[HTML]{FFFFFF}{\color[HTML]{FF0000} -0.4}                      & \cellcolor[HTML]{FFFFFF}{\color[HTML]{FF0000} -0.3}                      & \cellcolor[HTML]{FFFFFF}{\color[HTML]{FF0000} -0.3}                      & \cellcolor[HTML]{FFFFFF}{\color[HTML]{FF0000} -0.8}                      & \cellcolor[HTML]{FFFFFF}{\color[HTML]{FF0000} -0.5}                      & \cellcolor[HTML]{FFFFFF}{\color[HTML]{FF0000} -0.5}                      & \cellcolor[HTML]{FFFFFF}{\color[HTML]{FF0000} -1.1}                      & \cellcolor[HTML]{FFFFFF}{\color[HTML]{FF0000} -0.4}                       & \cellcolor[HTML]{FFFFFF}{\color[HTML]{FF0000} -0.4}                      & \cellcolor[HTML]{FFFFFF}{\color[HTML]{FF0000} -0.6}  & \cellcolor[HTML]{FFFFFF}{\color[HTML]{FF0000} -0.4}  \\ \hline 
                                       &                                        & \textbf{Simple DA}    & \cellcolor[HTML]{FFFFFF}{\color[HTML]{1F2328} 87.3}                      & \cellcolor[HTML]{FFFFFF}{\color[HTML]{1F2328} 91.0}                      & \cellcolor[HTML]{FFFFFF}{\color[HTML]{1F2328} 88.4}                      & \cellcolor[HTML]{FFFFFF}{\color[HTML]{1F2328} 92.0}                       & \cellcolor[HTML]{FFFFFF}{\color[HTML]{1F2328} 92.0}                      & \cellcolor[HTML]{FFFFFF}{\color[HTML]{1F2328} 91.7}                      & \cellcolor[HTML]{FFFFFF}{\color[HTML]{1F2328} 91.9}                      & \cellcolor[HTML]{FFFFFF}{\color[HTML]{1F2328} 92.5}                      & \cellcolor[HTML]{FFFFFF}{\color[HTML]{1F2328} 92.1}                      & \cellcolor[HTML]{FFFFFF}{\color[HTML]{1F2328} 92.5}                      & \cellcolor[HTML]{FFFFFF}{\color[HTML]{1F2328} 92.3}                      & \cellcolor[HTML]{FFFFFF}{\color[HTML]{1F2328} 93.1}                      & \cellcolor[HTML]{FFFFFF}{\color[HTML]{1F2328} 92.7}                       & \cellcolor[HTML]{FFFFFF}{\color[HTML]{1F2328} 92.8}                      & \cellcolor[HTML]{FFFFFF}{\color[HTML]{1F2328} 93.3}  & \cellcolor[HTML]{FFFFFF}{\color[HTML]{1F2328} 92.8}  \\ \cline{3-3} 
                                       &                                        & \textbf{Default}        & \cellcolor[HTML]{FFFFFF}{\color[HTML]{1F2328} 90.0}                      & \cellcolor[HTML]{FFFFFF}{\color[HTML]{1F2328} 88.1}                      & \cellcolor[HTML]{FFFFFF}{\color[HTML]{1F2328} 90.3}                      & \cellcolor[HTML]{FFFFFF}{\color[HTML]{1F2328} 94.9}                       & \cellcolor[HTML]{FFFFFF}{\color[HTML]{1F2328} 95.0}                      & \cellcolor[HTML]{FFFFFF}{\color[HTML]{1F2328} 94.1}                      & \cellcolor[HTML]{FFFFFF}{\color[HTML]{1F2328} 93.8}                      & \cellcolor[HTML]{FFFFFF}{\color[HTML]{1F2328} 94.5}                      & \cellcolor[HTML]{FFFFFF}{\color[HTML]{1F2328} 95.5}                      & \cellcolor[HTML]{FFFFFF}{\color[HTML]{1F2328} 95.3}                      & \cellcolor[HTML]{FFFFFF}{\color[HTML]{1F2328} 95.0}                      & \cellcolor[HTML]{FFFFFF}{\color[HTML]{1F2328} 95.3}                      & \cellcolor[HTML]{FFFFFF}{\color[HTML]{1F2328} 95.8}                       & \cellcolor[HTML]{FFFFFF}{\color[HTML]{1F2328} 94.9}                      & \cellcolor[HTML]{FFFFFF}{\color[HTML]{1F2328} 94.9}  & \cellcolor[HTML]{FFFFFF}{\color[HTML]{1F2328} 95.5}  \\ \cline{3-3} 
                                       & \multirow{-3}{*}{\textbf{EuroSAT}}     & $\Delta$            & \cellcolor[HTML]{FFFFFF}{\color[HTML]{FF0000} -2.7}                      & \cellcolor[HTML]{FFFFFF}{\color[HTML]{38761D} 2.9}                       & \cellcolor[HTML]{FFFFFF}{\color[HTML]{FF0000} -1.9}                      & \cellcolor[HTML]{FFFFFF}{\color[HTML]{FF0000} -2.9}                       & \cellcolor[HTML]{FFFFFF}{\color[HTML]{FF0000} -3.0}                      & \cellcolor[HTML]{FFFFFF}{\color[HTML]{FF0000} -2.4}                      & \cellcolor[HTML]{FFFFFF}{\color[HTML]{FF0000} -1.9}                      & \cellcolor[HTML]{FFFFFF}{\color[HTML]{FF0000} -2.0}                      & \cellcolor[HTML]{FFFFFF}{\color[HTML]{FF0000} -3.4}                      & \cellcolor[HTML]{FFFFFF}{\color[HTML]{FF0000} -2.8}                      & \cellcolor[HTML]{FFFFFF}{\color[HTML]{FF0000} -2.7}                      & \cellcolor[HTML]{FFFFFF}{\color[HTML]{FF0000} -2.2}                      & \cellcolor[HTML]{FFFFFF}{\color[HTML]{FF0000} -3.1}                       & \cellcolor[HTML]{FFFFFF}{\color[HTML]{FF0000} -2.1}                      & \cellcolor[HTML]{FFFFFF}{\color[HTML]{FF0000} -1.6}  & \cellcolor[HTML]{FFFFFF}{\color[HTML]{FF0000} -2.7}  \\ \hline 
                                       &                                        & \textbf{Simple DA}    & \cellcolor[HTML]{FFFFFF}{\color[HTML]{1F2328} 74.3}                      & \cellcolor[HTML]{FFFFFF}{\color[HTML]{1F2328} 75.0}                      & \cellcolor[HTML]{FFFFFF}{\color[HTML]{1F2328} 74.4}                      & \cellcolor[HTML]{FFFFFF}{\color[HTML]{1F2328} 80.1}                       & \cellcolor[HTML]{FFFFFF}{\color[HTML]{1F2328} 81.0}                      & \cellcolor[HTML]{FFFFFF}{\color[HTML]{1F2328} 78.5}                      & \cellcolor[HTML]{FFFFFF}{\color[HTML]{1F2328} 80.7}                      & \cellcolor[HTML]{FFFFFF}{\color[HTML]{1F2328} 80.6}                      & \cellcolor[HTML]{FFFFFF}{\color[HTML]{1F2328} 80.6}                      & \cellcolor[HTML]{FFFFFF}{\color[HTML]{1F2328} 81.6}                      & \cellcolor[HTML]{FFFFFF}{\color[HTML]{1F2328} 82.2}                      & \cellcolor[HTML]{FFFFFF}{\color[HTML]{1F2328} 81.5}                      & \cellcolor[HTML]{FFFFFF}{\color[HTML]{1F2328} 81.5}                       & \cellcolor[HTML]{FFFFFF}{\color[HTML]{1F2328} 82.2}                      & \cellcolor[HTML]{FFFFFF}{\color[HTML]{1F2328} 80.7}  & \cellcolor[HTML]{FFFFFF}{\color[HTML]{1F2328} 82.9}  \\ \cline{3-3} 
                                       &                                        & \textbf{Default}        & \cellcolor[HTML]{FFFFFF}{\color[HTML]{1F2328} 74.9}                      & \cellcolor[HTML]{FFFFFF}{\color[HTML]{1F2328} 81.6}                      & \cellcolor[HTML]{FFFFFF}{\color[HTML]{1F2328} 77.2}                      & \cellcolor[HTML]{FFFFFF}{\color[HTML]{1F2328} 84.2}                       & \cellcolor[HTML]{FFFFFF}{\color[HTML]{1F2328} 85.3}                      & \cellcolor[HTML]{FFFFFF}{\color[HTML]{1F2328} 80.9}                      & \cellcolor[HTML]{FFFFFF}{\color[HTML]{1F2328} 83.0}                      & \cellcolor[HTML]{FFFFFF}{\color[HTML]{1F2328} 83.2}                      & \cellcolor[HTML]{FFFFFF}{\color[HTML]{1F2328} 85.3}                      & \cellcolor[HTML]{FFFFFF}{\color[HTML]{1F2328} 86.5}                      & \cellcolor[HTML]{FFFFFF}{\color[HTML]{1F2328} 86.5}                      & \cellcolor[HTML]{FFFFFF}{\color[HTML]{1F2328} 86.0}                      & \cellcolor[HTML]{FFFFFF}{\color[HTML]{1F2328} 85.9}                       & \cellcolor[HTML]{FFFFFF}{\color[HTML]{1F2328} 85.9}                      & \cellcolor[HTML]{FFFFFF}{\color[HTML]{1F2328} 85.0}  & \cellcolor[HTML]{FFFFFF}{\color[HTML]{1F2328} 86.0}  \\ \cline{3-3} 
                                       & \multirow{-3}{*}{\textbf{Resisc45}}    & $\Delta$            & \cellcolor[HTML]{FFFFFF}{\color[HTML]{FF0000} -0.6}                      & \cellcolor[HTML]{FFFFFF}{\color[HTML]{FF0000} -6.6}                      & \cellcolor[HTML]{FFFFFF}{\color[HTML]{FF0000} -2.8}                      & \cellcolor[HTML]{FFFFFF}{\color[HTML]{FF0000} -4.1}                       & \cellcolor[HTML]{FFFFFF}{\color[HTML]{FF0000} -4.3}                      & \cellcolor[HTML]{FFFFFF}{\color[HTML]{FF0000} -2.4}                      & \cellcolor[HTML]{FFFFFF}{\color[HTML]{FF0000} -2.3}                      & \cellcolor[HTML]{FFFFFF}{\color[HTML]{FF0000} -2.6}                      & \cellcolor[HTML]{FFFFFF}{\color[HTML]{FF0000} -4.7}                      & \cellcolor[HTML]{FFFFFF}{\color[HTML]{FF0000} -4.9}                      & \cellcolor[HTML]{FFFFFF}{\color[HTML]{FF0000} -4.3}                      & \cellcolor[HTML]{FFFFFF}{\color[HTML]{FF0000} -4.5}                      & \cellcolor[HTML]{FFFFFF}{\color[HTML]{FF0000} -4.4}                       & \cellcolor[HTML]{FFFFFF}{\color[HTML]{FF0000} -3.7}                      & \cellcolor[HTML]{FFFFFF}{\color[HTML]{FF0000} -4.3}  & \cellcolor[HTML]{FFFFFF}{\color[HTML]{FF0000} -3.1}  \\ \hline 
                                       &                                        & \textbf{Simple DA}    & \cellcolor[HTML]{FFFFFF}{\color[HTML]{1F2328} 74.5}                      & \cellcolor[HTML]{FFFFFF}{\color[HTML]{1F2328} 73.6}                      & \cellcolor[HTML]{FFFFFF}{\color[HTML]{1F2328} 74.7}                      & \cellcolor[HTML]{FFFFFF}{\color[HTML]{1F2328} 76.3}                       & \cellcolor[HTML]{FFFFFF}{\color[HTML]{1F2328} 76.3}                      & \cellcolor[HTML]{FFFFFF}{\color[HTML]{1F2328} 76.7}                      & \cellcolor[HTML]{FFFFFF}{\color[HTML]{1F2328} 76.4}                      & \cellcolor[HTML]{FFFFFF}{\color[HTML]{1F2328} 76.4}                      & \cellcolor[HTML]{FFFFFF}{\color[HTML]{1F2328} 77.3}                      & \cellcolor[HTML]{FFFFFF}{\color[HTML]{1F2328} 75.6}                      & \cellcolor[HTML]{FFFFFF}{\color[HTML]{1F2328} 77.0}                      & \cellcolor[HTML]{FFFFFF}{\color[HTML]{1F2328} 77.1}                      & \cellcolor[HTML]{FFFFFF}{\color[HTML]{1F2328} 76.8}                       & \cellcolor[HTML]{FFFFFF}{\color[HTML]{1F2328} 76.2}                      & \cellcolor[HTML]{FFFFFF}{\color[HTML]{1F2328} 75.3}  & \cellcolor[HTML]{FFFFFF}{\color[HTML]{1F2328} 77.0}  \\ \cline{3-3} 
                                       &                                        & \textbf{Default}        & \cellcolor[HTML]{FFFFFF}{\color[HTML]{1F2328} 74.6}                      & \cellcolor[HTML]{FFFFFF}{\color[HTML]{1F2328} 73.6}                      & \cellcolor[HTML]{FFFFFF}{\color[HTML]{1F2328} 74.4}                      & \cellcolor[HTML]{FFFFFF}{\color[HTML]{1F2328} 73.9}                       & \cellcolor[HTML]{FFFFFF}{\color[HTML]{1F2328} 75.5}                      & \cellcolor[HTML]{FFFFFF}{\color[HTML]{1F2328} 75.2}                      & \cellcolor[HTML]{FFFFFF}{\color[HTML]{1F2328} 75.2}                      & \cellcolor[HTML]{FFFFFF}{\color[HTML]{1F2328} 74.8}                      & \cellcolor[HTML]{FFFFFF}{\color[HTML]{1F2328} 76.2}                      & \cellcolor[HTML]{FFFFFF}{\color[HTML]{1F2328} 75.2}                      & \cellcolor[HTML]{FFFFFF}{\color[HTML]{1F2328} 76.3}                      & \cellcolor[HTML]{FFFFFF}{\color[HTML]{1F2328} 75.4}                      & \cellcolor[HTML]{FFFFFF}{\color[HTML]{1F2328} 75.9}                       & \cellcolor[HTML]{FFFFFF}{\color[HTML]{1F2328} 75.7}                      & \cellcolor[HTML]{FFFFFF}{\color[HTML]{1F2328} 75.6}  & \cellcolor[HTML]{FFFFFF}{\color[HTML]{1F2328} 75.7}  \\ \cline{3-3} 
\multirow{-9}{*}{\textbf{Specialized}} & \multirow{-3}{*}{\textbf{Retinopathy}} & $\Delta$            & \cellcolor[HTML]{FFFFFF}{\color[HTML]{FF0000} -0.1}                      & \cellcolor[HTML]{FFFFFF}{\color[HTML]{FF0000} 0.0}                       & \cellcolor[HTML]{FFFFFF}{\color[HTML]{38761D} 0.3}                       & \cellcolor[HTML]{FFFFFF}{\color[HTML]{38761D} 2.4}                        & \cellcolor[HTML]{FFFFFF}{\color[HTML]{38761D} 0.8}                       & \cellcolor[HTML]{FFFFFF}{\color[HTML]{38761D} 1.5}                       & \cellcolor[HTML]{FFFFFF}{\color[HTML]{38761D} 1.2}                       & \cellcolor[HTML]{FFFFFF}{\color[HTML]{38761D} 1.6}                       & \cellcolor[HTML]{FFFFFF}{\color[HTML]{38761D} 1.1}                       & \cellcolor[HTML]{FFFFFF}{\color[HTML]{38761D} 0.4}                       & \cellcolor[HTML]{FFFFFF}{\color[HTML]{38761D} 0.7}                       & \cellcolor[HTML]{FFFFFF}{\color[HTML]{38761D} 1.7}                       & \cellcolor[HTML]{FFFFFF}{\color[HTML]{38761D} 0.9}                        & \cellcolor[HTML]{FFFFFF}{\color[HTML]{38761D} 0.5}                       & \cellcolor[HTML]{FFFFFF}{\color[HTML]{FF0000} -0.3}  & \cellcolor[HTML]{FFFFFF}{\color[HTML]{38761D} 1.3}   \\ \hline 
                                       &                                        & \textbf{Simple DA}    & \cellcolor[HTML]{FFFFFF}{\color[HTML]{1F2328} 22.6}                      & \cellcolor[HTML]{FFFFFF}{\color[HTML]{1F2328} 29.9}                      & \cellcolor[HTML]{FFFFFF}{\color[HTML]{1F2328} 24.3}                      & \cellcolor[HTML]{FFFFFF}{\color[HTML]{1F2328} 29.6}                       & \cellcolor[HTML]{FFFFFF}{\color[HTML]{1F2328} 28.7}                      & \cellcolor[HTML]{FFFFFF}{\color[HTML]{1F2328} 29.0}                      & \cellcolor[HTML]{FFFFFF}{\color[HTML]{1F2328} 29.0}                      & \cellcolor[HTML]{FFFFFF}{\color[HTML]{1F2328} 27.9}                      & \cellcolor[HTML]{FFFFFF}{\color[HTML]{1F2328} 28.9}                      & \cellcolor[HTML]{FFFFFF}{\color[HTML]{1F2328} 22.9}                      & \cellcolor[HTML]{FFFFFF}{\color[HTML]{1F2328} 30.9}                      & \cellcolor[HTML]{FFFFFF}{\color[HTML]{1F2328} 31.4}                      & \cellcolor[HTML]{FFFFFF}{\color[HTML]{1F2328} 30.5}                       & \cellcolor[HTML]{FFFFFF}{\color[HTML]{1F2328} 30.3}                      & \cellcolor[HTML]{FFFFFF}{\color[HTML]{1F2328} 32.5}  & \cellcolor[HTML]{FFFFFF}{\color[HTML]{1F2328} 28.4}  \\ \cline{3-3} 
                                       &                                        & \textbf{Default}        & \cellcolor[HTML]{FFFFFF}{\color[HTML]{1F2328} 29.4}                      & \cellcolor[HTML]{FFFFFF}{\color[HTML]{1F2328} 46.6}                      & \cellcolor[HTML]{FFFFFF}{\color[HTML]{1F2328} 43.1}                      & \cellcolor[HTML]{FFFFFF}{\color[HTML]{1F2328} 56.4}                       & \cellcolor[HTML]{FFFFFF}{\color[HTML]{1F2328} 53.9}                      & \cellcolor[HTML]{FFFFFF}{\color[HTML]{1F2328} 52.8}                      & \cellcolor[HTML]{FFFFFF}{\color[HTML]{1F2328} 52.1}                      & \cellcolor[HTML]{FFFFFF}{\color[HTML]{1F2328} 52.1}                      & \cellcolor[HTML]{FFFFFF}{\color[HTML]{1F2328} 56.6}                      & \cellcolor[HTML]{FFFFFF}{\color[HTML]{1F2328} 54.3}                      & \cellcolor[HTML]{FFFFFF}{\color[HTML]{1F2328} 53.0}                      & \cellcolor[HTML]{FFFFFF}{\color[HTML]{1F2328} 52.1}                      & \cellcolor[HTML]{FFFFFF}{\color[HTML]{1F2328} 55.3}                       & \cellcolor[HTML]{FFFFFF}{\color[HTML]{1F2328} 47.2}                      & \cellcolor[HTML]{FFFFFF}{\color[HTML]{1F2328} 53.1}  & \cellcolor[HTML]{FFFFFF}{\color[HTML]{1F2328} 53.1}  \\ \cline{3-3} 
                                       & \multirow{-3}{*}{\textbf{dSpr-Ori}}    & $\Delta$            & \cellcolor[HTML]{FFFFFF}{\color[HTML]{FF0000} -6.8}                      & \cellcolor[HTML]{FFFFFF}{\color[HTML]{FF0000} -16.7}                     & \cellcolor[HTML]{FFFFFF}{\color[HTML]{FF0000} -18.8}                     & \cellcolor[HTML]{FFFFFF}{\color[HTML]{FF0000} -26.8}                      & \cellcolor[HTML]{FFFFFF}{\color[HTML]{FF0000} -25.2}                     & \cellcolor[HTML]{FFFFFF}{\color[HTML]{FF0000} -23.8}                     & \cellcolor[HTML]{FFFFFF}{\color[HTML]{FF0000} -23.1}                     & \cellcolor[HTML]{FFFFFF}{\color[HTML]{FF0000} -24.2}                     & \cellcolor[HTML]{FFFFFF}{\color[HTML]{FF0000} -27.7}                     & \cellcolor[HTML]{FFFFFF}{\color[HTML]{FF0000} -31.4}                     & \cellcolor[HTML]{FFFFFF}{\color[HTML]{FF0000} -22.1}                     & \cellcolor[HTML]{FFFFFF}{\color[HTML]{FF0000} -20.7}                     & \cellcolor[HTML]{FFFFFF}{\color[HTML]{FF0000} -24.8}                      & \cellcolor[HTML]{FFFFFF}{\color[HTML]{FF0000} -16.9}                     & \cellcolor[HTML]{FFFFFF}{\color[HTML]{FF0000} -20.6} & \cellcolor[HTML]{FFFFFF}{\color[HTML]{FF0000} -24.7} \\ \hline 
                                       &                                        & \textbf{Simple DA}    & \cellcolor[HTML]{FFFFFF}{\color[HTML]{1F2328} 49.8}                      & \cellcolor[HTML]{FFFFFF}{\color[HTML]{1F2328} 48.7}                      & \cellcolor[HTML]{FFFFFF}{\color[HTML]{1F2328} 49.4}                      & \cellcolor[HTML]{FFFFFF}{\color[HTML]{1F2328} 52.7}                       & \cellcolor[HTML]{FFFFFF}{\color[HTML]{1F2328} 52.6}                      & \cellcolor[HTML]{FFFFFF}{\color[HTML]{1F2328} 54.7}                      & \cellcolor[HTML]{FFFFFF}{\color[HTML]{1F2328} 52.7}                      & \cellcolor[HTML]{FFFFFF}{\color[HTML]{1F2328} 50.6}                      & \cellcolor[HTML]{FFFFFF}{\color[HTML]{1F2328} 53.9}                      & \cellcolor[HTML]{FFFFFF}{\color[HTML]{1F2328} 53.6}                      & \cellcolor[HTML]{FFFFFF}{\color[HTML]{1F2328} 53.2}                      & \cellcolor[HTML]{FFFFFF}{\color[HTML]{1F2328} 52.2}                      & \cellcolor[HTML]{FFFFFF}{\color[HTML]{1F2328} 51.3}                       & \cellcolor[HTML]{FFFFFF}{\color[HTML]{1F2328} 52.9}                      & \cellcolor[HTML]{FFFFFF}{\color[HTML]{1F2328} 52.0}  & \cellcolor[HTML]{FFFFFF}{\color[HTML]{1F2328} 53.3}  \\ \cline{3-3} 
                                       &                                        & \textbf{Default}        & \cellcolor[HTML]{FFFFFF}{\color[HTML]{1F2328} 64.6}                      & \cellcolor[HTML]{FFFFFF}{\color[HTML]{1F2328} 77.9}                      & \cellcolor[HTML]{FFFFFF}{\color[HTML]{1F2328} 66.5}                      & \cellcolor[HTML]{FFFFFF}{\color[HTML]{1F2328} 77.9}                       & \cellcolor[HTML]{FFFFFF}{\color[HTML]{1F2328} 79.2}                      & \cellcolor[HTML]{FFFFFF}{\color[HTML]{1F2328} 81.0}                      & \cellcolor[HTML]{FFFFFF}{\color[HTML]{1F2328} 78.1}                      & \cellcolor[HTML]{FFFFFF}{\color[HTML]{1F2328} 81.4}                      & \cellcolor[HTML]{FFFFFF}{\color[HTML]{1F2328} 80.2}                      & \cellcolor[HTML]{FFFFFF}{\color[HTML]{1F2328} 79.6}                      & \cellcolor[HTML]{FFFFFF}{\color[HTML]{1F2328} 80.0}                      & \cellcolor[HTML]{FFFFFF}{\color[HTML]{1F2328} 80.2}                      & \cellcolor[HTML]{FFFFFF}{\color[HTML]{1F2328} 78.1}                       & \cellcolor[HTML]{FFFFFF}{\color[HTML]{1F2328} 79.9}                      & \cellcolor[HTML]{FFFFFF}{\color[HTML]{1F2328} 79.3}  & \cellcolor[HTML]{FFFFFF}{\color[HTML]{1F2328} 78.9}  \\ \cline{3-3} 
                                       & \multirow{-3}{*}{\textbf{KITTI}}       & $\Delta$            & \cellcolor[HTML]{FFFFFF}{\color[HTML]{FF0000} -14.8}                     & \cellcolor[HTML]{FFFFFF}{\color[HTML]{FF0000} -29.2}                     & \cellcolor[HTML]{FFFFFF}{\color[HTML]{FF0000} -17.1}                     & \cellcolor[HTML]{FFFFFF}{\color[HTML]{FF0000} -25.2}                      & \cellcolor[HTML]{FFFFFF}{\color[HTML]{FF0000} -26.6}                     & \cellcolor[HTML]{FFFFFF}{\color[HTML]{FF0000} -26.3}                     & \cellcolor[HTML]{FFFFFF}{\color[HTML]{FF0000} -25.4}                     & \cellcolor[HTML]{FFFFFF}{\color[HTML]{FF0000} -30.8}                     & \cellcolor[HTML]{FFFFFF}{\color[HTML]{FF0000} -26.3}                     & \cellcolor[HTML]{FFFFFF}{\color[HTML]{FF0000} -26.0}                     & \cellcolor[HTML]{FFFFFF}{\color[HTML]{FF0000} -26.8}                     & \cellcolor[HTML]{FFFFFF}{\color[HTML]{FF0000} -28.0}                     & \cellcolor[HTML]{FFFFFF}{\color[HTML]{FF0000} -26.8}                      & \cellcolor[HTML]{FFFFFF}{\color[HTML]{FF0000} -27.0}                     & \cellcolor[HTML]{FFFFFF}{\color[HTML]{FF0000} -27.3} & \cellcolor[HTML]{FFFFFF}{\color[HTML]{FF0000} -25.6} \\ \hline 
                                       &                                        & \textbf{Simple DA}    & \cellcolor[HTML]{FFFFFF}{\color[HTML]{1F2328} 15.0}                      & \cellcolor[HTML]{FFFFFF}{\color[HTML]{1F2328} 24.2}                      & \cellcolor[HTML]{FFFFFF}{\color[HTML]{1F2328} 12.8}                      & \cellcolor[HTML]{FFFFFF}{\color[HTML]{1F2328} 21.2}                       & \cellcolor[HTML]{FFFFFF}{\color[HTML]{1F2328} 21.9}                      & \cellcolor[HTML]{FFFFFF}{\color[HTML]{1F2328} 23.4}                      & \cellcolor[HTML]{FFFFFF}{\color[HTML]{1F2328} 18.8}                      & \cellcolor[HTML]{FFFFFF}{\color[HTML]{1F2328} 24.5}                      & \cellcolor[HTML]{FFFFFF}{\color[HTML]{1F2328} 25.1}                      & \cellcolor[HTML]{FFFFFF}{\color[HTML]{1F2328} 26.4}                      & \cellcolor[HTML]{FFFFFF}{\color[HTML]{1F2328} 24.8}                      & \cellcolor[HTML]{FFFFFF}{\color[HTML]{1F2328} 26.9}                      & \cellcolor[HTML]{FFFFFF}{\color[HTML]{1F2328} 25.7}                       & \cellcolor[HTML]{FFFFFF}{\color[HTML]{1F2328} 24.5}                      & \cellcolor[HTML]{FFFFFF}{\color[HTML]{1F2328} 23.4}  & \cellcolor[HTML]{FFFFFF}{\color[HTML]{1F2328} 18.2}  \\ \cline{3-3} 
                                       &                                        & \textbf{Default}        & \cellcolor[HTML]{FFFFFF}{\color[HTML]{1F2328} 17.3}                      & \cellcolor[HTML]{FFFFFF}{\color[HTML]{1F2328} 31.0}                      & \cellcolor[HTML]{FFFFFF}{\color[HTML]{1F2328} 15.2}                      & \cellcolor[HTML]{FFFFFF}{\color[HTML]{1F2328} 33.2}                       & \cellcolor[HTML]{FFFFFF}{\color[HTML]{1F2328} 30.1}                      & \cellcolor[HTML]{FFFFFF}{\color[HTML]{1F2328} 30.7}                      & \cellcolor[HTML]{FFFFFF}{\color[HTML]{1F2328} 24.3}                      & \cellcolor[HTML]{FFFFFF}{\color[HTML]{1F2328} 31.9}                      & \cellcolor[HTML]{FFFFFF}{\color[HTML]{1F2328} 33.8}                      & \cellcolor[HTML]{FFFFFF}{\color[HTML]{1F2328} 34.2}                      & \cellcolor[HTML]{FFFFFF}{\color[HTML]{1F2328} 33.0}                      & \cellcolor[HTML]{FFFFFF}{\color[HTML]{1F2328} 35.7}                      & \cellcolor[HTML]{FFFFFF}{\color[HTML]{1F2328} 38.6}                       & \cellcolor[HTML]{FFFFFF}{\color[HTML]{1F2328} 33.4}                      & \cellcolor[HTML]{FFFFFF}{\color[HTML]{1F2328} 32.8}  & \cellcolor[HTML]{FFFFFF}{\color[HTML]{1F2328} 27.8}  \\ \cline{3-3}
\multirow{-9}{*}{\textbf{Structured}}  & \multirow{-3}{*}{\textbf{sNORB-Azim}}  & $\Delta$            & {\cellcolor[HTML]{FFFFFF}{\color[HTML]{FF0000} -2.3}} & {\cellcolor[HTML]{FFFFFF}{\color[HTML]{FF0000} -6.8}} & {\cellcolor[HTML]{FFFFFF}{\color[HTML]{FF0000} -2.4}} & {\cellcolor[HTML]{FFFFFF}{\color[HTML]{FF0000} -12.0}} & {\cellcolor[HTML]{FFFFFF}{\color[HTML]{FF0000} -8.2}} & {\cellcolor[HTML]{FFFFFF}{\color[HTML]{FF0000} -7.3}} & {\cellcolor[HTML]{FFFFFF}{\color[HTML]{FF0000} -5.5}} & {\cellcolor[HTML]{FFFFFF}{\color[HTML]{FF0000} -7.4}} & {\cellcolor[HTML]{FFFFFF}{\color[HTML]{FF0000} -8.7}} & {\cellcolor[HTML]{FFFFFF}{\color[HTML]{FF0000} -7.8}} & {\cellcolor[HTML]{FFFFFF}{\color[HTML]{FF0000} -8.2}} & {\cellcolor[HTML]{FFFFFF}{\color[HTML]{FF0000} -8.8}} & {\cellcolor[HTML]{FFFFFF}{\color[HTML]{FF0000} -12.9}} & {\cellcolor[HTML]{FFFFFF}{\color[HTML]{FF0000} -8.9}} & \cellcolor[HTML]{FFFFFF}{\color[HTML]{FF0000} -9.4}  & \cellcolor[HTML]{FFFFFF}{\color[HTML]{FF0000} -9.6}  \\ \hline
\end{tabular}}
\caption{\small We apply simple data augmentations (DA) (RandomResizedCrop, RandomVerticalFlip and RandomHorizontalFlip) on three datasets in each group. Data augmentation does not benefit most of VTAB-1K datasets and thus, most recent PEFT papers~\citep{jie2023fact, convpass,luo2023rep, jia2022visual, zhang2022neural, luo2023rep, ssf} skip it.  \autoref{fig:image_transformations_vertical} shows examples of how some data augmentation transforms are harmful for a specific task. Therefore, to ensure that our results are directly comparable to existing papers, we don't apply data augmentation. }
\label{tab:simple_da}
\end{table*}

\begin{figure}[htbp]
    \centering
    \begin{subfigure}[b]{0.4\textwidth}
        \includegraphics[width=\textwidth]{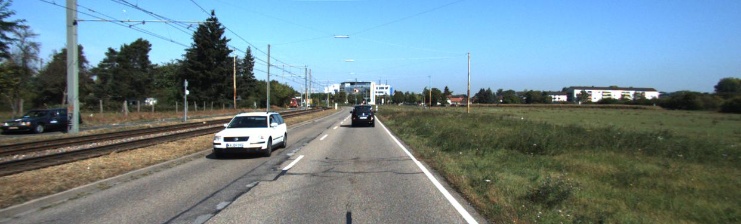}
        \caption{Original}
        \label{fig:original}
    \end{subfigure}
    \vskip\baselineskip
    \begin{subfigure}[b]{0.4\textwidth}
        \includegraphics[width=\textwidth]{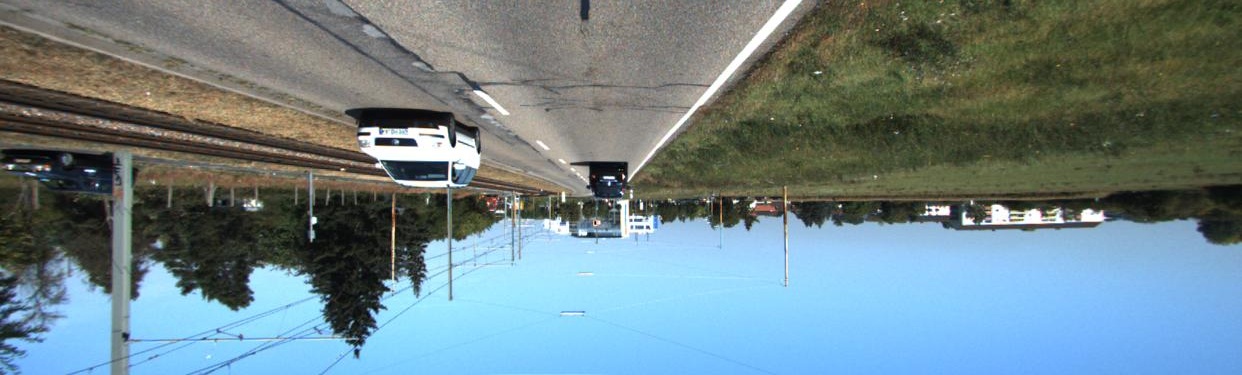}
        \caption{Vertical Flipped}
        \label{fig:flipped}
    \end{subfigure}
    \vskip\baselineskip
    \begin{subfigure}[b]{0.128\textwidth}
        \includegraphics[width=\textwidth]{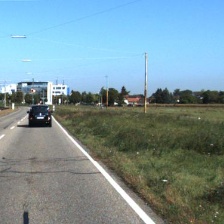}
        \caption{Cropped}
        \label{fig:cropped}
    \end{subfigure}
    \hspace{2em}
    \begin{subfigure}[b]{0.128\textwidth}
        \includegraphics[width=\textwidth]{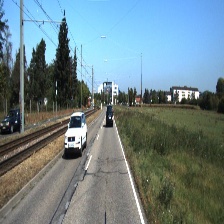}
        \caption{Resized}
        \label{fig:resized}
    \end{subfigure}
    \caption{Example of augmented images from KITTI: (a) Original, (b) RandomVerticalFlip, (c) RandomResizedCrop, (d) Resize only. The KITTI task needs to predict the depth to the nearest vehicle (car, van, or truck) in the image.  RandomResizedCrop may crop out the nearest vehicle. RandomVerticalFlip may make the task more difficult. }
    \label{fig:image_transformations_vertical}
\end{figure}

\section{Background}
\label{-sec: survey}
\begin{table*}
\centering
\begin{tabular}{cl}

\hline Symbol (Abbreviation) & Definition \\
\hline
$(H, W)$ & Resolution of input images\\
$C$ & Number of channels (input images)\\
$P$ & Resolution of patches\\
$N$ & Number of patches (tokens) \\
$N_h$ & Number of head in each Transformer layer\\
$D$ & Embedding dimension\\
$D_h$ & Embedding dimension for single-head attention\\
$L_m$ & $m$-th Transformer layer\\
$M$ & Number of Transformer layer\\
$Z_{m-1}$ & Input of $m$-th Transformer layer\\

\hline
ViT & Vision Transformer\\
LN & Layer Normalization \\
MSA & Multi-head Self-Attention\\
MLP & Multi-Layer Perceptron\\
FC &  Fully-connected layer\\
\hline
\end{tabular}
\caption{Definitions of symbols and abbreviation used in \autoref{-sec: survey}}
\end{table*}

\subsection{Vision Transformer}
\label{sec: vit}
\mypara{Overview of ViT.} Inspired by the recent success of Transformer-based models~\cite{vaswani2017attention} in NLP~\cite{wolf2020transformers}, Vision Transformer (ViT)~\cite{dosovitskiy2020image} has become widely used in computer vision. To handle 2D images, ViT divides an image $\mI \in \R^{H\times W \times C}$ into $N$ non-overlapping patches $\{\mI^{(n)} \in \R^{P^2 \times C}\}_{n=1}^N$, where $(H, W)$ is the resolution of the input image, $C$ is the number of channels, $N = HW / P^2$ and $(P, P)$ is the resolution of each patch. Each patch $\mI^{(n)}$ is flattened and embedded into a $D$-dimensional vector $\vx^{(n)}_0$ with a trainable linear projection. Incorporating the BERT design approach~\cite{kenton2019bert}, a ``Class'' token $\vx_0^{(\text{Class})}$ is prepended to the sequence of embedded patches, whose output state at the last Transformer layer is utilized as the image representation. Finally, position embeddings $\mathbf{E}_{\text {pos }} \in \mathbb{R}^{{D\times(1+N)}}$ are added to preserve positional information and form the input $\mathbf{Z}_0 \in \mathbb{R}^{D\times(1+N)} $ to the ViT, which can be formulated by: 

\begin{equation}
\mZ_0=\left[\vx_0^{(\text{Class})}, \vx^{(1)}_0, \vx^{(2)}_0, \cdots ,\vx^{(N)}_0 \right]+\mathbf{E}_{\text {pos }}
\end{equation}

\begin{figure}[th]
\centering
    \includegraphics[width=1\linewidth]{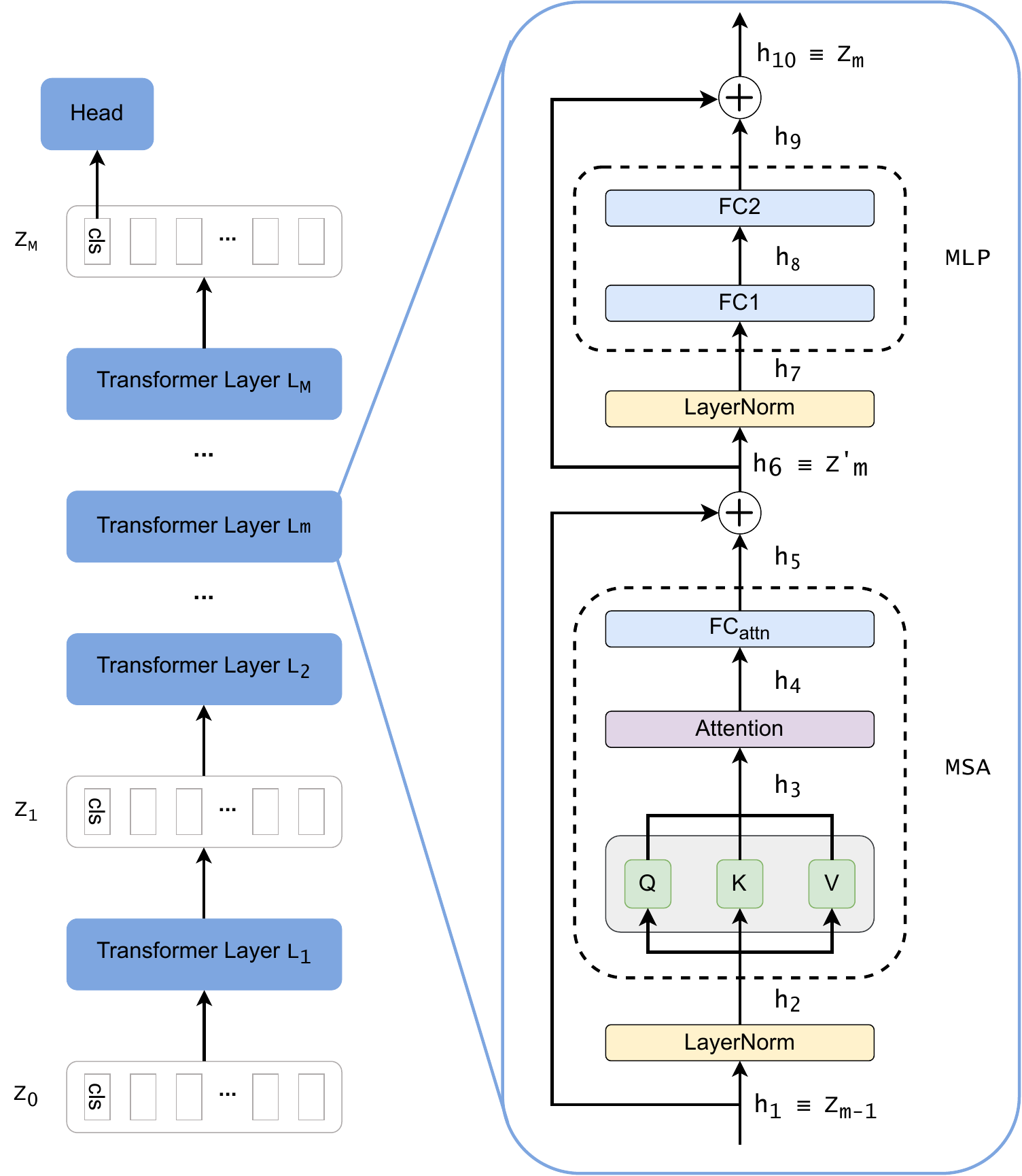}
    \caption{An overview of a Transformer block in ViT. We adopt the notation $\{h_1, \ldots, h_{10}\}$ to denote the intermediate features within a Transformer block to facilitate a clearer illustration of the PEFT methods discussed in \autoref{-sec: method}. }
    \label{fig:vit}
\end{figure}

As shown in the left part of \autoref{fig:vit}, a ViT typically consists of $M$ layers, denoted by $\{L_m\}_{m=1}^M$. The input $\mZ_0$ mentioned above is fed into the first layer $L_1$, producing the output $\mZ_1 = L_1(\mZ_0)=[\vx_1^{(\text{Class})}, \vx^{(1)}_1, \cdots, \vx^{(N)}_1]\in\R^{D\times(1+N)}$, which maintains the same size as $\mZ_0$. Namely, $\mZ_1$ comprises $1+N$ feature tokens, and each corresponds to the same column in $\mZ_0$. Similarly, for $m = 2, \cdots, M$, each layer $L_m$ takes the output of the previous layer as input and generates the output, $\mZ_m = L_m(\mZ_{m-1})$. Finally, the ``Class'' vector $\vx_M^{(\text{Class})}$ in $\mZ_M$ serves as  the image feature for prediction. When dealing with classification tasks, the predicted label $\hat{y} = \cst{Head}(\vx_M^{(\text{Class})})$ is generated through a linear head (\ie, a fully-connected layer).


\mypara{Details of each Transformer layer.} As shown in the right part of \autoref{fig:vit}, each Transformer layer consists of a Multi-head Self-Attention (MSA) block, a Multi-Layer Perceptron (MLP) block, and two Layer Normalization (LN) layers~\cite{ba2016layer}. Formally, a Transformer layer $L_m$ can be defined as

\begin{equation}
\begin{aligned}
\mZ_{m}^{\prime} & =\operatorname{MSA}\left(\operatorname{LN}\left(\mZ_{m-1}\right)\right)+\mZ_{m-1} \\
\mZ_{m} & =\operatorname{MLP}\left(\operatorname{LN}\left(\mZ_{m}^{\prime}\right)\right)+\mZ_{m}^{\prime}
\end{aligned}
\end{equation}
where $\mZ_{m-1}= [\vx_{m-1}^{(\text{Class})}, \vx^{(1)}_{m-1}, \cdots, \vx^{(N)}_{m-1}]\in\R^{D\times(1+N)}$ is the output of the preceding $(m-1)$-th Transformer layer. The MLP is applied to each column vector of $\mZ_{m}^{\prime}$ independently.

In order to encapsulate multiple complex relationships amongst different elements in the sequence, the MSA block comprises $N_h$ single-head self-attention blocks. For the $i^{th}$ single-head self-attention block, an generic input $\mZ$ is first projected into three matrices, namely Query $\mQ^{(i)}$, Key $\mK^{(i)}$, and Value $\mV^{(i)}$

\begin{align}
\mQ^{(i)} = \mW^{(i)}_Q \mZ, \quad \mK^{(i)} = \mW^{(i)}_K \mZ, \quad \mV^{(i)} = \mW_V^{(i)} \mZ, 
\end{align}
 where $\mW^{(i)}_{Q/K/V} \in \mathbb{R}^{D_h \times D}$ \footnote{For brevity, we ignore the layer index $m$ for the projection matrices $\mW_Q, \mW_K, \mW_V$, but each layer has its own projection matrices.} where $D_h$ is the embedding dimension for a single head self-attention block and typically set to $D/N_h$.  The $i^{th}$ self-attention head in MSA is formulated as 

 \begin{equation}
\operatorname{Attn}^{(i)}(\mZ)=\mV^{(i)} \times \operatorname{Softmax}\left(\frac{\left.  {\mK^{(i)}}^\top \mQ^{(i)}\right)}{\sqrt{D_h}}\right) \in \mathbb{R}^{D_h\times(1+N)}
\end{equation}

The outputs of all heads are concatenated and linearly projected by a fully connected layer ($FC_{attn}$) with weight $\mW_{O} \in \mathbb{R}^{D \times (D_h \cdot N_h)}$ as the output of the MSA block.

\begin{equation}
\operatorname{MSA}(\mZ)=\mW_{O} \left[\operatorname{Attn}^0(\mZ), \ldots, \operatorname{Attn}^{N_h}(\mZ)\right] 
\end{equation}

The MLP block can be defined as 

\begin{equation}
\operatorname{MLP}(\mZ)=GELU\left(\mZ \mW_1+b_1\right) \mW_2+b_2
\end{equation}

where $\mW_1 \in \mathbb{R}^{D \times 4D}$\footnote{4 is the MLP ratio in ViT-B}, $\mW_2 \in \mathbb{R}^{4D \times D}$, $b_1 \in \mathbb{R}^{4D}$, $b_2 \in \mathbb{R}^D$ are weights and biases for two FC layers ($FC_1$ and $FC_2$) respectively. 

Since PEFT methods often entail incorporating additional components to modify the intermediate features within or between Transformer layers, we adopt the notation $\{h_1, \ldots, h_{10}\}$ to denote the intermediate features in the unravelled view of a Transformer layer (as depicted in \autoref{fig:vit}) to facilitate a clearer illustration of the PEFT methods discussed in the subsequent section. 

\subsection{Evaluated Methods}
\label{-sec: method}
In this section, we dive into the details of 12 state-of-the-art PEFT approaches, categorized into three groups: Prompt-based, Adapter-based, and Selective Parameter Tuning. We will describe the distinctions and tradeoffs between them. A consolidated overview of these approaches is summarized in \autoref{table:summary}. 

\newcommand{\tabitem}{~~\llap{\textbullet}~~}

\begin{table*}[th!]
\footnotesize
        \caption{PEFT Methods Summary: Prompt-based and adapter-based methods incorporate additional parameters to modify features while keeping the original backbone intact. However, these added parameters introduce additional inference overhead. In contrast, selective tuning methods modify the backbone by updating selective parameters, thereby incurring no additional inference overhead. }
        \bgroup
\def\arraystretch{2}%

\resizebox{\textwidth}{!}{%
    \begin{tabular}{|c|Sc|Sc|c|c|c|}
        \hline
        Method & What & \thead{Tunable \\Parameters} & \thead{Hyper \\Parameters } & \thead{Modified \\ Type} &  \thead{Inference \\Efficient} \\
        \hline
        VPT-Deep& $\boldsymbol{h_1}=[\boldsymbol{h_1}, \boldsymbol{P}]$ & $\boldsymbol{P} \in \mathbb{R}^{l \times D}$& $l$: Number of prompts&  Feature& \XSolid \\
        \hline
        AdaptFormer& $\boldsymbol{h_{9}}= \boldsymbol{h_9}+ \operatorname{Adapter} \boldsymbol{(h_7)}$&$\boldsymbol{W}_{\text {down }/\text {up}} \in \mathbb{R}^{r \times D /D \times r}$ in $\operatorname{Adapter}$& \makecell{$s$: Scale factor in $\operatorname{Adapter} $\\ $r$: Bottleneck dimension} &Feature& \XSolid \\
        \hline
        
        Pfeif. Adapter& $\boldsymbol{h_9} = \operatorname{Adapter}(\boldsymbol{h_9}) $&$\boldsymbol{W}_{\text {down } /\text {up}} \in \mathbb{R}^{r \times D/D \times r}$ in $\operatorname{Adapter}$ &\makecell{$s$: Scale factor in  $\operatorname{Adapter}$ \\ $r$: Bottleneck dimension}&Feature& \XSolid\\
        \hline

        Houl. Adapter& \makecell{$\boldsymbol{h_5} = \operatorname{Adapter_1}(\boldsymbol{h_5})$\\ $\boldsymbol{h_9} = \operatorname{Adapter_2}(\boldsymbol{h_9})$}&\makecell{$\boldsymbol{W}^1_{\text {down } /\text {up}} \in \mathbb{R}^{r \times D /D \times r}$ in $\operatorname{Adapter_1}$ \\$\boldsymbol{W}^2_{\text {down } /\text {up}} \in \mathbb{R}^{ r\times D/D \times r}$ in $\operatorname{Adapter_2}$  }&\makecell{$s$: Scale factor in $\operatorname{Adapter}$\\ $r$: Bottleneck dimension}&Feature& \XSolid\\ 
        \hline

        Convpass&\makecell{$\boldsymbol{h_5} = \operatorname{Convpass_1}(\boldsymbol{h_2}) + \boldsymbol{h_5}$\\ $\boldsymbol{h_9} = \operatorname{Convpass_2}(\boldsymbol{h_7}) + \boldsymbol{h_9}$}& \makecell{ \makecell[l]{$\boldsymbol{W}^1_{conv2d} \in \mathbb{R}^{r \times r \times k \times k}$\\ $\boldsymbol{W}^1_{\text {down } / \text {up}} \in \mathbb{R}^{r \times D/ D \times r}$} in $\operatorname{Convpass_1}$\\\makecell[l]{$\boldsymbol{W}^2_{conv2d} \in \mathbb{R}^{r \times r \times k \times k}$\\ $\boldsymbol{W}^2_{\text {down } / \text {up}} \in \mathbb{R}^{r \times D/ D \times r}$} in $\operatorname{Convpass_2}$  } &\makecell{$s$: Scale factor in $\operatorname{Convpass}$\\ $r$: Bottleneck dimension \\ $k$: Kernel size of $conv2d$}&Feature& \XSolid\\ 
        \hline

        RepAdpater& \makecell{$\boldsymbol{h_2} = \operatorname{RepAdapter_1}(\boldsymbol{h_2})$\\ $\boldsymbol{h_7} =  \operatorname{RepAdapter_2}(\boldsymbol{h_7})$}& \makecell{ \makecell[l]{$\boldsymbol{W}^1_{conv1d} \in \mathbb{R}^{r \times D}$\\ $\boldsymbol{b}^1 \in \mathbb{R}^{r}$} in $\operatorname{RepAdapter_1}$\\\makecell[l]{$\boldsymbol{W}^2_{conv1d} \in \mathbb{R}^{D \times \frac{r}{G} }$\\ $\boldsymbol{b}^2 \in \mathbb{R}^{D}$} in $\operatorname{RepAdapter_2}$} &\makecell{$s$: Scale factor in $\operatorname{RepAdapter}$\\ $r$: Bottleneck dimension \\ $G$: Number of groups}&Feature& \XSolid\\
        \hline

        LayerNorm &  \makecell{$\boldsymbol{h_2} = \operatorname{LayerNorm_1}(\boldsymbol{h_1})$\\ $\boldsymbol{h_7} = \operatorname{LayerNorm_2}(\boldsymbol{h_6})$}&\makecell{
            $\boldsymbol{W}^{1(2)},\boldsymbol{b}^{1(2)}  \in \mathbb{R}^{D}$ in $\operatorname{LayerNorm_{1(2)}}$}& N/A&Backbone& \Checkmark\\ 
        \hline
        
        BitFit& \makecell{Fine-tune all bias terms \\ in the network} &\makecell[l]{
         $\boldsymbol{b}^{1(2)}  \in \mathbb{R}^{D}$ in $\operatorname{LayerNorm_{1(2)}}$\\
         $\boldsymbol{b}^{Q/K/V}  \in \mathbb{R}^{D}$ in $\operatorname{Q/K/V}$\\
         $\boldsymbol{b}^{FC_{attn}}  \in \mathbb{R}^{D}$ in $\operatorname{FC_{attn}}$\\
         $\boldsymbol{b}^{1}  \in \mathbb{R}^{4D},$ in $\operatorname{FC_1}$, $\boldsymbol{b}^{2}  \in \mathbb{R}^{D}$ in $\operatorname{FC_2}$
         } & N/A&Backbone& \Checkmark\\
        \hline
        
        \multirow{2}{*}{DiffFit} & \tabitem LayerNorm + BitFit&\tabitem \makecell{All tunable parameters \\in LayerNorm \& BitFit} & \multirow{2}{*}{N/A}&\multirow{2}{*}{Backbone}& \multirow{2}{*}{\Checkmark}\\
        &\tabitem \makecell{$\boldsymbol{h_5} = \gamma_1 \cdot \boldsymbol{h_5}$\\ $\boldsymbol{h_{9}} = \gamma_2 \cdot \boldsymbol{h_9}$}&\tabitem $\gamma_1, \gamma_2 \in \mathbb{R}^D$& &&\\
        \hline

        SSF&\makecell{$\boldsymbol{h_2} = \operatorname{SSF_2}(\boldsymbol{h_2})$, {} $\boldsymbol{h_3} = \operatorname{SSF_3}(\boldsymbol{h_3})$\\$\boldsymbol{h_5} = \operatorname{SSF_5}(\boldsymbol{h_5})$, {} $\boldsymbol{h_7} = \operatorname{SSF_7}(\boldsymbol{h_7})$\\$\boldsymbol{h_8} = \operatorname{SSF_7}(\boldsymbol{h_8})$, {} $\boldsymbol{h_9} = \operatorname{SSF_9}(\boldsymbol{h_9})$}& \makecell[l]{$\boldsymbol{W}^{2,5,7,9} \in \mathbb{R}^{D}, \boldsymbol{b}^{2,5,7,9}  \in \mathbb{R}^{D}$\\  $\boldsymbol{W}^{3} \in \mathbb{R}^{3D}, \boldsymbol{b}^{3}  \in \mathbb{R}^{3D}$\\ $\boldsymbol{W}^{8} \in \mathbb{R}^{4D}, \boldsymbol{b}^{8}  \in \mathbb{R}^{4D}$} & N/A&Backbone& \Checkmark\\
        \hline 
        
        LoRA&$\boldsymbol{h_3} = \operatorname{LoRA}(\boldsymbol{h_2}) + \boldsymbol{h_3}$& $\boldsymbol{W}_{\text {down} / \text {up}}^{Q/K/V} \in \mathbb{R}^{r \times D / D\times r} $ in $\operatorname{LoRA}$& $r$: Bottleneck dimension&Backbone& \Checkmark \\
        \hline 
        
        \multirow{2}{*}{FacT\textsubscript{TT(TK)}}& \multirow{2}{*}{\shortstack{$\boldsymbol{h_3} = \operatorname{FacT_{TT(TK)}}(\boldsymbol{h_2}) + \boldsymbol{h_3}$\\ $\boldsymbol{h_5} = \operatorname{FacT_{TT(TK)}}(\boldsymbol{h_4})+ \boldsymbol{h_5}$ \\$\boldsymbol{h_8} = \operatorname{FacT_{TT(TK)}}(\boldsymbol{h_7})+ \boldsymbol{h_8}$\\ $\boldsymbol{h_9} = \operatorname{FacT_{TT(TK)}}(\boldsymbol{h_8})+ \boldsymbol{h_9}$}}& \shortstack[l]{$\boldsymbol{U} \in \mathbb{R}^{D \times r}, \boldsymbol{V} \in \mathbb{R}^{D \times r}$, \\ $\boldsymbol{\Sigma} \in \mathbb{R}^{12 L \times r \times r}$ in $\operatorname{FacT_{TT}}$}&\multirow{2}{*}{\shortstack{$s$: Scale factor in $\operatorname{FacT_{TT(TK)}}$\\ $r$: Bottleneck dimension}}&\multirow{2}{*}{Backbone}& \multirow{2}{*}{\Checkmark}\\
        \cline{3-3}
        &&\shortstack[l]{$\boldsymbol{U} \in \mathbb{R}^{D \times r}, \boldsymbol{V} \in \mathbb{R}^{D \times r}$, \\ $\boldsymbol{A} \in \mathbb{R}^{12 L \times r},  \boldsymbol{B} \in \mathbb{R}^{r \times r \times r}$ in $\operatorname{FacT_{TK}}$}&&&\\
        \hline

    \end{tabular}
    }
    \egroup

    \label{table:summary}
\end{table*}

\subsubsection{Prompt-based Methods}
Prompt-based learning emerged in NLP as an effective approach to adapt pre-trained models for downstream tasks~\cite{promptnlpsurvey, petlnkp}. The core concept involves augmenting the model input with task-specific hints (prompts), which aid the pre-trained model in addressing novel tasks with its existing knowledge. Hard prompts are human-interpretable natural language hints, encompassing task instructions, in-context examples, or supporting information. Alternatively, soft prompts are continuous vector hints that are incorporated into the input embeddings of the input layers or hidden states of other layers. Soft prompts are updated during the fine-tuning process using gradient-based methods, guided by the downstream task-specific loss functions, while the pre-trained model itself remains fixed. The splendent success of prompts in NLP has sparked a growing interest in adopting it in computer vision~\cite{yu2023visualtuning, tu2023visual} and multi-modal domains~\cite{promptVLsurvey}.

In this paper, we investigate a prominent and strong prompt-based method called \textbf{Visual Prompt Tuning (VPT)}~\cite{jia2022visual}, which represents one of the early endeavours in introducing prompts to computer vision. Specifically, VPT-Shallow adds $l$ prompts $\mP_0 \in \R^{l \times D}$ to the input of the first Transformer layer $\mZ_0$ and the output $\tilde{\mP_0}$ of $\mP_0$ serves as the input for the next layer as depicted in \autoref{eq: vpt-s}. VPT-Shallow can be perceived as the addition of learnable pixels to the original images. On the other hand, VPT-Deep inserts $l$ prompts $\{\mP_m \in \R^{l \times D}\}_{m=0}^{M}$ to the input of every Transformer layer $\mZ_m$ but their outputs are discarded at the end of the layer as illustrated in \autoref{eq: vpt-d}. 

\begin{equation}
\label{eq: vpt-s}
\begin{aligned}
[\tilde{\mP_1}, \mZ_1] &= L_m([\mP_0, \mZ_{0}])\\
[\tilde{\mP_m}, \mZ_m] &= L_m([\tilde{\mP}_{m-1}, \mZ_{m-1}]) \quad m=2,3, \ldots, M\\
\end{aligned}
\end{equation}

\begin{equation}
\label{eq: vpt-d}
[\_, \mZ_m] = L_m([{\mP}_{m-1}, \mZ_{m-1}]) \quad m=1,2,3, \ldots, M\\
\end{equation}

Throughout the adaptation process, the pre-trained model is frozen and no additional weights are introduced to the model, thereby preserving the model's original behaviour. During the forward pass, the output $\mZ_m$ of layer $m$ is changed because of the interaction between $\mZ_{m-1}$ and $\mP_{m-1}$ (or $\tilde{\mP}_{m-1}$) in the MSA block. Thus, the output feature is adapted to the downstream tasks by iteratively tuning the prompts through gradient descent.

\subsubsection{Adapter-based Methods}
\label{-sec: adapter}
Adapter-based methods typically introduce additional trainable parameters into a frozen pre-trained model to facilitate learning of downstream tasks~\cite{petlnkp}. Initially developed for multi-domain adaptation~\cite{rebuffi2017learning, rebuffi2018efficient} and continual learning~\cite{rosenfeld2018incremental, mai2022online}, the idea of Adapters is subsequently embraced by Houlsby \etal~\cite{houlsby2019parameter} in the NLP domain to adapt Transformer-based networks for downstream tasks, and it also has garnered increasing interest in the computer vision field~\cite{yu2023visualtuning}. In this comparative analysis, we concentrate on\emph{five} popular Adapter-based methods, encompassing the original Adapter, along with variants focusing on adjusting the positions of Adapters~\cite{chen2022adaptformer, pfeiffer2021adapterfusion}, introducing visual inductive biases~\cite{convpass}, as well as employing re-parameterization to reduce the number of trainable parameters and inference latency~\cite{luo2023rep}.

\begin{figure*}
\centering
  \begin{subfigure}{0.2\textwidth}
  \centering
    \includegraphics[width=\textwidth]{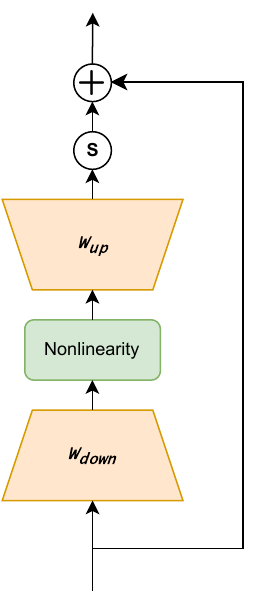}
    \caption{Adapter}
    \label{fig:adapter}
  \end{subfigure}
  \begin{subfigure}{0.2\textwidth}
  \centering
    \includegraphics[width=0.515\textwidth]{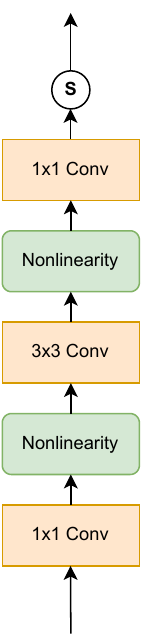}
    \caption{Convpass}
    \label{fig:convpass}
  \end{subfigure}
  \begin{subfigure}{0.2\textwidth}
  \centering
    \includegraphics[width=\textwidth]{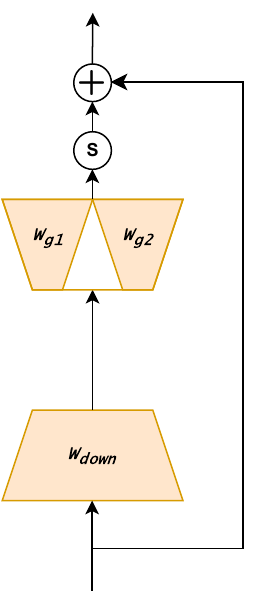}
    \caption{RepAdapter}
    \label{fig:repadapter}
  \end{subfigure}
    \caption{Comparison of three Adapter structures. }
\end{figure*}

\subparagraph{Houl. Adapter}~\cite{houlsby2019parameter} inserts two lightweight bottleneck-structured modules into each Transformer layer: one after the MSA block and the other after the MLP block. As depicted in \autoref{fig:adapter}, the Adapter is composed of a down-projection layer with $\boldsymbol{W}_{\text {down }} \in \mathbb{R}^{r \times D} $, a nonlinear activation function $\sigma$, an up-projection layer with $\boldsymbol{W}_{\text {up}} \in \mathbb{R}^{D \times r} $, a scaling factor $s$ and a skip-connection.  To limit the number of trainable parameters, the bottleneck dimension is much smaller than the feature dimension $r \ll D$. Formally, Houl. Adapter can be defined as:

\begin{gather}
\label{eq: h-adapter}
h_5 = \operatorname{Adapter_1}(h_5) \quad h_9 = \operatorname{Adapter_2}(h_9) \\
\operatorname{Adapter}(h) = s \cdot \mW_{\text {up}} \sigma(\mW_{\text {down }}h) + h
\label{eq: adapter}
\end{gather}

\subparagraph{Pfeif. Adapter}~\cite{pfeiffer2021adapterfusion} is a more efficient variant that introduces the Adapter solely after the MLP block, a strategy that has demonstrated effectiveness in recent studies~\cite{hu2021lora}. Pfeif. Adapter can be defined formally as $h_9 = \operatorname{Adapter}(h_9)$ where $\operatorname{Adapter}$ follows \autoref{eq: adapter}. 

\subparagraph{AdaptFormer}~\cite{chen2022adaptformer} proposed to insert the Adapter in parallel with the MLP block, which differs from the sequential design of Houl. and Pfeif. Adapter. The rationale behind this parallel design lies in the belief that the domain-specific features generated by the Adapter can complement the domain-agnostic features derived from the original MLP block, leading to an improved feature ensemble~\cite{inception}.  Formally, AdaptFormer can be defined as $\boldsymbol{h_{9}}= \boldsymbol{h_9}+ \operatorname{Adapter} \boldsymbol{(h7)}$ where $\operatorname{Adapter}$ follows \autoref{eq: adapter}.

\subparagraph{ConvPass}(\underline{Conv}olutional By-\underline{Pass}es)~\cite{convpass} addresses the concern that many existing Adapters lack visual inductive bias, potentially limiting their performance for downstream vision tasks with limited data. To this end, the authors introduce a convolutional bottleneck module, running in parallel with the MSA or(and) MLP block. This module encompasses a $1 \times 1$ convolution reducing the channel with $\boldsymbol{W}_{\text {down }} \in \mathbb{R}^{r \times D} $, a $3 \times 3$ convolution with the same input and output channel, a $1 \times 1$ convolution expanding the channel $\boldsymbol{W}_{\text {up}} \in \mathbb{R}^{D \times r} $, two nonlinear functions $\sigma$ and a scaling factor $s$, as shown in \autoref{fig:convpass}. The authors argue that Convpass is more efficient at capturing visual information in low-data scenarios due to its hard-coded locality of convolutional layers. The formal definition of Convpass is shown in \autoref{eq: convpass}. 
\begin{equation}
\begin{gathered}
\label{eq: convpass}
h_5 = \operatorname{Convpass_1}(h_2) + h_5 \quad h_9 = \operatorname{Convpass_2}(h_7) + h_9 \\
\operatorname{Convpass}(h) = s \cdot \mW_{\text {up}}\sigma(\operatorname{Conv2d}(\sigma(\mW_{\text {down }}h)))
\end{gathered}
\end{equation}

\subparagraph{RepAdapter}~\cite{luo2023rep} found that the removal of the nonlinear function in the Adapter does not result in performance degradation for vision tasks. In light of this finding, the authors propose a linear Adapter with group-wise transformation~\cite{luo2022towards} and sequentially added two of these linear Adapters to both MSA and MLP blocks. Owing to the sequential placement of the RepAdapter and its inherent linearity, the additional parameters can be re-parameterized to the original MSA or MLP block after training, thereby incurring zero additional costs during inference. RepAdapter is illustrated in \autoref{fig:repadapter} and formally defined in \autoref{eq: repadapter}. 

\begin{equation}
\begin{gathered}
\label{eq: repadapter}
h_5 = \operatorname{RepAdapter_1}(h_2)  \quad h_7 = \operatorname{RepAdapter_2}(h_7) \\
\operatorname{RepAdapter}(h) = s \cdot \phi_{\text {up}}(\phi_{\text {down }}(h)) + h\\
\tilde{h} = \phi_{\text {down }}(h) =\boldsymbol{W}_{\text {down }} h  \\
\phi_{\text {up}}(\tilde{h}) = [\boldsymbol{W}_{g1}\tilde{h}_{g1}, \ldots, \boldsymbol{W}_{gG}\tilde{h}_{gG}]\\
\end{gathered}
\end{equation}

where $ \boldsymbol{W}_{\text {down }} \in \mathbb{R}^{r \times D} $, $\tilde{h}_{g(1, \ldots,G)} \in \mathbb{R}^{ \frac{r}{G}\times (N+1)  } $ is the features splitted from $\tilde{h} \in \mathbb{R}^{r \times (N+1) }$ and $G$ is the number of groups in group-wise transformation~\cite{luo2022towards}. $\boldsymbol{W}_{g(1, \ldots,G)} \in \mathbb{R}^{\frac{D}{G} \times \frac{r}{G}  }$ is the projection weight matrix. 

\subsubsection{Selective Parameter Tuning  Methods}
The methods falling within this category aim to selectively update the parameters of a pre-trained model for downstream tasks. Within transfer learning, two prominent strategies, namely \textit{full fine-tuning} and \textit{linear probing}~\cite{kornblith2019better, zhuang2020comprehensive}, represent the two extremes of this category. \textit{Full fine-tuning} updates all the model parameters end-to-end based on the new dataset while \textit{linear probing} treats the pre-trained model as a feature extractor and only updates the prediction heads while keeping the backbone frozen. Although \textit{full fine-tuning} generally exhibits superior performance compared to \textit{linear probing}~\cite{zhai2019large}, it possesses certain limitations that may hinder its practicality in real-world production settings. Firstly, it requires running gradient descent for all parameters and necessitates storing a separate fine-tuned model for each task, incurring significant computational, memory, and storage overhead. These challenges become more salient with Transformer-based models whose parameters grow exponentially. Secondly, \textit{full fine-tuning} may distort pre-trained features and underperform \textit{linear probing} in out-of-distribution (OOD) scenarios~\cite{kumar2021fine}. 

To cope with the above issues, a cohort of PEFT methods has emerged under this category. In addition to the two common approaches mentioned above, our investigation encompasses seven methods that can be further categorized into two groups: direct selective tuning~\cite{zaken2022bitfit, ln-tune, xie2023difffit}, which involves the direct modification of selective weights, and efficient selective tuning~\cite{hu2021lora, jie2023fact, ssf}, which approximates the weight updates with low-rank factors. 

Notably, an extra advantage of methods in this category is that they introduce \textbf{no additional inference latency}, making them particularly favourable when inference efficiency is a priority. Methods within the direct selective tuning group abstain from introducing any new modules, thus inherently avoiding extra inference latency. Meanwhile, for methods in the efficient selective tuning group, the added modules can often be seamlessly integrated into weights of the pre-trained models through the re-parameterization techniques~\cite{jacob2018quantization, ding2021repvgg}, thereby ensuring the absence of increased inference latency as well.

\paragraph{Direct Selective Tuning}
\paragraph{BitFit}~\cite{zaken2022bitfit} is a simple yet effective method that only tunes the bias parts of the pre-trained model. For each Transformer layer in ViT, BitFit updates the bias terms in the QKV projections and the FC layer in the MSA block, two FC layers in the MLP block and two LN blocks. It also updates the bias in the projection for patch embedding. The original authors underscore BitFit's capability to achieve performance comparable to full fine-tuning or even surpass it under low and medium-data scenarios in BERT models~\cite{kenton2019bert}.

\subparagraph{LayerNorm}~\cite{ln-tune} represents another simple but strong baseline that solely tunes the two LN blocks in each Transformer layer - one before the MSA block and another before the MLP block. Given that each LN block contains merely two trainable parameters $\{\mW_{LN}, \vb_{LN}\} \in \mathbb{R}^{D}$, LN-tune stands out as an exceedingly light-weight approach compared to other PEFT methods. For instance, ViT-B/16 ($\sim$86M parameters) has only $\sim$38K LN parameters, accounting for $\sim$0.04\% of the total parameters. 

\subparagraph{DiffFit}~\cite{xie2023difffit} is a recently proposed PEFT strategy designed for adapting large pre-trained diffusion models to the new domains. DiffFit exclusively fine-tunes the bias terms and the LN blocks within the network. Furthermore, it inserts learnable scale factors $\gamma$ to shift the features after the MSA and the MLP blocks, as shown in \autoref{eq: difffit}. Consequently, DiffFit can be regarded as a combination of the BitFit and Ln-Tune, incorporating additional feature shift factors.

\begin{equation}
\begin{gathered}
\label{eq: difffit}
{h_5} = \gamma_1 \cdot {h_5}\\ 
{h_{9}} = \gamma_2 \cdot {h_9}
\end{gathered}
\end{equation}

\paragraph{Efficient Selective Tuning}
\subparagraph{LoRA}(\underline{Lo}w-\underline{R}ank \underline{A}daptation)~\cite{hu2021lora} drew inspiration from recent investigations demonstrating that the learned over-parametrized models in fact reside on a low intrinsic dimension~\cite{li2018measuring, aghajanyan-etal-2021}. Building upon this insight, the authors hypothesize that the change in weights during model adaptation also exhibits a low intrinsic rank and injects trainable low-rank decomposition matrices to approximate the weight updates. The LoRA update methodology is strategically applied to the Query/Value projection weights $\mW_{Q/V} \in \mathbb{R}^{D \times D}$ within the MSA block. Concretely, the weight updates are approximated as $\mW_{Q/V} + \Delta\mW_{Q/V} = \mW_{Q/V} + \mW^{Q/V}_{\text{down}}\mW^{Q/V}_{\text{up}}$ where $\mW^{Q/V}_{\text{down}/\text{up}} \in \mathbb{R}^{D \times r/r \times D}$ and rank $r \ll D$. The authors use a random Gaussian initialization for $\mW^{Q/V}_{\text{up}}$ and zero for $\mW^{Q/V}_{\text{down}}$ so that $\Delta\mW_{Q/V} = \mW^{Q/V}_{\text{down}}\mW^{Q/V}_{\text{up}}$ is zero at the beginning of training. The formal definition of LoRA is articulated in \autoref{eq: lora}, utilizing the notations delineated in \autoref{fig:vit}.  



\begin{equation}
\begin{gathered}
\label{eq: lora}
{h_3} = \operatorname{LoRA}({h_2}) + {h_3}\\
h_3 = [\mQ, \mK, \mV]\\
\operatorname{LoRA}({h_2}) = [\mW^{Q}_{\text{down}}\mW^{Q}_{\text{up}}h_2, 0, \mW^{V}_{\text{down}}\mW^{V}_{\text{up}}h_2]
\end{gathered}
\end{equation}






\subparagraph{FacT}(\underline{Fac}tor \underline{T}uning)~\cite{jie2023fact} is inspired by the recent advances in Transformer compression~\cite{wang2022exploring, zhang2022minivit}and exploited the low-rank update paradigm (e.g., LoRA) to the extreme. While LoRA posits that the update for an individual weight matrix manifests a low-rank characteristic during fine-tuning, FacT advances the proposition that the weight updates spanning different matrices can also be effectively approximated using low-rank decomposition matrices. Specifically, FacT encapsulates the four weight matrices $\mW_{Q/K/V/O} \in \mathbb{R}^{D \times D}$ in the MSA block and the two weight matrices $\mW_1 \in \mathbb{R}^{D \times 4D}$, $\mW_2 \in \mathbb{R}^{4D \times D}$ in the MLP block into a single $\mW_{FacT} \in \mathbb{R}^{12M\times D\times D}$ tensor where $M$ is the number of Transformer layer. The update of $\mW_{FacT}$, $\Delta \mW_{FacT}$, can be decomposed into several factors to promote parameter efficiency. To this end, the authors leverage the well-established Tensor-Train (TT)~\cite{tt}and the Tucker (TK)~\cite{tk} format to decompose $\Delta \mW_{FacT}$. $\text{FacT}_{\text{TT}}$ and $\text{FacT}_{\text{TK}}$ are used to denote different decomposition formats for FacT and their formal definitions can be found in \autoref{eq: fact}.


\begin{flalign}
\label{eq: fact}
&\text{FacT}_{\text{TT}}: \Delta \mW_{FacT} = s \cdot \boldsymbol{\Sigma} \times_2 \mU^{\top} \times_3 \mV^{\top}\\
&\text{FacT}_{\text{TK}}: \Delta \mW_{FacT} = s \cdot \mA \times_1 \mB^{\boldsymbol{\top}} \times_2 \mU^{\boldsymbol{\top}} \times_3 \mV^{\boldsymbol{\top}}
\end{flalign}
where $\mU \in \mathbb{R}^{D \times r}, \mV \in \mathbb{R}^{D \times r}, \boldsymbol{\Sigma} \in \mathbb{R}^{12 L \times r \times r}, \mB \in \mathbb{R}^{12L \times r} \mA \in \mathbb{R}^{ r \times r \times r}$ and the $\times_{j}$ denotes mode-$j$ product and $s$ is the scaling factor. 

Since $\Delta \mW_{FacT}$ contains the updates for $\mW_{Q/K/V/O}, \mW_{1/2}$, the modified forward pass inherently influences $h_3, h_5, h_8, h_9$. Let's consider $h_5$ for elucidation.  Once the weight update $\Delta \mW_{FacT}$ is calculated with $\text{FacT}_{\text{TT(TK)}}$ in \autoref{eq: fact}, the corresponding update for $\mW_O$, $\Delta \mW_{O}$, is extracted from $\Delta \mW_{FacT}$. Similar to the modified forward pass of LoRA,  $h_5 = h_4 \Delta \mW_{O} + h_5$.

\subparagraph{SSF}(\underline{S}cale \& \underline{S}hift deep \underline{F}eatures)~\cite{ssf} employs linear transformations to adapt the intermediate features extracted by a pre-trained model. Motivated by the feature modulation methods~\cite{huang2017arbitrary, perez2018film}, SSF is designed to accommodate the distribution difference between the upstream and downstream datasets. Specifically, SSF modulates the features residing at $h_2, h_3, h_5, h_7, h_8, h_9$ by incorporating scale and shift factors. To demonstrate the mechanism of SSF, let's consider $h_5 \in \mathbb{R}^{(N+1) \times D}$ as an illustrative example and other features can similarly undergo the same transformative process. Formally, the modulated $h_5$ is formulated as follows.

\begin{equation}
\begin{gathered}
\label{eq: ssf}
h_5 = \operatorname{SSF_5}(h_5) =  \vw^5 \odot h_5 + \vb^5 
\end{gathered}
\end{equation}
where $\vw^5\in \mathbb{R}^{D}$, $\vb^5 \in \mathbb{R}^{D}$ are the scale and shift factors affiliated with the SSF module attributed to $h_5$, and $\odot$ is the dot product. It is noteworthy that each modulated feature has its own SSF module with corresponding scale and shift factors. The modification details for other features are summarized in \autoref{table:summary}.

\section{More Detailed Results}
\label{-sec:results}

\begin{figure}
\centering
  \includegraphics[width=.9\linewidth]{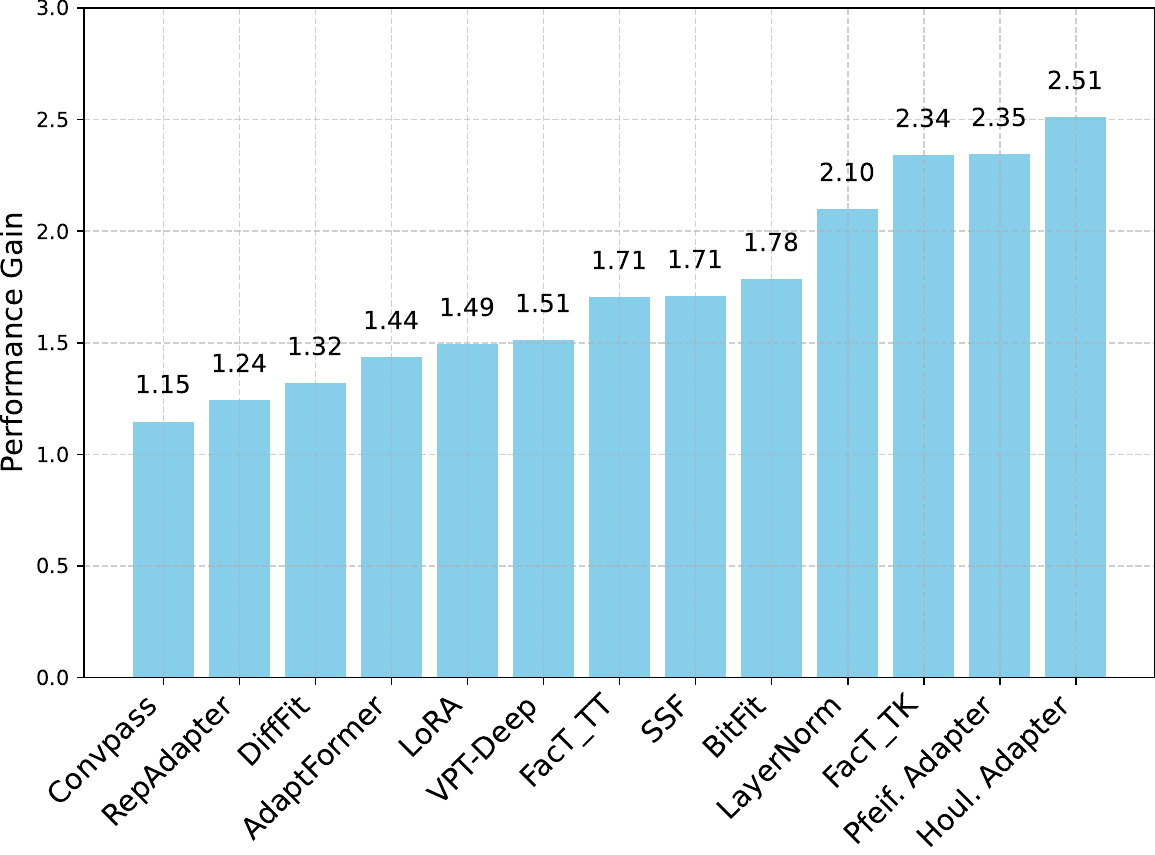}
  \caption{\small Performance gain for PEFT methods by turning drop-path-rate on. }
  \label{fig:dpr}
\end{figure}

\paragraph{Drop-path-rate. } Learning with low-shot data is prone to over-fitting. We find that if the drop path rate --- which stochastically drops a transformer block per sample~\cite{huang2016deep} --- is set not as default (\ie, nonzero), all the methods can benefit from such a regularization. \autoref{fig:dpr} shows the performance gain by tuning the drop-path-rate on compared with the default 0.

\paragraph{More results on prediction similarity analysis. }

\begin{figure*}
    \centering
    \begin{subfigure}{0.22\linewidth}
    \includegraphics[width=\linewidth]{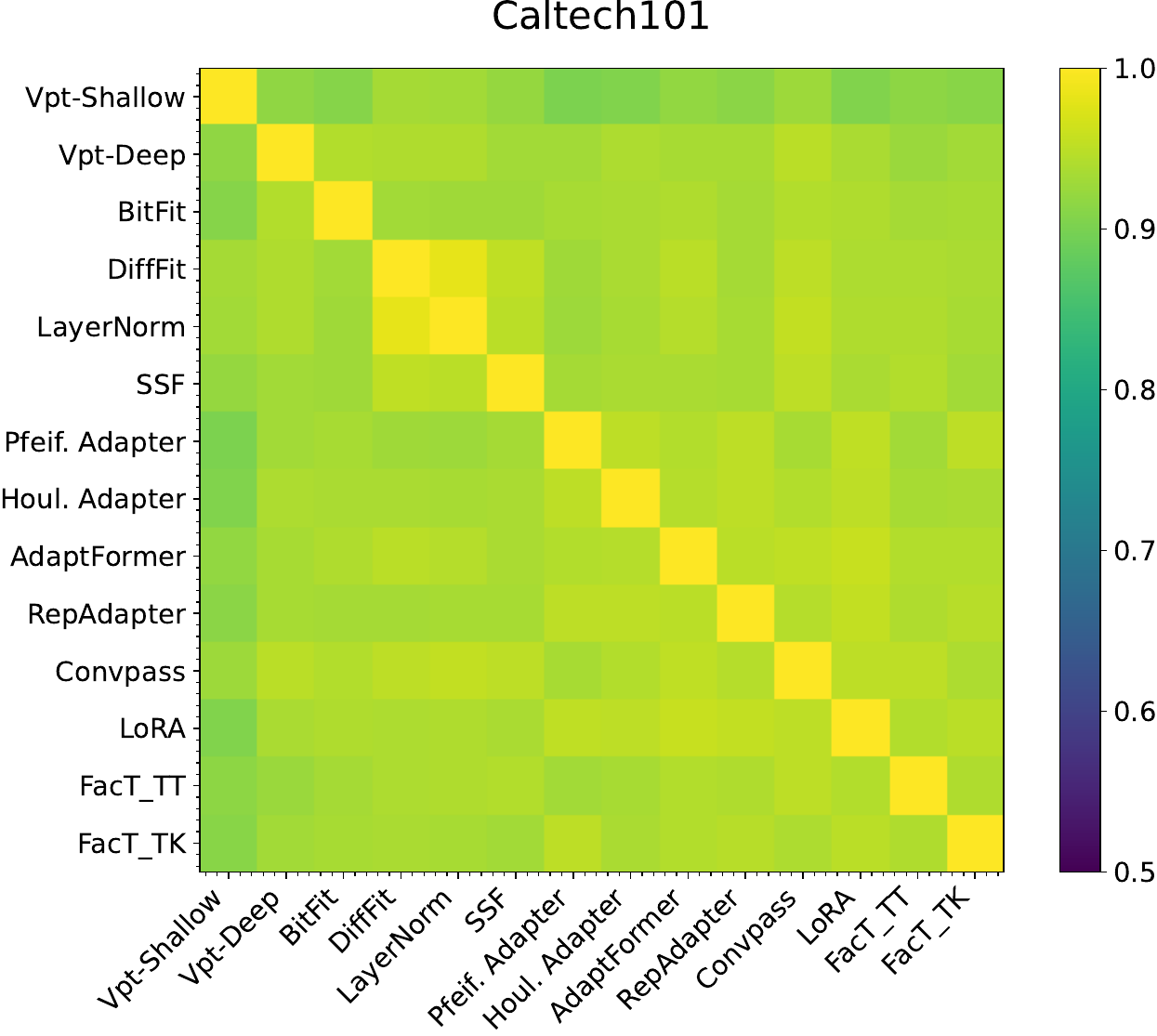}
    \caption{Caltech101}
    \end{subfigure}%
    \hfill
    \begin{subfigure}{0.22\linewidth}
    \includegraphics[width=\linewidth]{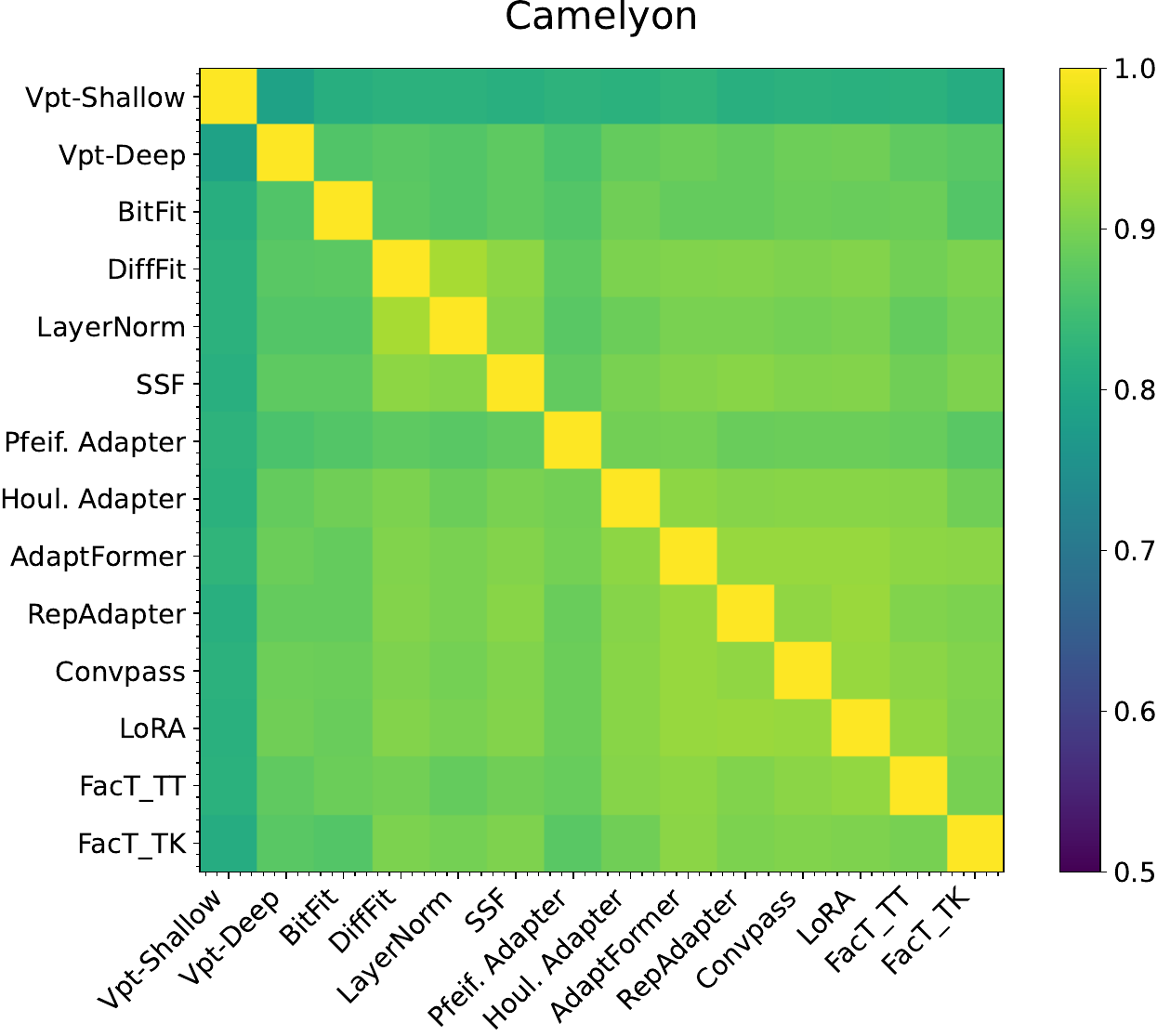}
    \caption{Camelyon}
    \end{subfigure}%
    \hfill
    \begin{subfigure}{0.22\linewidth}
    \includegraphics[width=\linewidth]{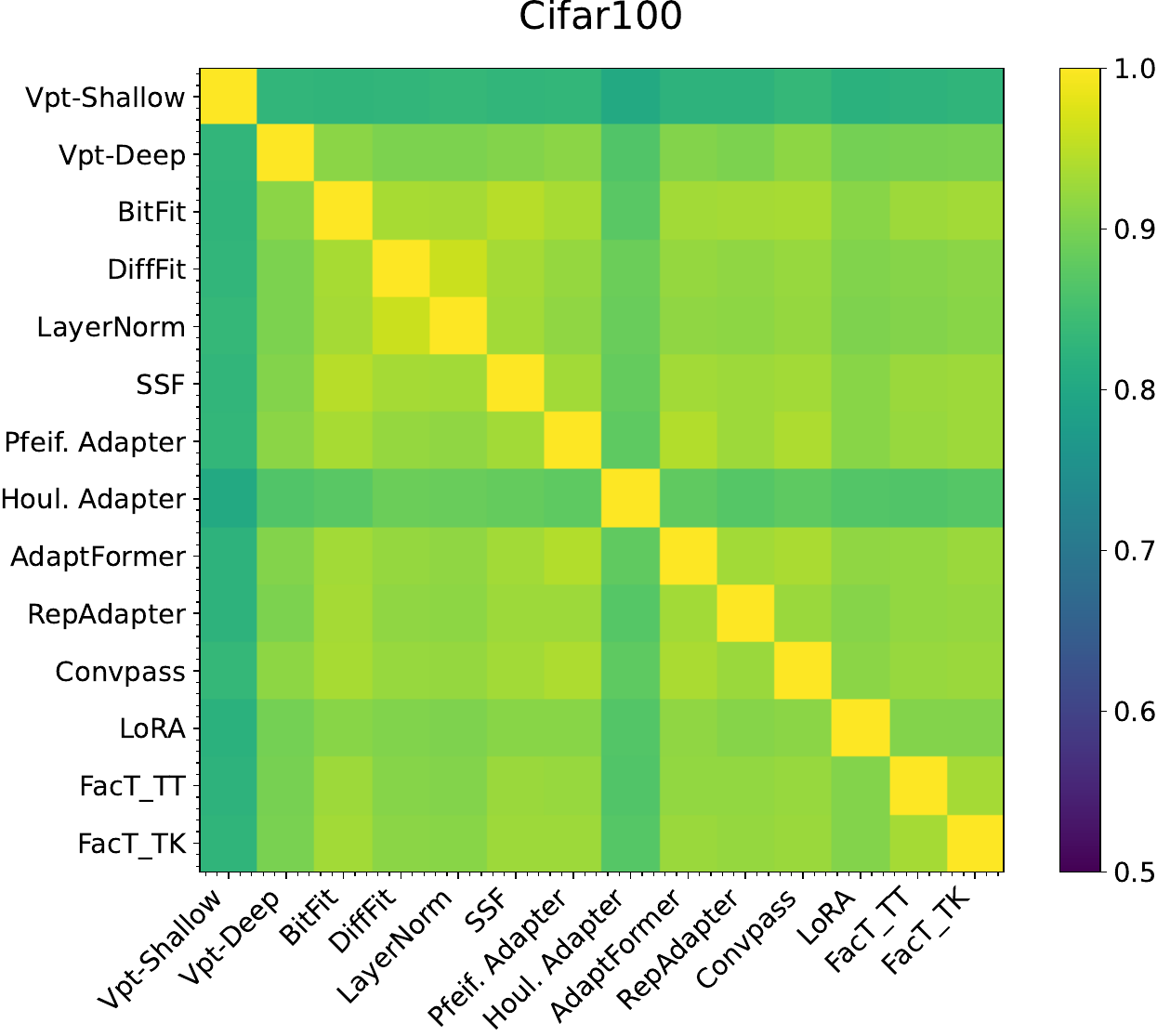}
    \caption{Cifar100}
    \end{subfigure}%
    \hfill
    \begin{subfigure}{0.22\linewidth}
    \includegraphics[width=\linewidth]{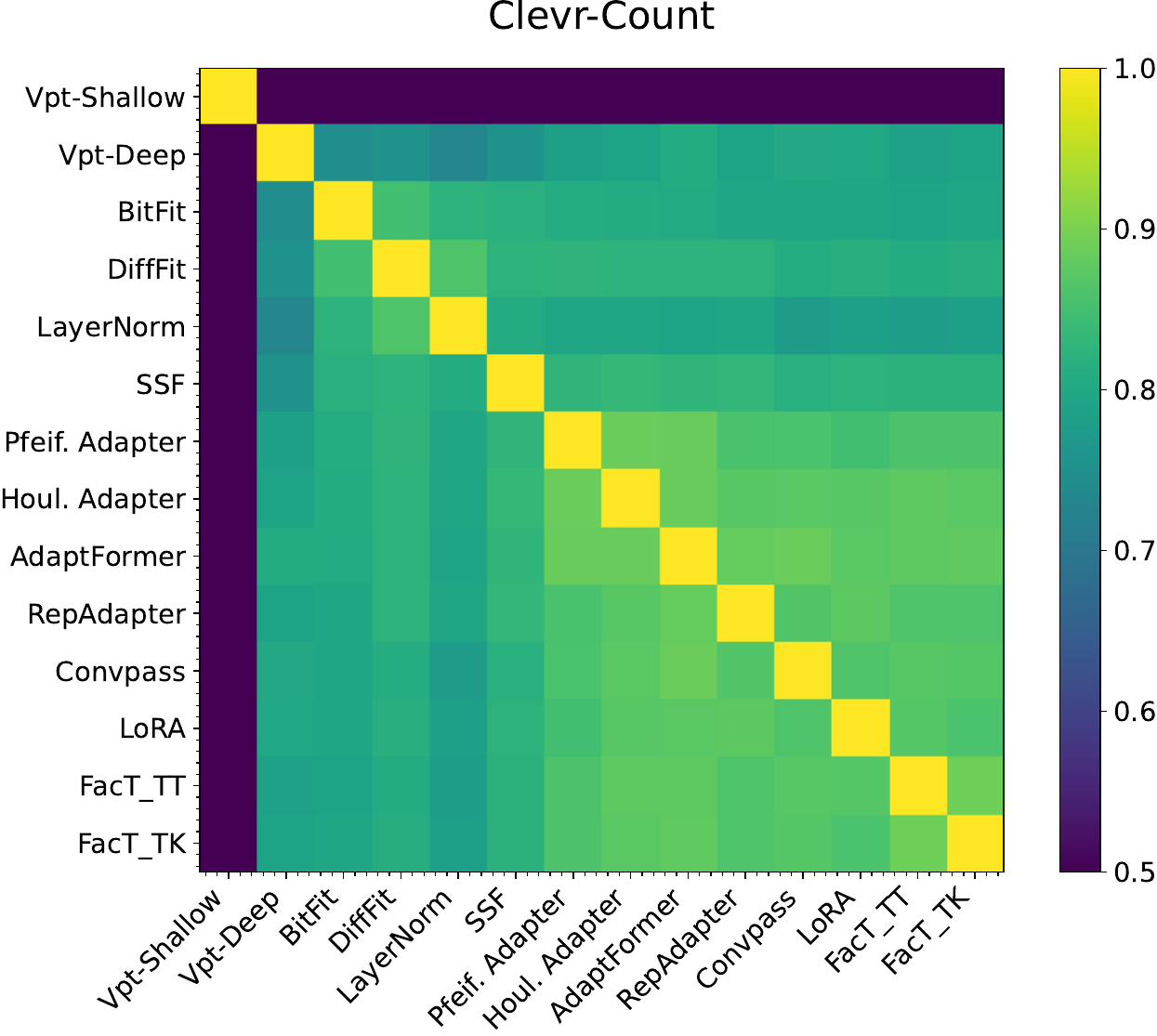}
    \caption{Clevr-Count}
    \end{subfigure}%
    
    \begin{subfigure}{0.22\linewidth}
    \includegraphics[width=\linewidth]{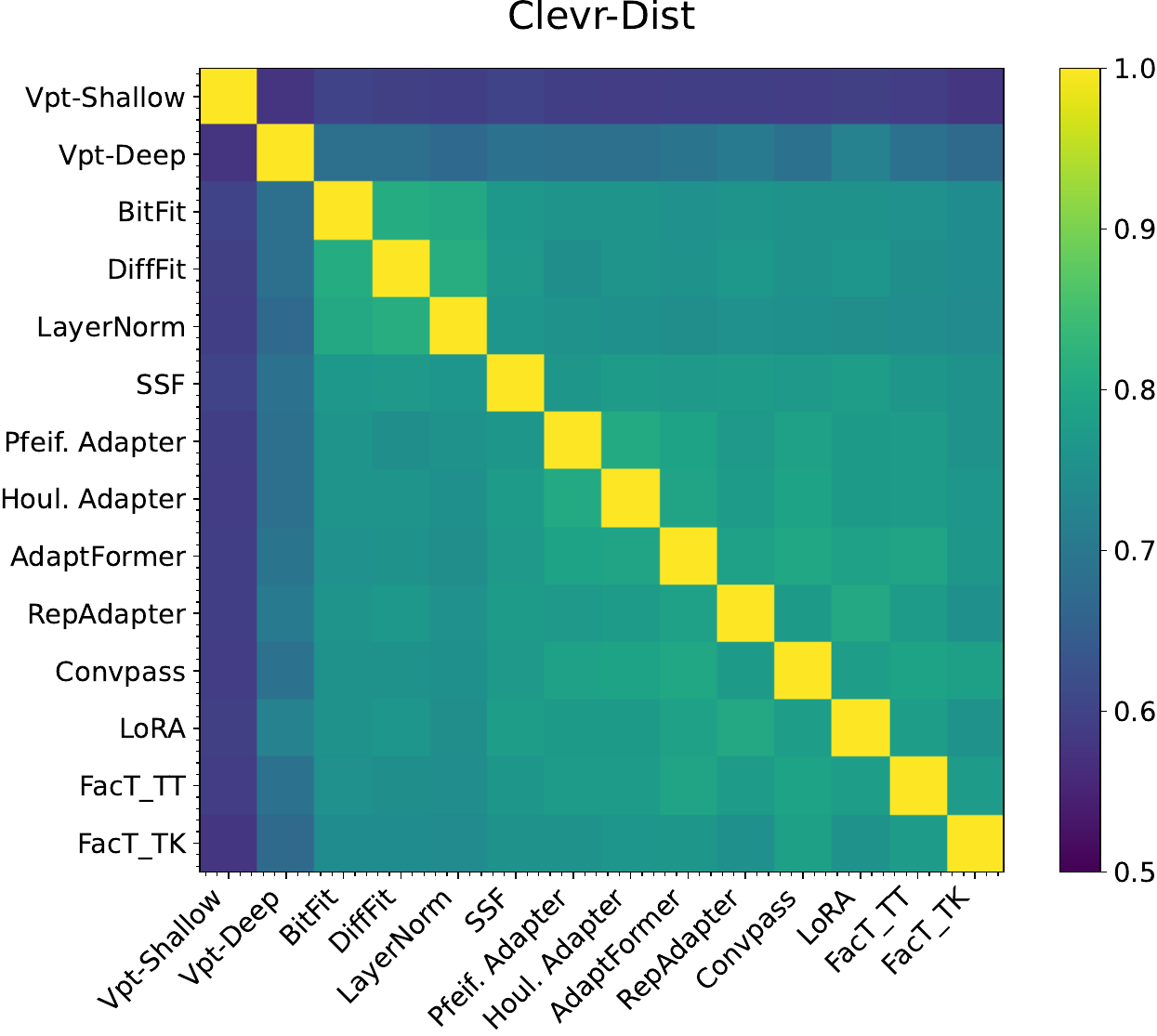}
    \caption{Clevr-Dist}
    \end{subfigure}%
    \hfill
    \begin{subfigure}{0.22\linewidth}
    \includegraphics[width=\linewidth]{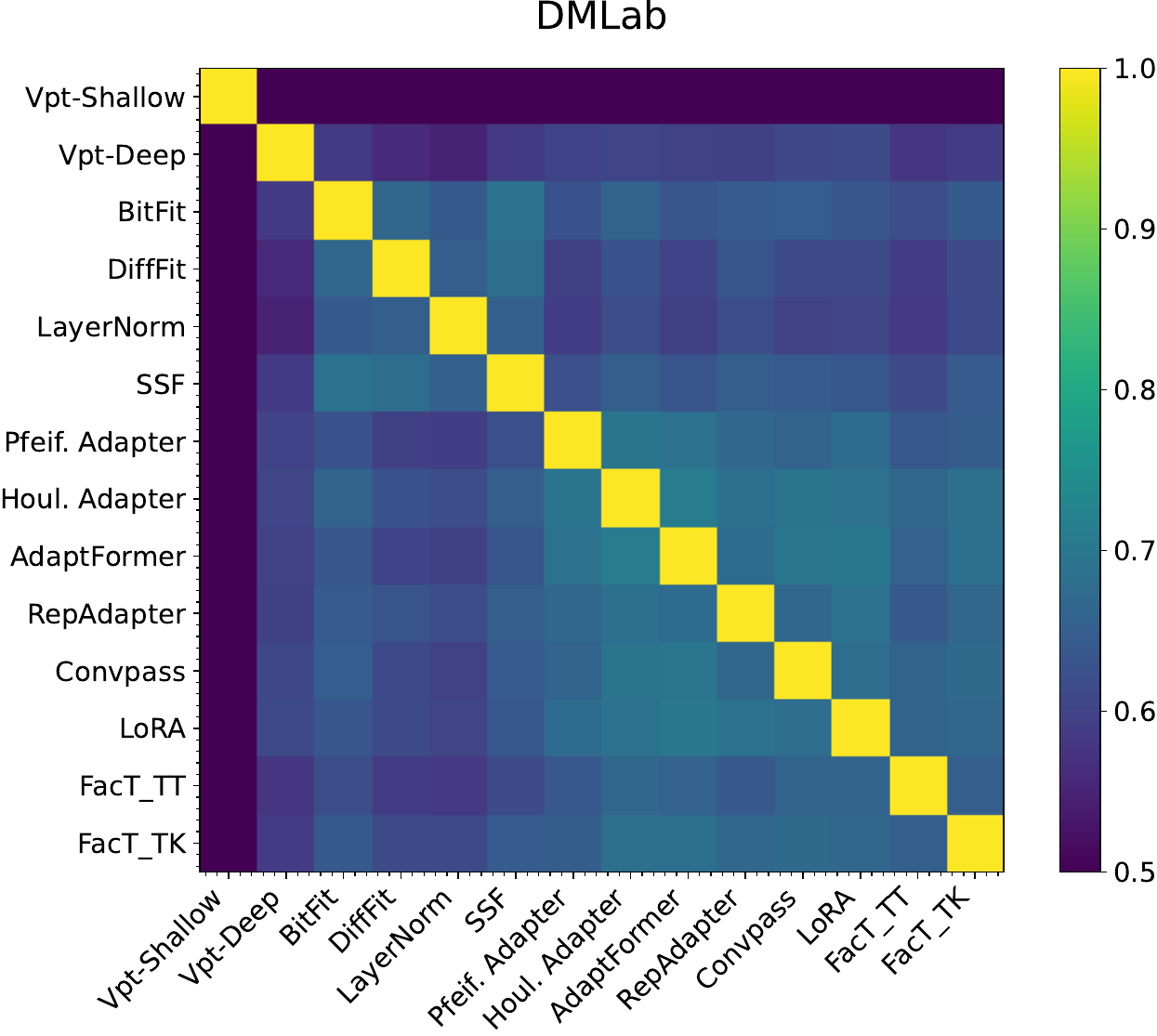}
    \caption{DMLab}
    \end{subfigure}%
    \hfill
    \begin{subfigure}{0.22\linewidth}
    \includegraphics[width=\linewidth]{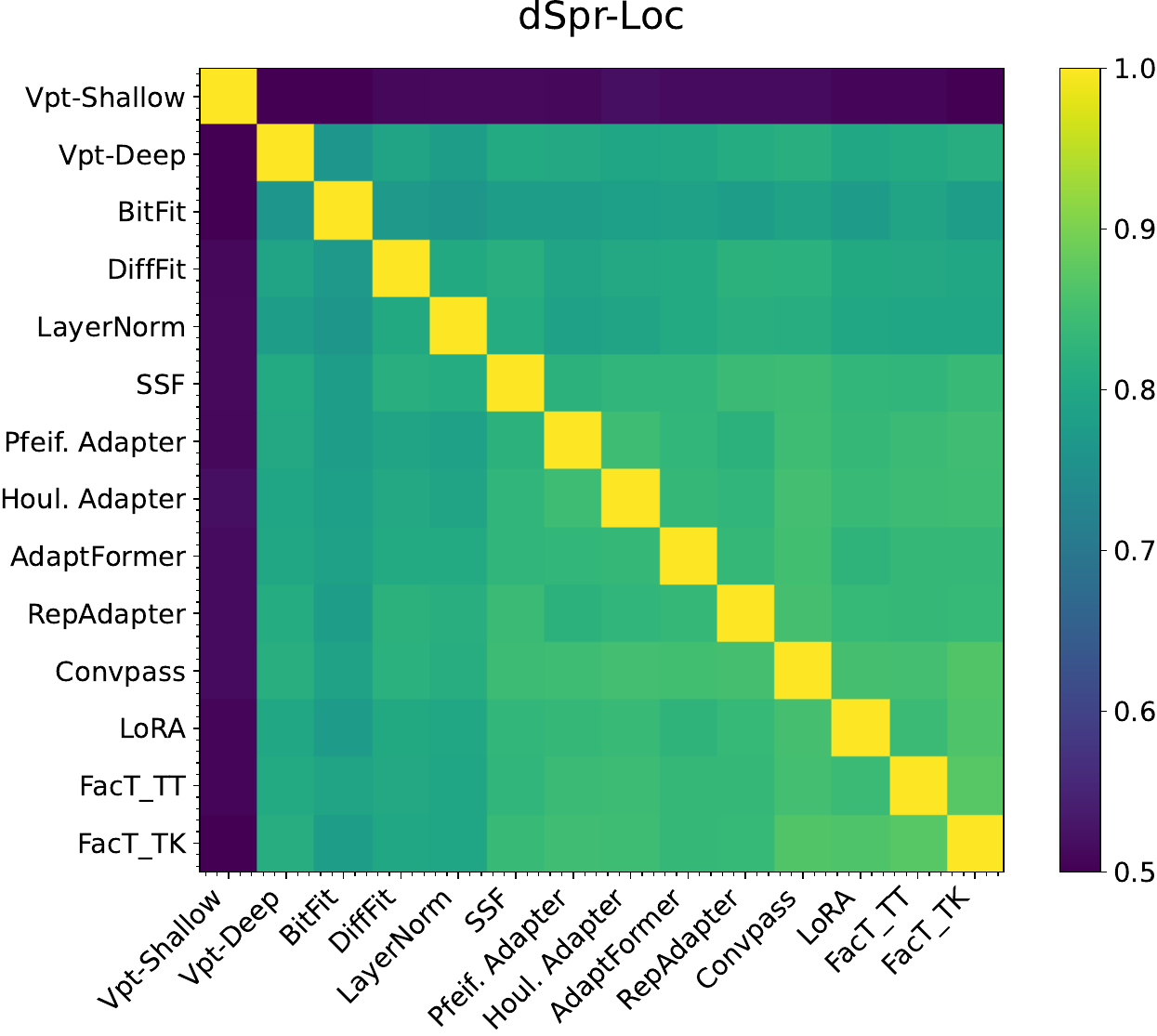}
    \caption{dSpr-Loc}
    \end{subfigure}%
    \hfill
    \begin{subfigure}{0.22\linewidth}
    \includegraphics[width=\linewidth]{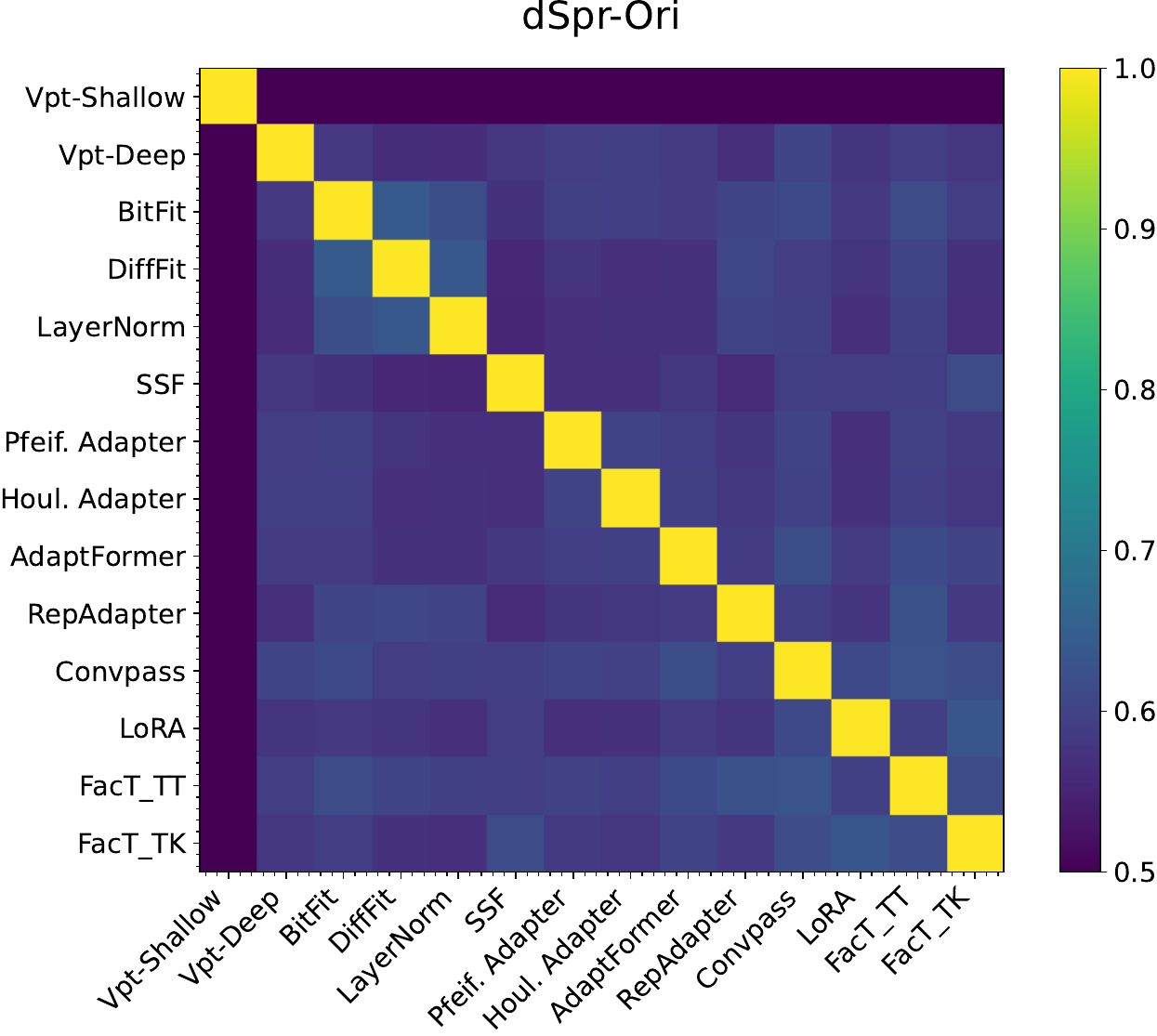}
    \caption{dSpr-Ori}
    \end{subfigure}%

    \begin{subfigure}{0.22\linewidth}
    \includegraphics[width=\linewidth]{figures/pred_sim/dSpr-Ori.pdf}
    \caption{dSpr-Ori}
    \end{subfigure}%
    \hfill
    \begin{subfigure}{0.22\linewidth}
    \includegraphics[width=\linewidth]{figures/pred_sim/dSpr-Ori.pdf}
    \caption{dSpr-Ori}
    \end{subfigure}%
    \hfill
    \begin{subfigure}{0.22\linewidth}
    \includegraphics[width=\linewidth]{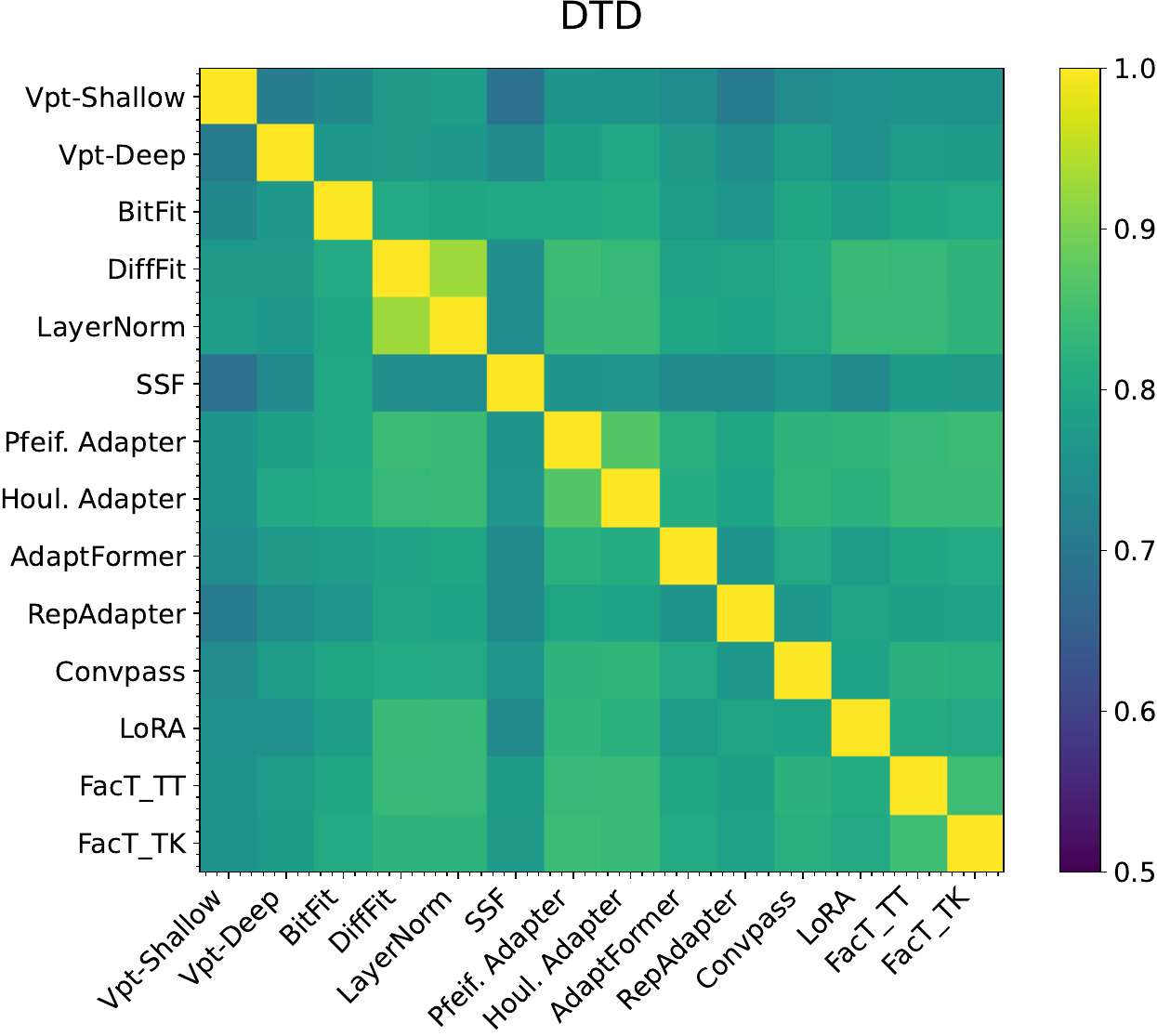}
    \caption{DTD}
    \end{subfigure}%
    \hfill
    \begin{subfigure}{0.22\linewidth}
    \includegraphics[width=\linewidth]{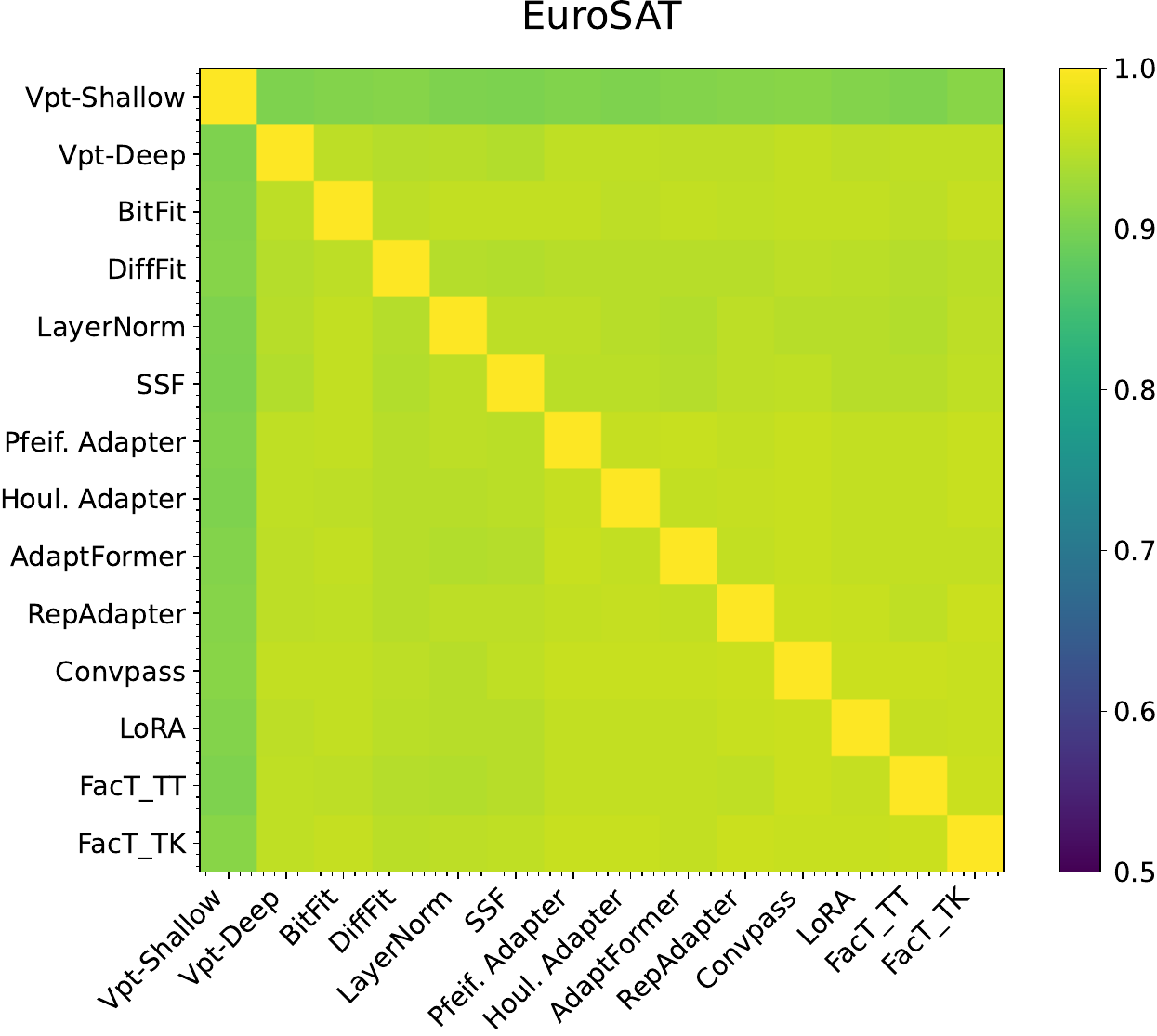}
    \caption{EuroSAT}
    \end{subfigure}%

    \begin{subfigure}{0.22\linewidth}
    \includegraphics[width=\linewidth]{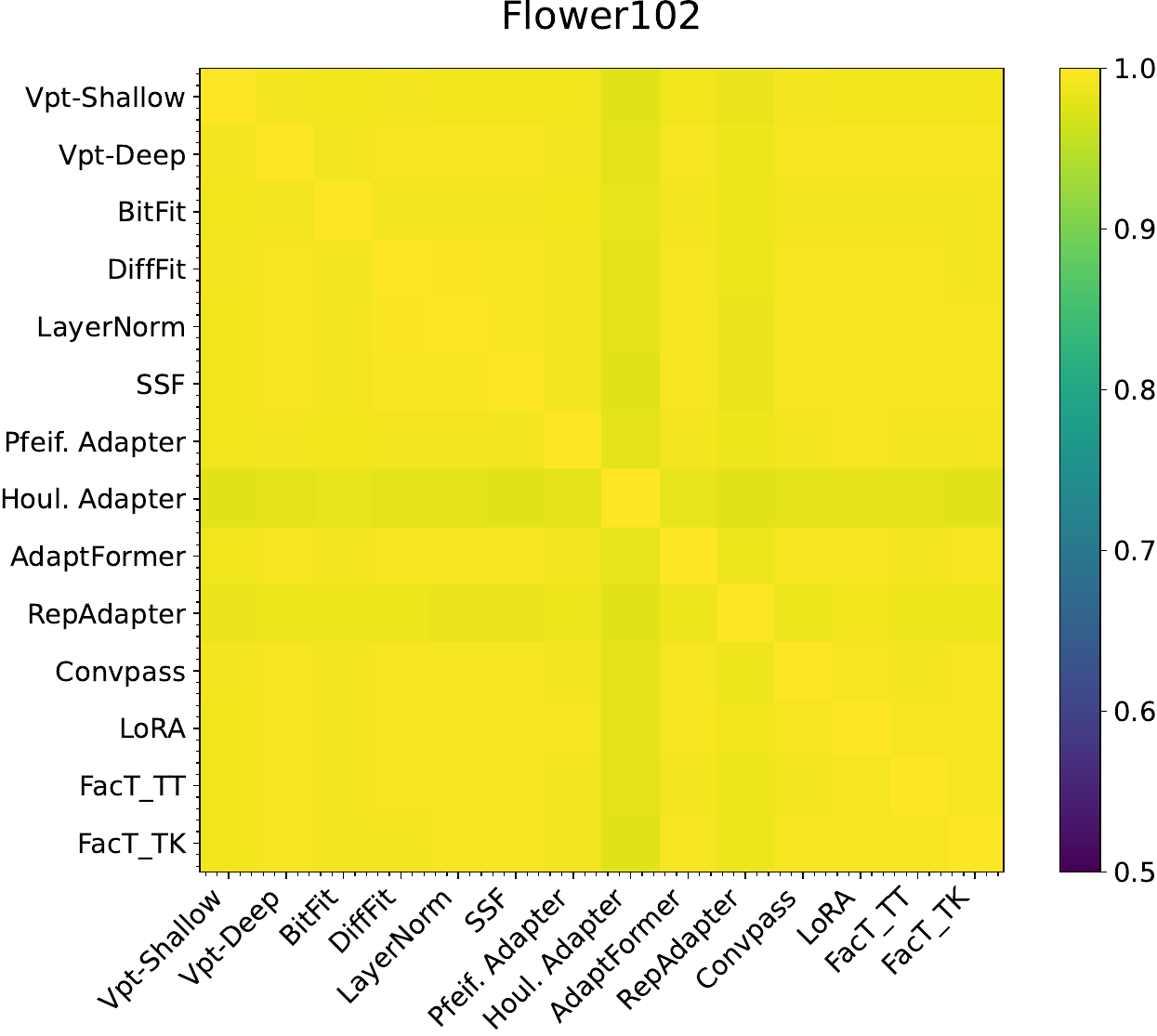}
    \caption{Flower102}
    \end{subfigure}%
    \hfill
    \begin{subfigure}{0.22\linewidth}
    \includegraphics[width=\linewidth]{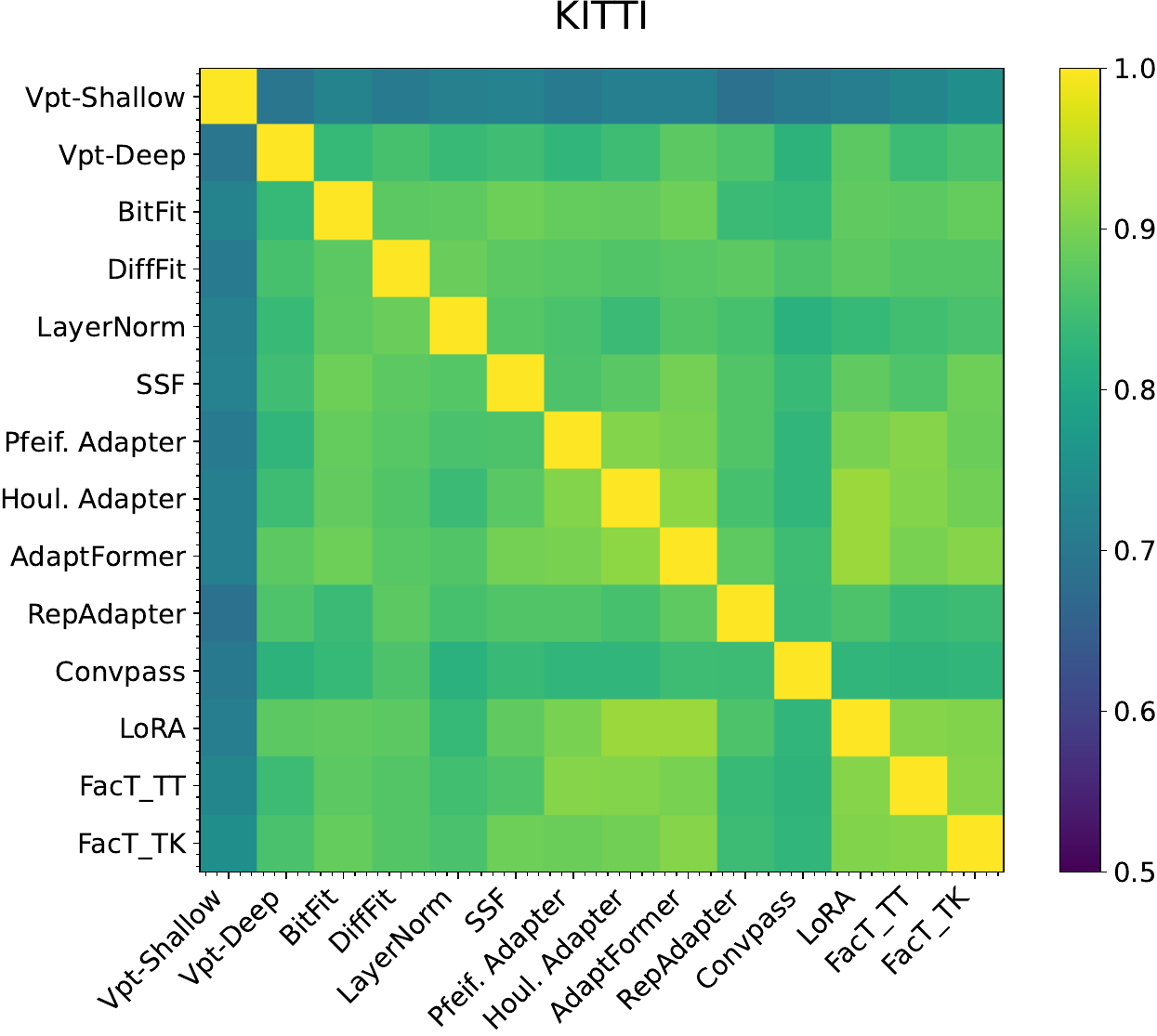}
    \caption{KITTI}
    \end{subfigure}%
    \hfill
    \begin{subfigure}{0.22\linewidth}
    \includegraphics[width=\linewidth]{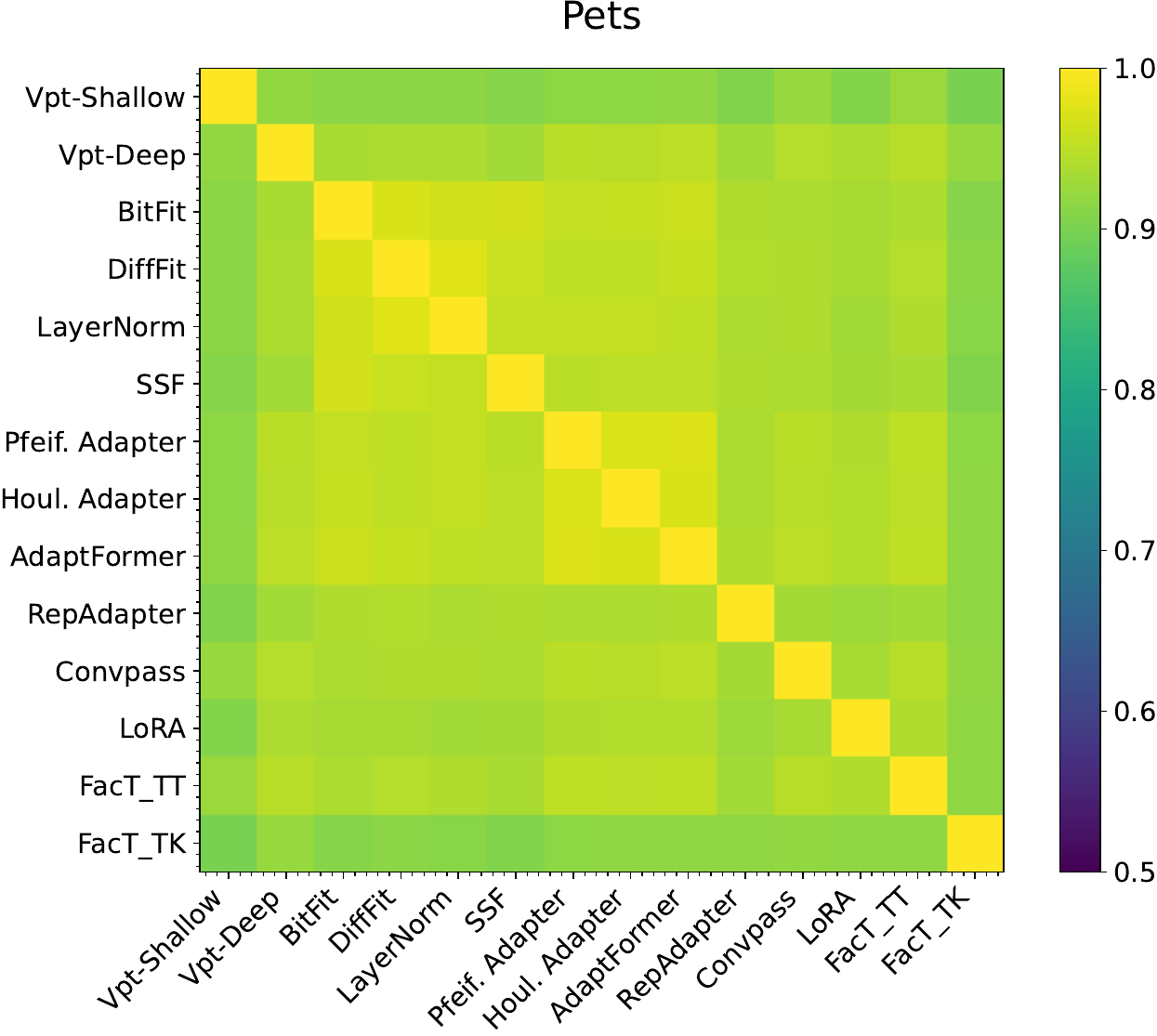}
    \caption{Pets}
    \end{subfigure}%
    \hfill
    \begin{subfigure}{0.22\linewidth}
    \includegraphics[width=\linewidth]{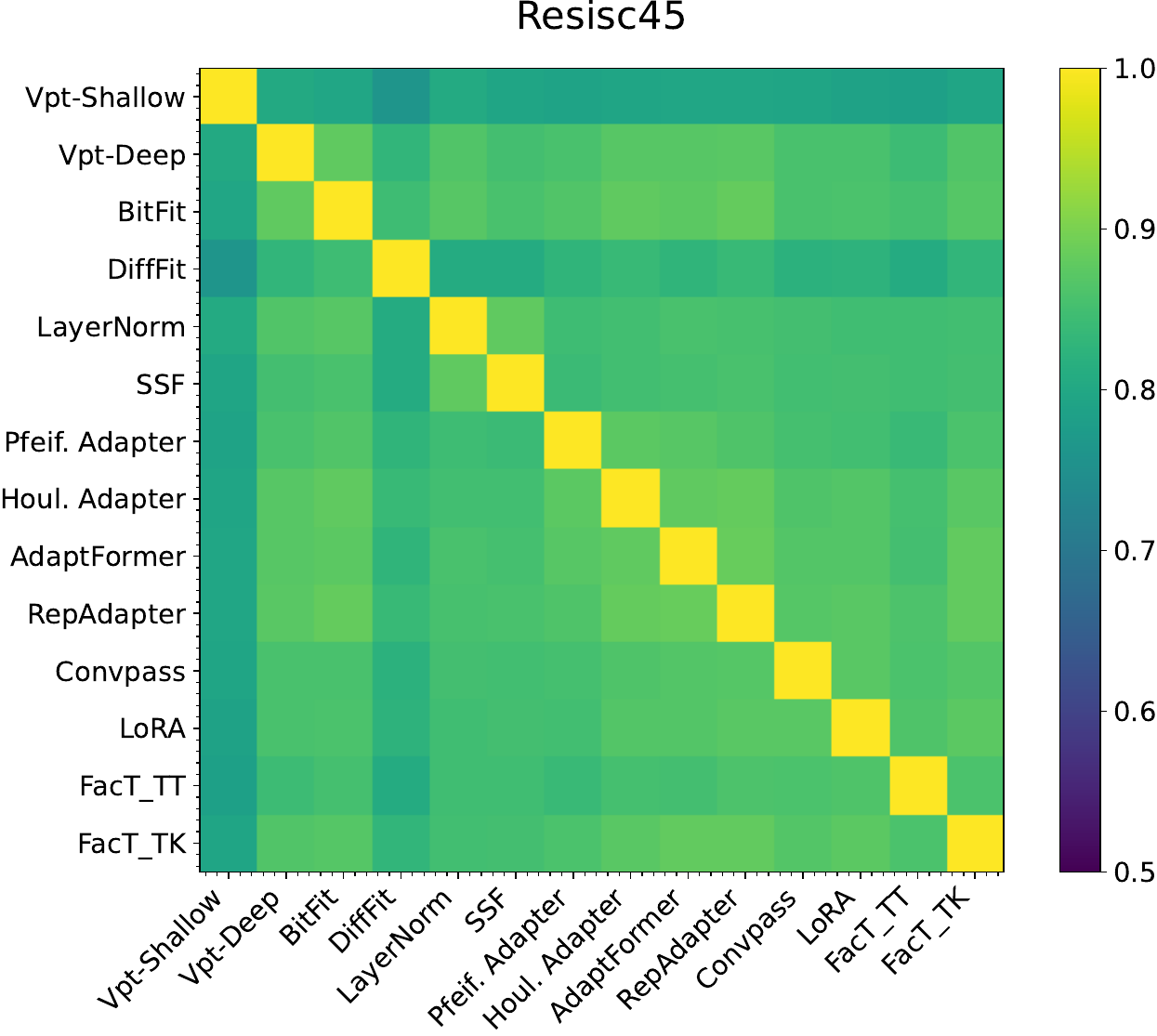}
    \caption{Resisc45}
    \end{subfigure}%

    \begin{subfigure}{0.22\linewidth}
    \includegraphics[width=\linewidth]{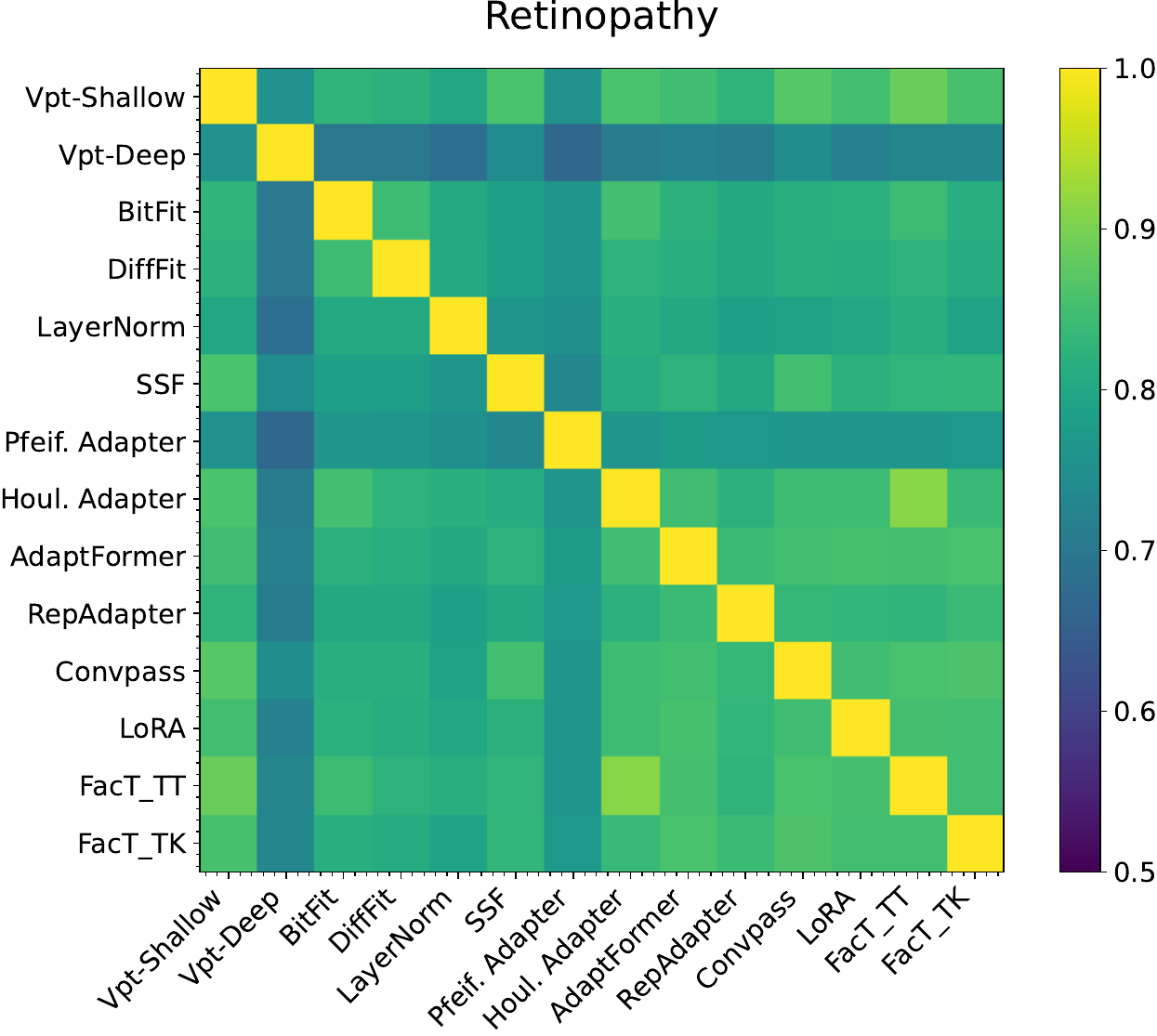}
    \caption{Retinopathy}
    \end{subfigure}%
    \hfill
    \begin{subfigure}{0.22\linewidth}
    \includegraphics[width=\linewidth]{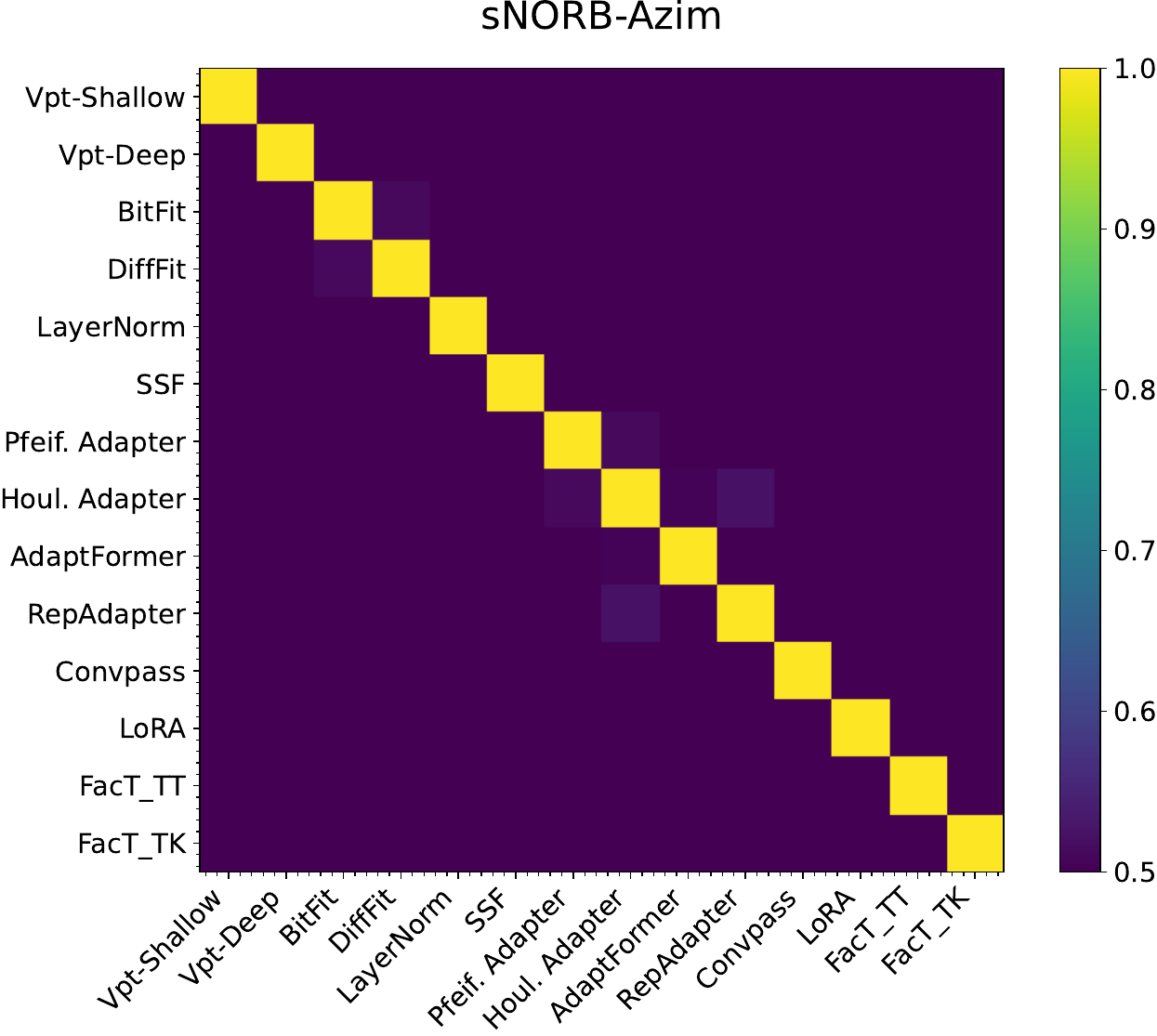}
    \caption{\scriptsize sNORB-Azim}
    \end{subfigure}%
    \hfill
    \begin{subfigure}{0.22\linewidth}
    \includegraphics[width=\linewidth]{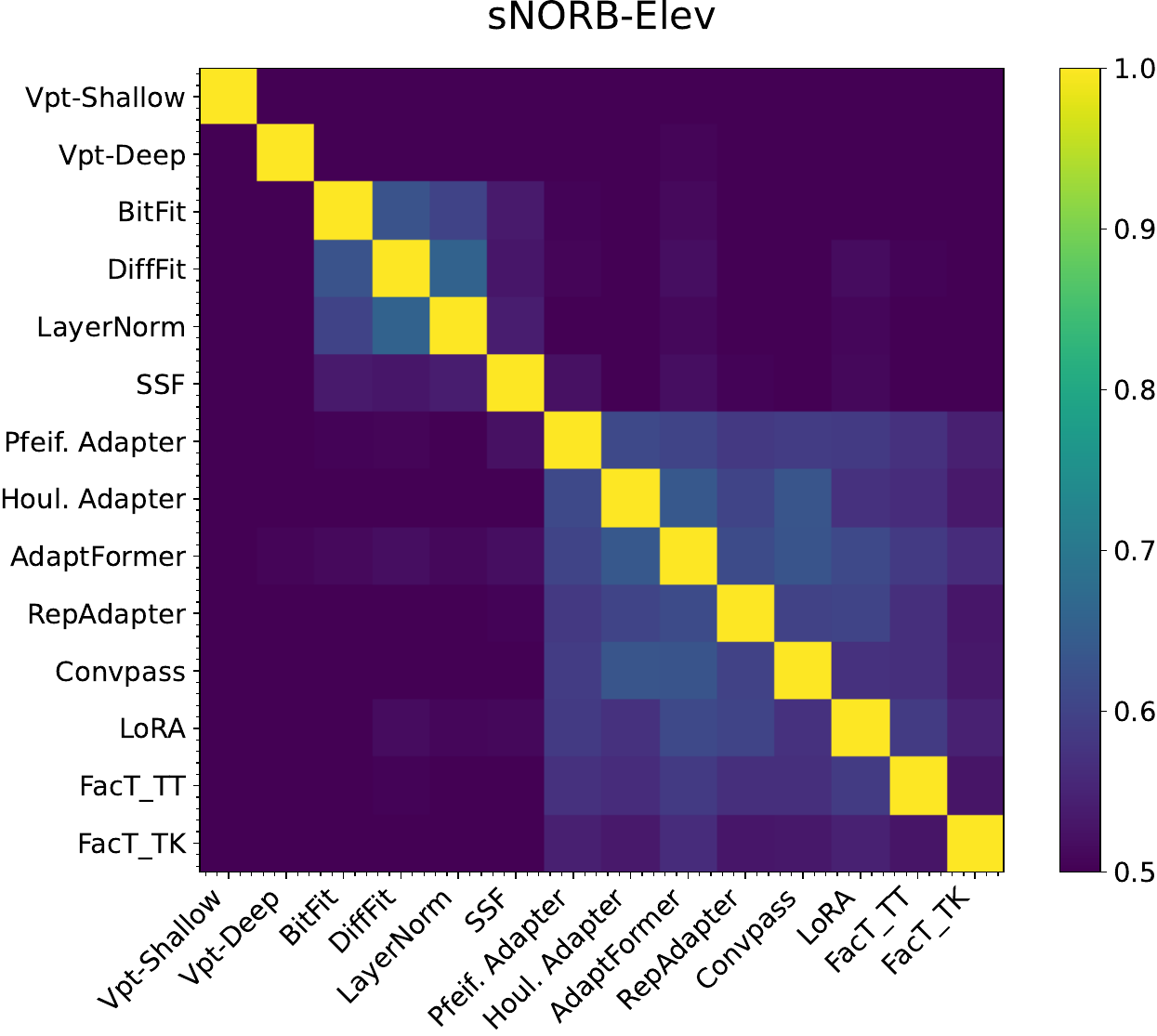}
    \caption{sNORB-Elev}
    \end{subfigure}%
    \hfill
    \begin{subfigure}{0.22\linewidth}
    \includegraphics[width=\linewidth]{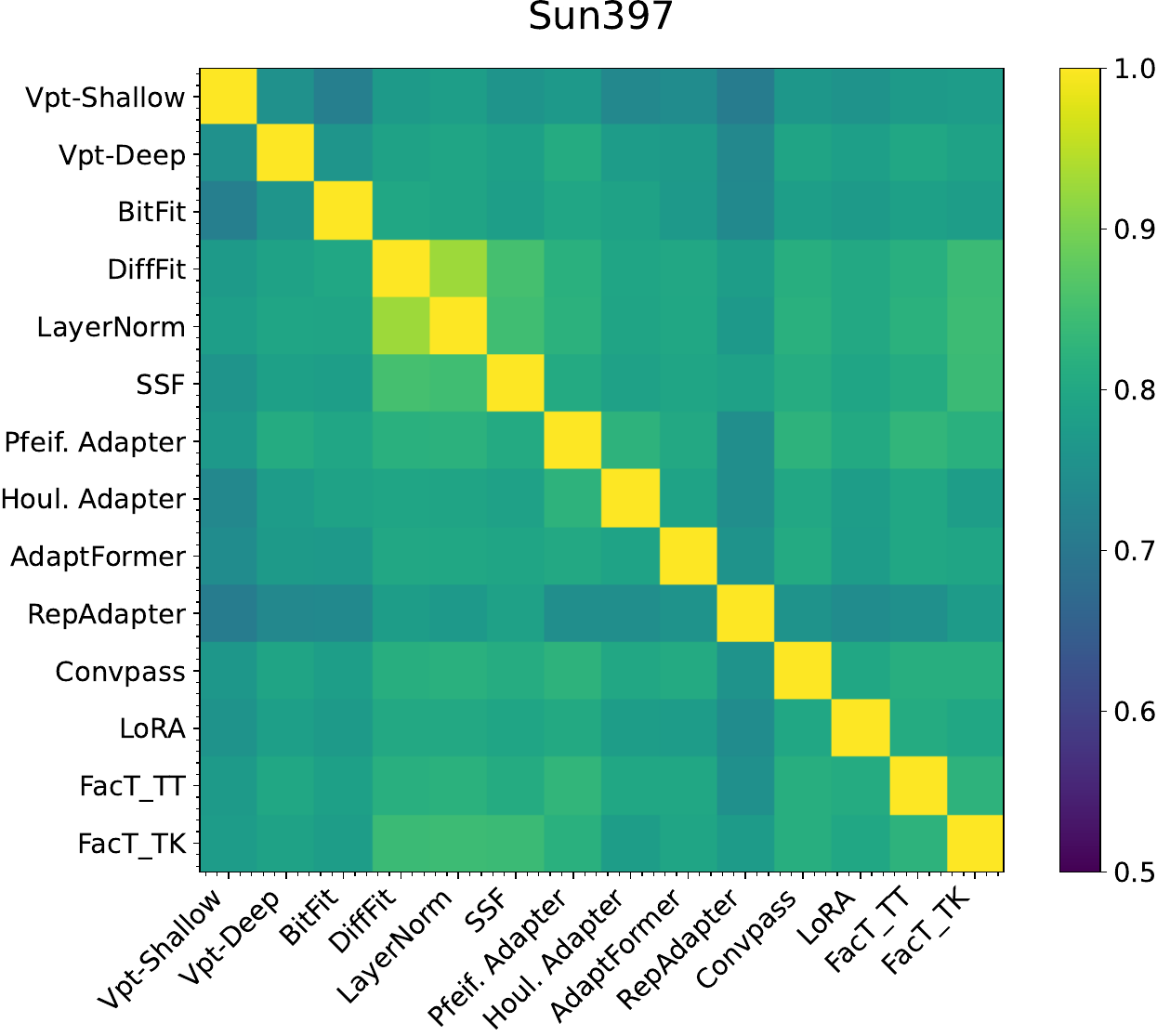}
    \caption{Sun397}
    \end{subfigure}%

    \begin{subfigure}{0.22\linewidth}
    \includegraphics[width=\linewidth]{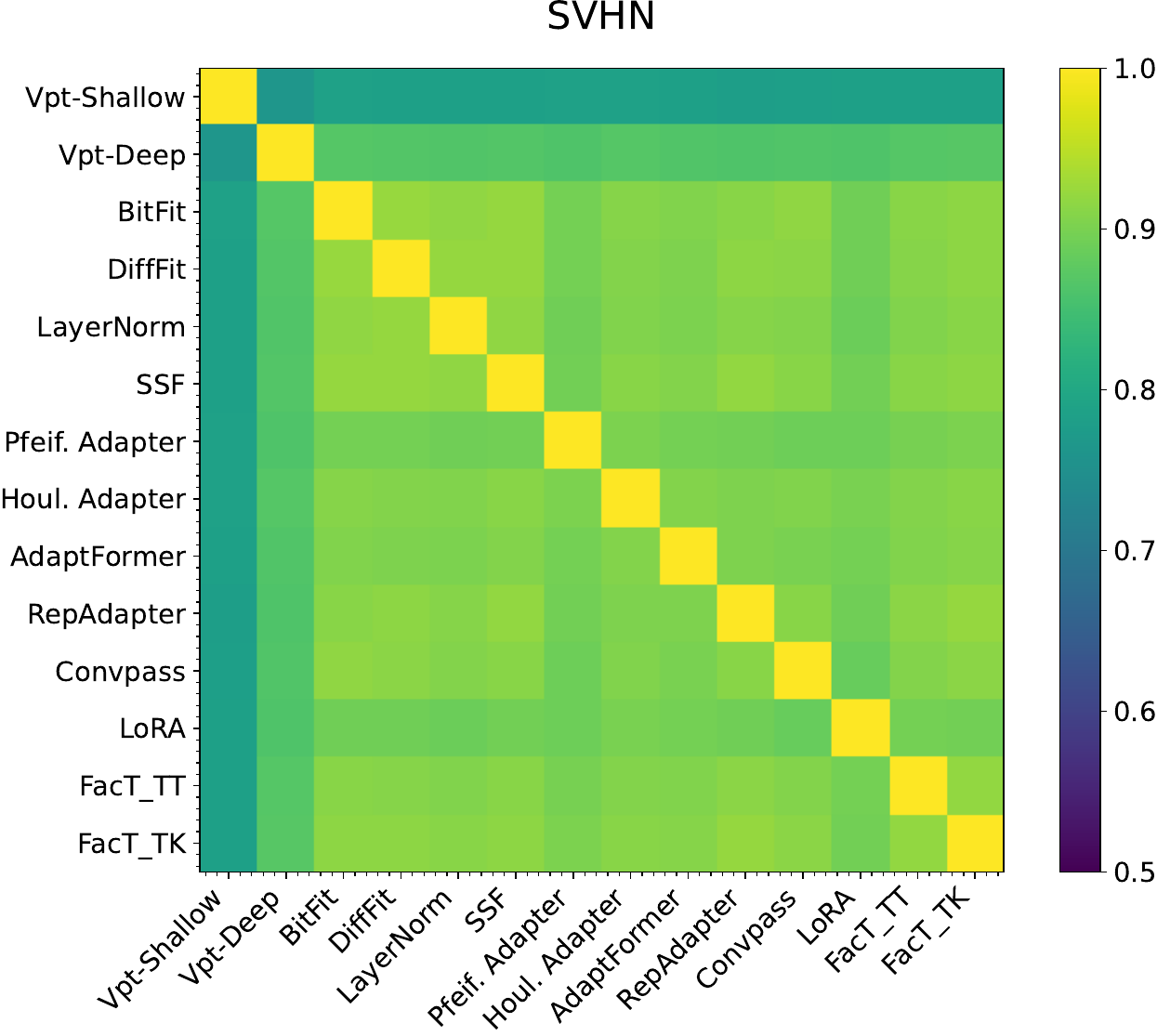}
    \caption{SVHN}
    \end{subfigure}%
    \hfill

    \caption{\small Prediction similarity analysis on other datasets.}
    \label{-fig:pred_sim_others}
\end{figure*}

\autoref{-fig:pred_sim_others} shows the prediction analysis discussed in \autoref{sec:complememtary} for all the datasets in VTAB-1K. It is expected that their predictions are similar for datasets with very high accuracy, such as Flowers102 (avg 99.1\%) and Caltech101 (avg 91.4\%). Beyond them, we find that most PEFT methods show diverse predictions in other datasets in VTAB-1K.

\paragraph{Prediction similarity within the same PEFT group. }

\begin{figure*}[h!]
    \centering
    \small 
    \begin{minipage}[b]{0.34\textwidth}
        \centering
        \includegraphics[width=\textwidth]{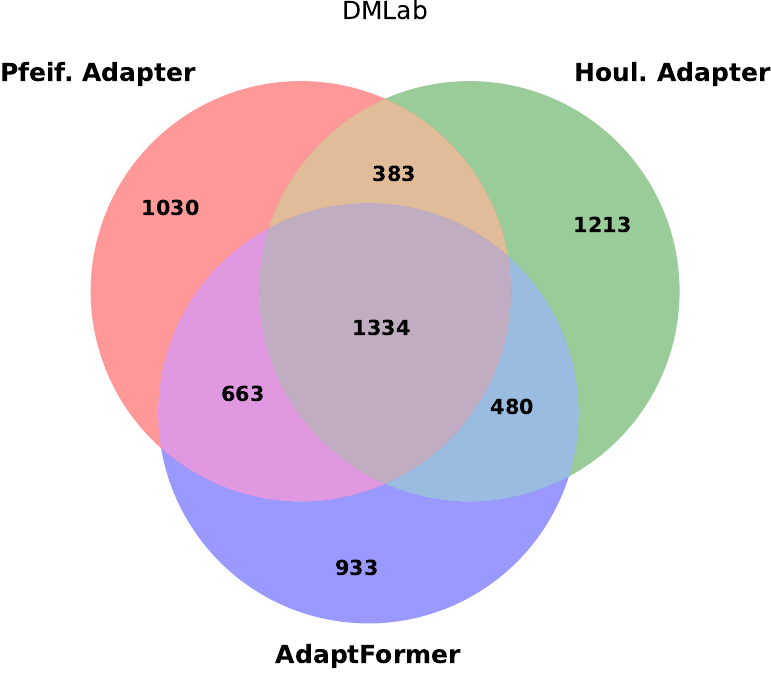}
        \subcaption{\small Adapter-based methods}
    \end{minipage}
    \begin{minipage}[b]{0.3\textwidth}
        \centering
        \includegraphics[width=\textwidth]{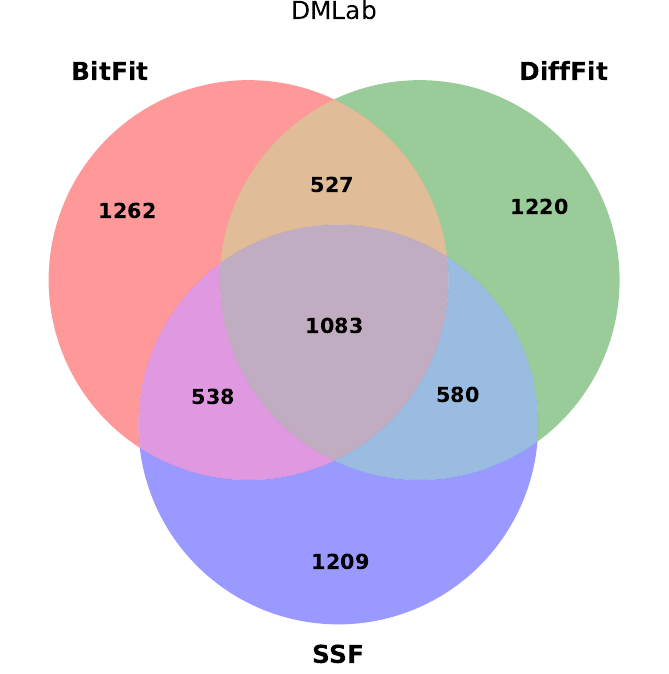}
        \subcaption{\small Selective tuning methods}
    \end{minipage}
    \begin{minipage}[b]{0.3\textwidth}
        \centering
        \includegraphics[width=\textwidth]{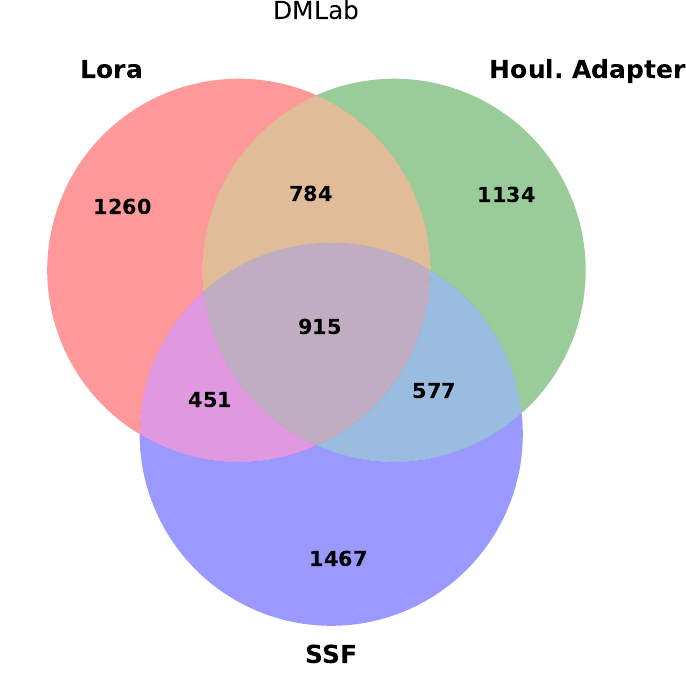}
        \subcaption{\small Methods from different groups}
    \end{minipage}
    \caption{Prediction overlap for the 5K most confident samples. Although methods from the same group share slightly more prediction overlap than methods from other groups, they still have quite different predictions}
    \label{fig:within_group_overlap}
\end{figure*}

To verify if methods within the same PEFT group share more prediction similarity, we plotted the prediction overlap for adapter-based methods, selective-tuning methods, and methods from different groups. As shown in \autoref{fig:within_group_overlap}, methods within the same group share slightly more prediction similarity than those from different groups, but they still exhibit distinct predictions. \autoref{fig:diversit_pred_part1} in the main paper also supports this observation. Methods are grouped based on the categories defined in \autoref{sec:PEFT-all}. If methods within the same group had very high similarities, we would see bright squares, which are only slightly evident around BitFit, DiffFit, LayerNorm, and SSF.

\begin{figure*}
    \centering
    \begin{subfigure}{0.4\linewidth}
    \includegraphics[width=\linewidth]{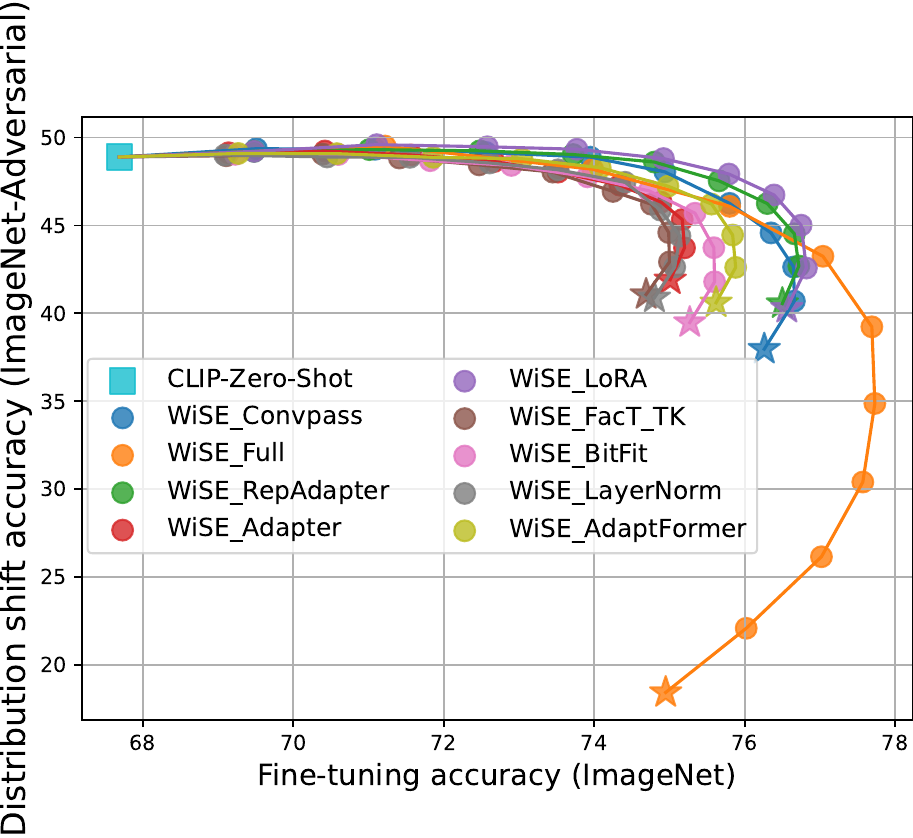}
    \caption{ImageNet-A}
    \end{subfigure}%
    \hfill
    \begin{subfigure}{0.4\linewidth}
    \includegraphics[width=\linewidth]{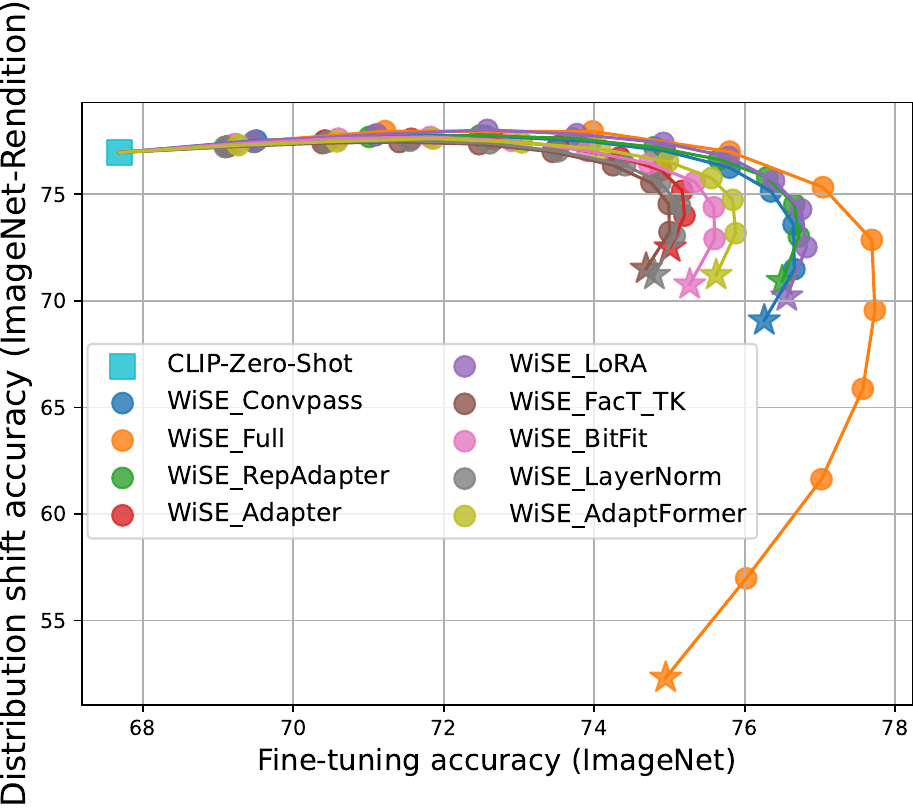}
    \caption{ImageNet-R}
    \end{subfigure}%
    
    \begin{subfigure}{0.4\linewidth}
    \includegraphics[width=\linewidth]{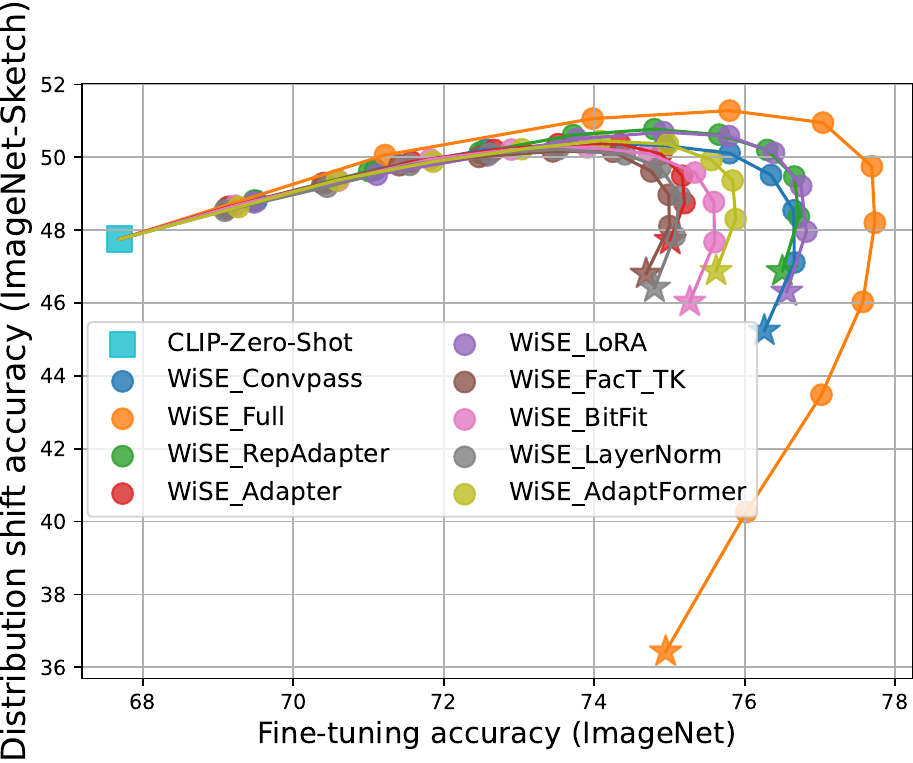}
    \caption{ImageNet-S}
    \end{subfigure}%
    \hfill
    \begin{subfigure}{0.4\linewidth}
    \includegraphics[width=\linewidth]{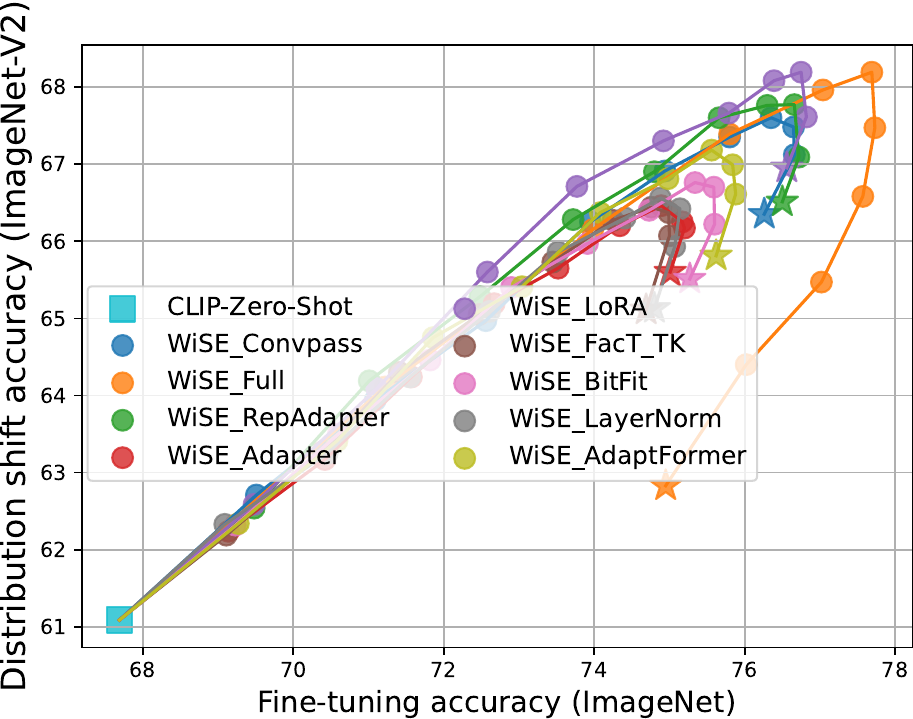}
    \caption{ImageNet-V2}
    \end{subfigure}%
    
    \caption{\small WiSE PEFT performance on all distribution shift datasets. Target distribution  vs.~distribution shifts}
    \label{-fig:merge_others}
\end{figure*}

\paragraph{WiSE PEFT results for all distribution shift datasets. }
We provide detailed WiSE PEFT performance for each distribution shift dataset in \autoref{-fig:merge_others}. WiSE improves both the robustness and the in-distribution performance of PEFT methods. Interestingly, even though full fine-tuning is generally less robust than PEFT methods, applying WiSE allows it to achieve better performance in both target distribution and distribution shift data.

\paragraph{Performance comparison between DINOv2 and IN21k.} To compare the performance of DINOv2 and IN21k, we selected several PEFT methods and datasets from VTAB-1K. The results presented in \autoref{tb:dino_results} reveal several interesting findings:

(1) \textbf{Improved Linear Probing Performance}: Linear probing generally shows improved results with DINOv2, indicating that its extracted features are more robust and discriminative than those from IN21k.

(2) \textbf{Deteriorated Full Fine-Tuning Performance}: Conversely, full fine-tuning performance significantly worsens with DINOv2, suggesting that models fully fine-tuned on DINOv2 are more susceptible to overfitting.

(3) \textbf{Adapter-Based Methods Performance}: Among the three adapter-based methods evaluated—Houl. Adapter, AdaptFormer, and Convpass—we observe performance enhancements in most datasets for AdaptFormer and Convpass. In contrast, Houl. Adapter exhibits significant degradation across all datasets. This disparity may be attributed to architectural differences: AdaptFormer and Convpass insert their adapter modules in parallel with existing modules such as Multi-Head Self-Attention (MSA) and/or Multi-Layer Perceptron (MLP), whereas Houl. Adapter inserts its adapter sequentially after the MSA and MLP layers. We hypothesize that the sequential design of Houl. Adapter leads to more substantial alterations of intermediate features compared to the parallel design, potentially explaining the observed decrease in performance.

\begin{table*}
\centering
\resizebox{0.7\textwidth}{!}{

\begin{tabular}{|c|c|c|ccccccc|}
\hline
\multicolumn{1}{|l|}{} & \multicolumn{1}{l|}{}        & \multicolumn{1}{l|}{}    & \textbf{Linear} & \textbf{Full} & \textbf{SSF} & \makecell{\textbf{Houl.}\\\textbf{Adapter}} & \makecell{\textbf{Adapt-}\\\textbf{Former}} & \textbf{Convpass} & \textbf{LoRA} \\ \hline

\multirow{12}{*}{\textbf{Natural}} & \multirow{3}{*}{\textbf{Caltech101}} & \textbf{Dinov2} & 89.8 & 83.2 & 90.3 & 22.1 & 92.5 & 91.7 & 92.3 \\ \cline{3-10}
& & \textbf{IN21k} & 86.6 & 89.9 & 89.8 & 92.1 & 91.8 & 92.1 & 92.6 \\ \cline{3-10}
& & $\Delta$ & \textcolor[HTML]{38761D}{\textbf{3.2}} & \textcolor[HTML]{FF0000}{\textbf{-6.7}} & \textcolor[HTML]{38761D}{\textbf{0.5}} & \textcolor[HTML]{FF0000}{\textbf{-70.0}} & \textcolor[HTML]{38761D}{\textbf{0.7}} & \textcolor[HTML]{FF0000}{\textbf{-0.4}} & \textcolor[HTML]{FF0000}{\textbf{-0.3}} \\ \cline{2-10}

& \multirow{3}{*}{\textbf{DTD}} & \textbf{Dinov2} & 74.9 & 45.2 & 77.0 & 14.6 & 78.8 & 77.0 & 78.4 \\ \cline{3-10}
& & \textbf{IN21k} & 65.7 & 61.9 & 68.8 & 72.3 & 70.5 & 72.0 & 69.8 \\ \cline{3-10}
& & $\Delta$ & \textcolor[HTML]{38761D}{\textbf{9.2}} & \textcolor[HTML]{FF0000}{\textbf{-16.7}} & \textcolor[HTML]{38761D}{\textbf{8.2}} & \textcolor[HTML]{FF0000}{\textbf{-57.7}} & \textcolor[HTML]{38761D}{\textbf{8.3}} & \textcolor[HTML]{38761D}{\textbf{5.0}} & \textcolor[HTML]{38761D}{\textbf{8.6}} \\ \cline{2-10}

& \multirow{3}{*}{\textbf{Pets}} & \textbf{Dinov2} & 93.4 & 68.7 & 92.6 & 7.1 & 94.0 & 92.3 & 94.1 \\ \cline{3-10}
& & \textbf{IN21k} & 89.3 & 85.8 & 91.4 & 91.7 & 91.8 & 91.3 & 90.5 \\ \cline{3-10}
& & $\Delta$ & \textcolor[HTML]{38761D}{\textbf{4.1}} & \textcolor[HTML]{FF0000}{\textbf{-17.1}} & \textcolor[HTML]{38761D}{\textbf{1.2}} & \textcolor[HTML]{FF0000}{\textbf{-84.6}} & \textcolor[HTML]{38761D}{\textbf{2.2}} & \textcolor[HTML]{38761D}{\textbf{1.0}} & \textcolor[HTML]{38761D}{\textbf{3.6}} \\ \cline{2-10}

& \multirow{3}{*}{\textbf{Sun397}} & \textbf{Dinov2} & 55.1 & 23.9 & 52.5 & 2.6 & 56.5 & 56.1 & 55.7 \\ \cline{3-10}
& & \textbf{IN21k} & 53.2 & 36.8 & 56.5 & 55.4 & 56.7 & 55.9 & 55.5 \\ \cline{3-10}
& & $\Delta$ & \textcolor[HTML]{38761D}{\textbf{1.9}} & \textcolor[HTML]{FF0000}{\textbf{-12.9}} & \textcolor[HTML]{FF0000}{\textbf{-4.0}} & \textcolor[HTML]{FF0000}{\textbf{-52.8}} & \textcolor[HTML]{FF0000}{\textbf{-0.2}} & \textcolor[HTML]{38761D}{\textbf{0.2}} & \textcolor[HTML]{38761D}{\textbf{0.2}} \\ \hline

\multirow{12}{*}{\textbf{Specialized}} & \multirow{3}{*}{\textbf{Camelyon}} & \textbf{Dinov2} & 83.2 & 77.9 & 83.9 & 77.1 & 84.4 & 86.3 & 85.4 \\ \cline{3-10}
& & \textbf{IN21k} & 83.1 & 81.6 & 86.1 & 88.7 & 86.8 & 87.7 & 87.5 \\ \cline{3-10}
& & $\Delta$ & \textcolor[HTML]{38761D}{\textbf{0.1}} & \textcolor[HTML]{FF0000}{\textbf{-3.7}} & \textcolor[HTML]{FF0000}{\textbf{-2.2}} & \textcolor[HTML]{FF0000}{\textbf{-11.6}} & \textcolor[HTML]{FF0000}{\textbf{-2.4}} & \textcolor[HTML]{FF0000}{\textbf{-1.4}} & \textcolor[HTML]{FF0000}{\textbf{-2.1}} \\ \cline{2-10}

& \multirow{3}{*}{\textbf{EuroSAT}} & \textbf{Dinov2} & 89.2 & 66.9 & 93.9 & 60.2 & 93.8 & 93.8 & 94.2 \\ \cline{3-10}
& & \textbf{IN21k} & 90.0 & 88.1 & 94.5 & 95.3 & 95.0 & 95.8 & 94.9 \\ \cline{3-10}
& & $\Delta$ & \textcolor[HTML]{FF0000}{\textbf{-0.8}} & \textcolor[HTML]{FF0000}{\textbf{-21.2}} & \textcolor[HTML]{FF0000}{\textbf{-0.6}} & \textcolor[HTML]{FF0000}{\textbf{-35.1}} & \textcolor[HTML]{FF0000}{\textbf{-1.2}} & \textcolor[HTML]{FF0000}{\textbf{-2.0}} & \textcolor[HTML]{FF0000}{\textbf{-0.7}} \\ \cline{2-10}

& \multirow{3}{*}{\textbf{Resisc45}} & \textbf{Dinov2} & 78.8 & 25.9 & 82.0 & 24.5 & 88.6 & 87.6 & 84.2 \\ \cline{3-10}
& & \textbf{IN21k} & 74.9 & 81.6 & 83.2 & 86.5 & 86.5 & 85.9 & 85.9 \\ \cline{3-10}
& & $\Delta$ & \textcolor[HTML]{38761D}{\textbf{3.9}} & \textcolor[HTML]{FF0000}{\textbf{-55.7}} & \textcolor[HTML]{FF0000}{\textbf{-1.2}} & \textcolor[HTML]{FF0000}{\textbf{-62.0}} & \textcolor[HTML]{38761D}{\textbf{2.1}} & \textcolor[HTML]{38761D}{\textbf{1.7}} & \textcolor[HTML]{FF0000}{\textbf{-1.7}} \\ \cline{2-10}

& \multirow{3}{*}{\textbf{Retinopathy}} & \textbf{Dinov2} & 75.3 & 73.6 & 76.0 & 73.6 & 76.0 & 76.0 & 75.5 \\ \cline{3-10}
& & \textbf{IN21k} & 74.6 & 73.6 & 74.8 & 75.2 & 76.3 & 75.9 & 75.7 \\ \cline{3-10}
& & $\Delta$ & \textcolor[HTML]{38761D}{\textbf{0.7}} & \textcolor[HTML]{FF0000}{\textbf{0.0}} & \textcolor[HTML]{38761D}{\textbf{1.2}} & \textcolor[HTML]{FF0000}{\textbf{-1.6}} & \textcolor[HTML]{FF0000}{\textbf{-0.3}} & \textcolor[HTML]{38761D}{\textbf{0.1}} & \textcolor[HTML]{FF0000}{\textbf{-0.2}} \\ \hline

\multirow{12}{*}{\textbf{Structured}} & \multirow{3}{*}{\textbf{Clevr-Count}} & \textbf{Dinov2} & 47.5 & 27.3 & 71.2 & 38.1 & 91.2 & 87.8 & 89.8 \\ \cline{3-10}
& & \textbf{IN21k} & 37.5 & 56.2 & 80.1 & 82.9 & 82.9 & 82.3 & 82.9 \\ \cline{3-10}
& & $\Delta$ & \textcolor[HTML]{38761D}{\textbf{10.0}} & \textcolor[HTML]{FF0000}{\textbf{-28.9}} & \textcolor[HTML]{FF0000}{\textbf{-8.9}} & \textcolor[HTML]{FF0000}{\textbf{-44.8}} & \textcolor[HTML]{38761D}{\textbf{8.3}} & \textcolor[HTML]{38761D}{\textbf{5.5}} & \textcolor[HTML]{38761D}{\textbf{6.9}} \\ \cline{2-10}

& \multirow{3}{*}{\textbf{DMLab}} & \textbf{Dinov2} & 44.1 & 30.7 & 51.8 & 39.1 & 51.8 & 53.1 & 54.5 \\ \cline{3-10}
& & \textbf{IN21k} & 36.5 & 48.2 & 53.0 & 53.8 & 52.8 & 53.8 & 51.8 \\ \cline{3-10}
& & $\Delta$ & \textcolor[HTML]{38761D}{\textbf{7.6}} & \textcolor[HTML]{FF0000}{\textbf{-17.5}} & \textcolor[HTML]{FF0000}{\textbf{-1.2}} & \textcolor[HTML]{FF0000}{\textbf{-14.7}} & \textcolor[HTML]{FF0000}{\textbf{-1.0}} & \textcolor[HTML]{FF0000}{\textbf{-0.7}} & \textcolor[HTML]{38761D}{\textbf{2.7}} \\ \cline{2-10}

& \multirow{3}{*}{\textbf{KITTI}} & \textbf{Dinov2} & 60.3 & 47.1 & 81.0 & 46.8 & 83.4 & 82.6 & 83.8 \\ \cline{3-10}
& & \textbf{IN21k} & 64.6 & 77.9 & 81.4 & 79.6 & 80.0 & 78.1 & 79.9 \\ \cline{3-10}
& & $\Delta$ & \textcolor[HTML]{FF0000}{\textbf{-4.3}} & \textcolor[HTML]{FF0000}{\textbf{-30.8}} & \textcolor[HTML]{FF0000}{\textbf{-0.4}} & \textcolor[HTML]{FF0000}{\textbf{-32.8}} & \textcolor[HTML]{38761D}{\textbf{3.4}} & \textcolor[HTML]{38761D}{\textbf{4.5}} & \textcolor[HTML]{38761D}{\textbf{3.9}} \\ \cline{2-10}

& \multirow{3}{*}{\textbf{dSpr-Ori}} & \textbf{Dinov2} & 47.2 & 17.5 & 56.1 & 10.0 & 57.9 & 55.6 & 57.2 \\ \cline{3-10}
& & \textbf{IN21k} & 29.4 & 46.6 & 52.1 & 54.3 & 53.0 & 55.3 & 47.2 \\ \cline{3-10}
& & $\Delta$ & \textcolor[HTML]{38761D}{\textbf{17.8}} & \textcolor[HTML]{FF0000}{\textbf{-29.1}} & \textcolor[HTML]{38761D}{\textbf{4.0}} & \textcolor[HTML]{FF0000}{\textbf{-44.3}} & \textcolor[HTML]{38761D}{\textbf{4.9}} & \textcolor[HTML]{38761D}{\textbf{0.3}} & \textcolor[HTML]{38761D}{\textbf{10.0}} \\ \hline

\end{tabular}}
\caption{Performance comparison between DINOv2 and IN21k. }
\label{tb:dino_results}
\end{table*}

\paragraph{\autoref{fig:vtab_vis} details. }  This figure illustrates the relative performance compared to linear probing ({$\times$}) on VTAB-1K. The range between the highest and lowest accuracy across 14 PEFT methods is represented by {\color{red}$\bullet$-$\bullet$}, while ({\color{blue}$\blacksquare$}) denotes the performance of full fine-tuning.

\paragraph{\autoref{fig:merge} details. } The X-axis represents the accuracy on ImageNet-1K, while the Y-axis shows the distribution shift accuracy (averaged across ImageNet-V2, ImageNet-S, ImageNet-R, and ImageNet-A). The cyan squares ({\color{cyan}$\blacksquare$}) represent the zero-shot performance of the CLIP model, and stars ($\star$) denote the performance of fine-tuned models. Each curve corresponds to the WiSE+PEFT method, with dots $\bullet$ indicating different mixing coefficients $\alpha$ as described in \autoref{sec: robust}.

\paragraph{\autoref{fig:ranking} details. } For each dataset in VTAB-1K, 15 methods (14 PEFT methods plus linear probing) are ranked by accuracy. Within each dataset group (\eg., Natural), the element $(i, j)$ in the ranking frequency matrix indicates how often method $i$ ranks $j^{th}$. For instance, in the Natural group matrix, the entry (1, 3) equals 2, meaning DiffFit ranked 3rd in two datasets within this group. The row sums correspond to the total number of datasets in each group (\eg., 7 datasets for the Natural group). Methods are sorted by their average rank (shown in brackets), and the parameters column indicates the number of trainable parameters in millions.

\paragraph{\autoref{fig:diversit_pred_part1} details. } Each entry $(i,j)$ in the prediction similarity matrix represents the percentage of test samples for which methods $i$ and $j$ made the same prediction. The diagonal entries are always 1, indicating perfect agreement with themselves. To compute $(i,j)$, predictions from models fine-tuned by methods $i$ and $j$ are compared, with $(i,j)$ equaling the number of matching predictions divided by the total test samples.

\paragraph{\autoref{fig:diversit_pred_part2} details. } The Venn diagrams are generated by fine-tuning a pre-trained model on CIFAR100 (VTAB-1K) using LoRA, SSF, and Adapter methods. For \autoref{fig:diversit_pred_part2}(a), we selected the correct predictions from the top 5K most confident samples for each method and visualized the overlap among the three methods. For \autoref{fig:diversit_pred_part2}(b), we did the same for the wrong predictions, selecting from the 5K least confident samples.

\paragraph{\autoref{fig:ensemble} details. } For each VTAB-1K dataset, the worst-performing PEFT method serves as the baseline ($\crossline$). Each $\redcircle$ represents the relative performance of other PEFT methods compared to this baseline. An ensemble prediction ($\greentriangle$) is generated based on the average logits of all PEFT methods for each test sample.

\paragraph{\autoref{fig:many_shots} details. }  Different colors represent various PEFT methods. Each $\bullet$ along a curve (corresponding to a single PEFT method) indicates the accuracy at a specific tunable parameter size, allowing us to observe how the size of tunable parameters impacts accuracy.

\paragraph{\autoref{fig:why} details. } Each sub-figure displays accuracy on the Y-axis, with columns representing linear probing (left), the best PEFT methods (middle), and full fine-tuning (right). Sub-figures (a) and (b) correspond to VTAB-1K (low-shot), while (c) and (d) correspond to many-shot settings. Different colors represent distinct datasets.

\section{Broader Impacts}
\label{-sec:impacts}

Our study provides a unifying study of PEFT in visual recognition. We expect it to serve as a valuable practical user guide to benefit society. Specifically, fine-tuning large models needs significant computation. A unifying study of PEFT will ease end-users to apply more parameter-efficient and computation-efficient ways for fine-tuning. To our knowledge, our paper does not introduce any additional negative societal impacts compared to existing papers on PEFT.


\end{document}